\documentclass{article}
\usepackage[utf8]{inputenc}
\usepackage{amsmath}
\usepackage{amsfonts}
\usepackage{amssymb}
\usepackage{graphicx}
\usepackage[ruled,vlined]{algorithm2e}
\usepackage{pdflscape}
\usepackage[a4paper, total={6in, 8in}]{geometry}
\usepackage{lmodern}
\usepackage[dvipsnames]{xcolor}
\usepackage{setspace} 
\usepackage{epstopdf}
\usepackage{subfigure}
\usepackage{hyperref}
\usepackage{natbib}

\begin{document}

	
\title{A continuous-state cellular automata algorithm for global optimization}

\date{March, 2021}

\author{Juan Carlos Seck-Tuoh-Mora, Norberto Hernandez-Romero, Pedro Lagos-Eulogio, \\
Joselito Medina-Marin, Nadia Samantha Zuñiga-Peña \\
Área Académica de Ingeniería, Instituto de Ciencias Básicas e Ingeniería \\
Universidad Autónoma del Estado de Hidalgo \\ Carr Pachuca-Tulancingo Km 4.5. Pachuca 42184 Hidalgo. Mexico \\
jseck@uaeh.edu.mx, nhromero@uaeh.edu.mx, plagos@uaeh.edu.mx\\
jmedina@uaeh.edu.mx, zu450520@uaeh.edu.mx}

\maketitle

\begin{abstract}
\noindent Cellular automata are capable of developing complex behaviors based on simple local interactions between their elements. Some of these characteristics have been used to propose and improve meta-heuristics for global optimization; however, the properties offered by the evolution rules in cellular automata have not yet been used directly in optimization tasks. Inspired by the complexity that various evolution rules of cellular automata can offer, the continuous-state cellular automata algorithm (CCAA) is proposed. In this way, the CCAA takes advantage of different evolution rules to maintain a balance that maximizes the exploration and exploitation properties in each iteration. The efficiency of the CCAA is proven with $33$ test problems widely used in the literature, $4$ engineering applications that were also used in recent literature, and the design of adaptive infinite-impulse response (IIR) filters, testing  $10$ full-order IIR reference functions. The numerical results prove its competitiveness in comparison with state-of-the-art algorithms. The source codes of the CCAA are publicly available at \url{https://github.com/juanseck/CCAA.git}. \\


\noindent Submitted to Expert Systems With Applications
\end{abstract}



\section{Introduction}
Nowadays, the constant demand for reducing costs and production times and the unstoppable competition in the engineering area has driven the search for advanced decision-making methods, such as optimization methods, to design and produce goods economically and efficiently. Among these optimization methods are meta-heuristics, which are high-level algorithms with which good enough solutions to optimization problems can be found. \citep{Dwivedi2020}.

Meta-heuristic algorithms are of great relevance because, in general, they are easy to implement and their operation is based on simple concepts that do not require information regarding the gradient of the function to be optimized. A good meta-heuristic algorithm can escape from local optima and be used in a wide range of optimization, design, and parameter identification problems.

For that purpose, meta-heuristic algorithms must perform in an efficient and balanced way the exploration of the solution space and the exploitation of promising areas of said space. Since these actions can conflict and leave the algorithm trapped in a local minimum, the correct balance of both processes is essential to obtain a meta-heuristic algorithm that calculates good solutions, avoiding stagnation or premature convergence towards local minima.

The optimization problem, which consists of finding the best sets of parameters that optimize a given objective function, subject to restrictions, has been approached from different perspectives. One of them is represented by algorithms inspired by physical and natural systems, with variants and modifications that intensify their exploitation and exploration characteristics. One of the best-known algorithms is the genetic algorithm (GA) \citep{holland1984genetic}, which has been hybridized and adapted multiple times to optimize a broader range of problems. An example is where it is combined with neural networks to enhance the capacity to optimize complex nonlinear problems \citep{qiao2020nature}. Another example is the particle swarm optimization (PSO) algorithm  \citep{eberhart1995new}, in which, unlike the GA, all particles have memory and share knowledge. Like the GA, it has been modified and improved multiple times; for example, in the study by \citep{yang2020improved}, PSO was improved based on entropy, which reduces the swarm's randomness and increases the diversity of the population. An algorithm that can be considered a modification and hybridization of the GA and the PSO \citep{yang2020improved} is the differential evolution (DE) algorithm, which treats individuals as strings of real numbers, making encoding and decoding for continuous problems unnecessary. In the case of the DE, we can also find multiple modifications and improvements, such as proposing a dual strategy strengthening global exploration and population diversity \citep{zhong2020improved}.

Another algorithm based on non-trivial collective behavior that originates from sharing information among the population members is the ant colony optimization algorithm (ACO) \citep{dorigo1997ant}. ACO has been  studied deeply and multiple recent variants have been presented \citep{chatterjee2020adaptive} \citep{liu2020improving} \citep{yu2020novel}. Another algorithm inspired by the global behavior that performs a complex task, such as foraging, based on its individuals' interactions, is the artificial bee colony algorithm (ABC)  \citep{karaboga2005idea}. In the same way, many variants and improvements have been recently proposed \citep {huang2019improved} \citep {wang2020artificial}, which shows that algorithms based on population behaviors continue to be an active research area.

In addition to the variants and modifications of population-based algorithms, we can find in the recent literature various works where algorithms are hybridized to enhance the search properties for particular cases \citep{Lagos-Eulogio2017} \citep{qiao2020nature} \citep{qu2020novel} \citep{moayedi2020hybridizing} or those that adapt not only the parameters but also the procedure used, such as the drone squad optimization (DSO) algorithm   \citep{de2018drone}.

Meta-heuristic algorithms have shown great capacity and adaptability in finding global minima in different test problems and practical applications over the years. However, these types of algorithms are always in constant improvement since according to the No Free Lunch (NFL) theorem, no meta-heuristic can optimize all kinds of problems \citep{wolpert1997no}. As a result of the need to optimize increasingly complex problems with a larger number of dimensions, the emergence of new meta-heuristic algorithms has increased in recent years \citep{Dokeroglua2019}.

In this work, a new meta-heuristic algorithm based on the neighborhood concept and evolution rules of cellular automata is proposed for the global optimization of problems defined in multiple dimensions. As this algorithm uses continuous variables, it has been named the continuous cellular automata algorithm (CCAA).

The algorithm defines a set of initial solutions, each one of which generates a neighborhood of possible new solutions to improve the fitness (or cost) of these solutions concerning the problem to be optimized, which, in the case of this work, is the minimization of functions. This neighborhood is formed using rules inspired by cellular automata, where some of them serve to exchange information with other solutions and others only take information from the solution and the best cost obtained to induce changes in its elements.

The contribution of the CCAA lies in the direct use of the concepts of neighborhood and evolution rule in the dynamics of the solutions to carry out the optimization process. These features lead to the implementation of an algorithm that is very simple to implement and which, at the same time, is highly competitive compared to other recently published algorithms with proven performance.

There are many meta-heuristics applications in the area of engineering since they have been used with satisfactory results in complex tasks such as the identification of system parameters from experimental data, the optimal modeling of materials, and multi-dimensional optimization, to mention a few \citep{Dwivedi2020}. Examples of these applications are mechanical gear train design, helical compression spring design \citep{sandgren1990nonlinear}, pressure vessel design \citep{sandgren1990nonlinear} \citep{salih2019}, process control \citep{8979108}, and the design of adaptive digital filters, one of the more popular application fields in recent years due to its relevance to real-world problems  \citep{jiang2015new} \citep{Lagos-Eulogio2017} \citep{Dwivedi2018}. The operation of the CCAA is tested using some of these applications to check its performance against other specialized algorithms.

The structure of the paper is as follows. Section \ref{sec:conceptos} describes the basic concepts of cellular automata, the proposals for meta-heuristic algorithms based on cellular automata, and the opportunity to improve this type of implementations. Section \ref{sec:ccaa} presents the CCAA algorithm and explains the adaptations of various evolution rules to carry out exploration and exploitation tasks concurrently as well as the general strategy of the CCAA. Section \ref{sec:experimentos} shows the experimental results of $33$ test functions in $30$ and $500$ dimensions, presenting a statistical comparison with other state-of-the-art algorithms. Section \ref{sec:aplicaciones} applies the CCAA in $4$ engineering problems that are commonly considered in the specialized literature, in addition to utilizing the proposed algorithm in the design of adaptive IIR filters of full order. These results are compared with previously published findings to show the effectiveness and performance of the CCAA. The last section provides the conclusions and future proposals for the development of the CCAA.
 
\section{Preliminaries}
\label{sec:conceptos}

A cellular automaton is a discrete dynamical system composed of cells that initially take their values from a finite set of possible states. The dynamics of the system is given in discrete steps. In each step, a cell takes into account its current state and those of its neighbors to update its state. This assignment of neighborhoods to states is known as the evolution rule. Thus, a cellular automaton is discrete in time and space.

\begin{figure}[h!]
    \centering
    \includegraphics[scale=0.33]{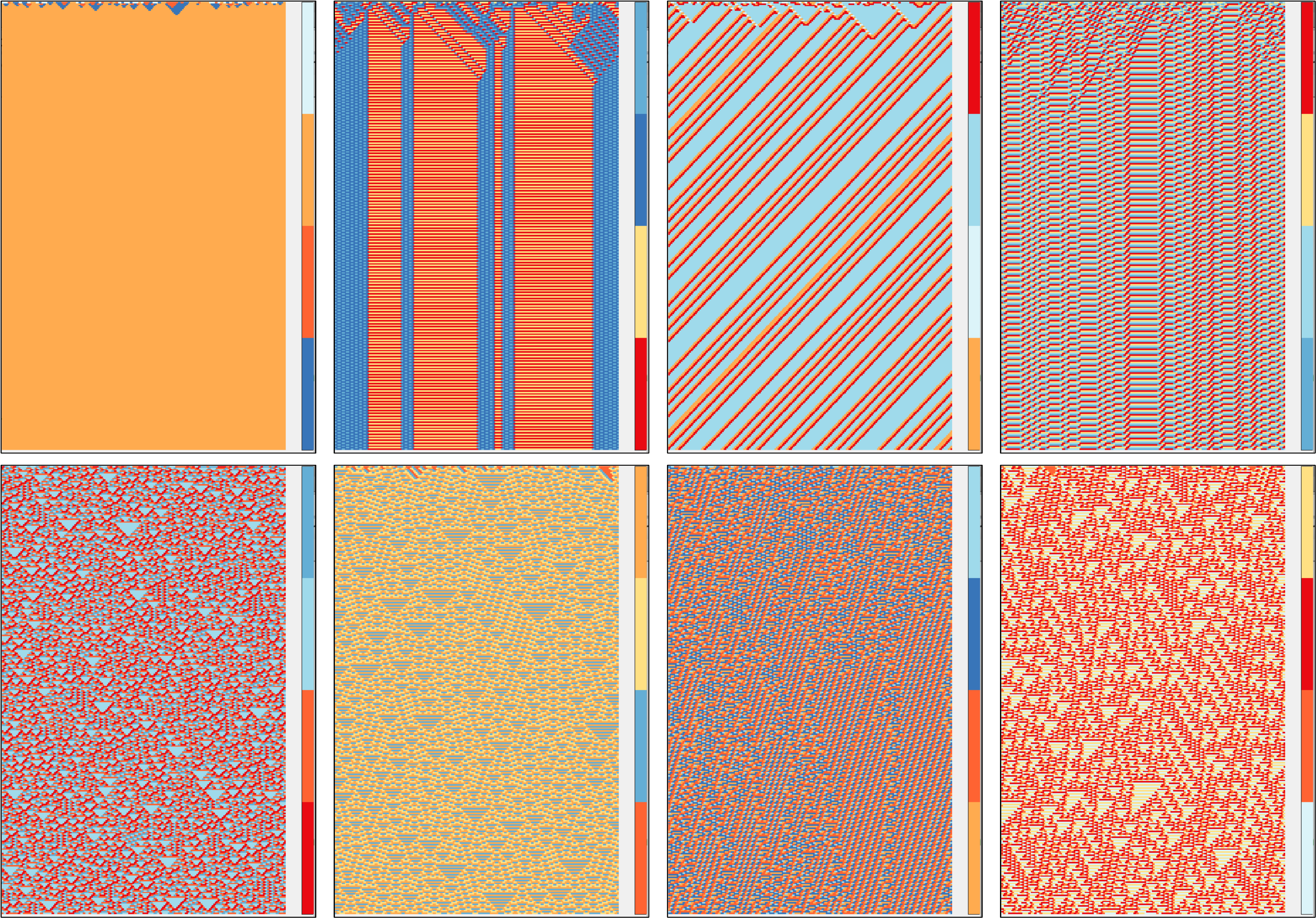}
    \caption{Examples of different cellular automata with $4$ states}
    \label{fig:Ejemplos_AC}
\end{figure}

Cellular automata are easy to implement in a computer, given the simplicity of their specification. However, these systems can develop chaotic and complex global behaviors, which have been widely investigated and used to solve various computational and engineering problems \citep{mcintosh2009one} \citep{wolfram2002new}. Fig. \ref{fig:Ejemplos_AC} shows the evolutions of different cellular automata with $4$ states, with distinct dynamical behaviors.

Although concepts and ideas of cellular automata have been used before for proposing new meta-heuristics, very few have used the main concept that characterizes a cellular automaton, the simplicity of its neighborhood, and the versatility that can be found in its evolution rules. One of the most recent works that uses the concept of neighborhood in cellular automata to propose a global optimization meta-heuristic is the cellular particle swarm optimization (CPSO) \citep{shi2011cellular}. CPSO has been shown to be very useful for global optimization tasks, although the proposed evolution rule continues to be based on an adaptation of the classical PSO equations. Other recent meta-heuristics were proposed for the implementation of cellular automata concepts oriented to particular engineering problems, such as for the design of adaptive IIR filters \citep{Lagos-Eulogio2017} and for the simultaneous optimization of the plant layout and the sequencing of tasks in a job-shop-type system \citep{hernandez2020solution}.

However, the versatility of the evolution rules that are widely used in cellular automata for various modeling problems and for computational complexity has not been directly adapted in an algorithm for global optimization tasks. This is the main contribution of this paper, whose implementation is described in the next section.

\section{Continuous cellular automata algorithm (CCAA)}
\label{sec:ccaa}

The evolution rules of cellular automata have been widely studied for the complex behaviors they represent, model, and construct. Within these rules, we can find that globally, cellular automata can generate periodic, chaotic, complex, reversible behaviors or are capable of solving complex global tasks such as global synchronization, classification, or task scheduling problems.

The aim is to take these types of rules as inspiration to adapt them into an optimization algorithm, where solutions are able to share information and generate spontaneous changes that are useful in carrying out exploration and exploitation tasks in a solution space.

The function that is to be optimized (in this work, the case of minimization is taken) is defined as $f(\mathbf{x}) \rightarrow \mathbb{R}$ in $n$ dimensions, where $x \in \mathbb{R}^{n}$ and each $x_i$ value in $\mathbf{x}$ is bounded between a lower bound $lb_i$ and an upper bond $ub_i$. If the bounds are the same for all the values of $\mathbf{x} $, then these limit values will simply be stated as $lb$ and $ub$.

The CCAA first takes an initial population $S$ of $smart\_n$ initial solutions or $smart\_cells$. Each $smart\_cell$ is represented as $s \in \mathbb{R}^n$ and its cost is given by $f(\mathbf{s})$. Given a population $S$, the best solution will be represented as $b_S$ such that its cost $f(b_S)$ is minimal with respect to all the other $smart\_cells$ in $S$. 

The algorithm uses a total of $12$ evolution rules adapted from cellular automata to apply them on the $smart\_cells$ in $S$ and their costs in $f$. The general strategy is to generate a neighborhood of each $smart\_cell$ with $m$ neighbors and from that neighborhood, take the best solution as the new $smart\_cell$ in a probabilistic way or if it improves the cost of the original $smart\_cell$. The evolution rules used to carry out the CCAA's exploration and exploitation actions are described in the upcoming section.


\subsection{Rules for approaching a neighbor}

In this type of rule, a $smart\_cell$ $s_i$ will approach another $smart\_cell$ $s_j$ depending on the cost $f(s_i)$ and $f(s_j)$. The structure of the rule is as follows:

\begin{algorithm}[H]
\SetAlgoLined
\KwResult{New $smart\_cell$ $evol$ }
 Input: $s_i$, $s_j$, $f(s_i)$, $f(s_j)$, $prop$ \;
 $evol=s_i$\;
 \If{$condition(f(s_i),f(s_j))$}
 {
  $dec=(s_i-s_j)*prop*rand$\;
 }
$evol=evol-dec$\;
\caption{Approach rule}
\label{alg:regla_acercamiento}
\end{algorithm}

The rule in the Algorithm \ref{alg:regla_acercamiento} is very simple: if $condition(f(s_i),f (s_j)) $ is true, a vector is calculated with the difference between each element of $s_i$ and $s_j$ and the rule weights these differences by a proportion that goes between $0$ and $prop$ in a uniform random way. If the condition is false, $s_i$ remains unchanged.

This makes $s_i$ get closer to $s_j$ as $prop$ indicates, if this value is low the approach will be limited. This rule serves the purposes of differentiated exploitation of the elements in $s_i$, since the vector of differences is calculated element by element. In the CCAA, one version of this rule,$R_1$, is used, which takes $f(s_i) \neq f(s_j)$ as  condition, and $prop = lower_p$, where $lower_p$ has a low value. This allows the original $s_i$ to be moved in its close neighborhood in the direction of a neighbor with a different cost.

\subsection{Rules for staying away from a neighbor}

\begin{algorithm}[H]
\SetAlgoLined
\KwResult{New $smart\_cell$ $evol$ }
 Input: $s_i$, $s_j$, $f(s_i)$, $f(s_j)$, $prop$ \;
 $evol=s_i$\;
 \If{$condition(f(s_i),f(s_j))$}
 {
  $inc=(s_i-s_j)*prop*rand$\;
 }
$evol=evol+inc$\;
\caption{Rule of taking away}
\label{alg:regla_alejamiento}
\end{algorithm}

The rule in Algorithm \ref{alg:regla_alejamiento} is also simple. If $condition(f(s_i), f(s_j))$ is true, the vector of differences $s_i-s_j$ is calculated and weighted with a uniform random value between $0$ and $prop$. If the condition is false, $s_i$ does not change.

This rule has the effect of moving $s_i$ away from $s_j$, which implies that if $s_i$ is close to $s_j$, then the rule can continue to exploit areas further away from the vicinity of $s_i$ and help to escape local minima. In the other case, where $s_i$ is far from $s_j$, then the new solution will be far from $s_j$ and will serve the exploration of new areas in the search space for both $s_i$ and $s_j$.

In the CCAA, two versions of this rule are used: $R_2$ with the condition $f(s_i) \neq f(s_j)$ and $prop = upper_b$ is considered, where $upper_b$ has a high value that allows for a significant difference from $s_j$, and $R_3$, which takes as condition $f(s_i) < f(s_j)$ and $prop = lower_b$, causing $s_i$ to deviate a little from $s_j$ only when the cost $f( s_j)$ is higher. This rule serves to intensify the task of exploiting the information in $s_i$ in a direction away from $s_j$.

\subsection{Rules for changes in a $smart\_cell$}

\begin{algorithm}[H]
\SetAlgoLined
\KwResult{New $smart\_cell$ $evol$ }
 Input: $s_i$, $s_j$, $f(s_i)$, $f(s_j)$, $dist$ \;
 $evol=s_i$\;
 $sum=f(s_i)+f(s_j)$\;
 $pond=1-(f(s_j)/sum)$\;
 $r=(rand*dist)-(dist/2)$\;
 \ForAll{$k$ in $length(evol)$}
 {
 \If{$rand<=pond$}
 {
  $evol_k=evol_k+r*s_{jk}$\;
 }
 }
\caption{Change rule}
\label{alg:regla_cambio}
\end{algorithm}

The rule in Algorithm \ref{alg:regla_cambio} is to estimate how optimal $s_i$ is compared to $s_j$ taking into account their costs. If $f (s_i)$ is very high compared to $f(s_j)$, then $pond$ will have a higher value; this will imply a greater probability that each element of $s_i$ will be modified. This probability of change will decrease as the value of $f(s_i)$ becomes lower, which is expected in the optimization process. The change consists of increasing the element $s_{ik} $ with some influence of $s_ {jk}$. This weight depends on the parameter $dist$ with $k=1 \ldots dim_n$, where $dim_n$ is the number of dimensions of the problem to be optimized.

The effect of this rule is to cause a change, element by element, in $s_i$, taking as a factor of change the neighbor $s_j$. Therefore, it is a rule focused on exploring new areas of the search space, mainly if the value of $f(s_i)$ is very high compared to $f(s_j)$.

In the CCAA, two versions of this rule are used: $R_4$, where the parameter $dist = dist\_M$, with $dist\_M$ being a value that allows extensive changes, and $R_5$, with $dist = dist\_m$ and $dist\_m$ defined as a smaller value that allows moderate changes.

\subsection{Rules for increasing values in a $smart\_cell$}

\begin{algorithm}[H]
\SetAlgoLined
\KwResult{New $smart\_cell$ $evol$ }
 Input: $s_i$, $f(s_i)$, $best_f$, $dist$ \;
 $evol=s_i$\;
 $sum=f(s_i)+best_f$\;
 $pond=1-(f(s_i)/sum)$\;
 $r=(rand*dist)-(dist/2)$\;
 \ForAll{$k$ in $length(evol)$}
 {
 \If{$rand<=pond$}
 {
  $evol_k=evol_k+r*evol_k$\;
 }
 }
\caption{Increment rule}
\label{alg:regla_incremento}
\end{algorithm}

The rule in Algorithm \ref{alg:regla_incremento} is very similar to the change rule, but in its definition, it only takes into account the best cost obtained in the population ($best_f$) and an increment parameter $dist$. The weighting is done taking into account $f(s_i)$ and $best_f$. If the value $f(s_i)$ is larger, then $pond$ will be small, inducing little change in $s_i$. If $f(s_i)$ is close to $best_f$, the probability of making changes to the elements of $s_i$ will be increased. The applied changes will also be in proportion to the same values of $s_i$ weighted by the $dist$ parameter.

This rule's effect is to induce self-generated changes by the values $s_i$, so it is a rule focused on exploiting the information of $s_i$, especially when it has a better cost.

In the CCAA, two versions of this rule are used: $R_6$, which applies a parameter $dist = dist_M$ to induce larger increments on $s_i$, and $R_7$, which uses $dist = dist_m $ to only cause small modifications to the $smart\_cell$.

\subsection{Rules for majority values in a $smart\_cell$}

\begin{algorithm}[H]
\SetAlgoLined
\KwResult{New $smart\_cell$ $evol$ }
 Input: $s_i$, $dist$\;
 $elem =$ most repeated element in $s_i$\;
 $cam = (s_i-elem)*dist*rand$\;
 $evol = s_i - cam$\;
\caption{Majority rule}
\label{alg:regla_mayoria}
\end{algorithm}

The rule in Algorithm \ref{alg:regla_mayoria} is very simple. It takes the element $elem$ that is most repeated in $ s_i $ and forms a vector $cam$ of differences between $s_i$ and $elem$ weighted by a parameter between $0$ and $dist$. The vector $cam$ is subtracted from $s_i$ to form a new solution in order to bring the values of $s_i$ closer to the most repeated value.

This rule's effect is to autogenerate a change in $s_i$ that makes its values more homogeneous, which is suitable for many optimization problems where vectors with good costs have elements with the same value.

In the CCAA, two versions of this rule are used: $R_{8} $, which takes the most repeated element of $s_i$ and the analogous $R_{9} $, which can be considered as a minority rule as it chooses the least repeated element of $s_i$.

\subsection{Rules for rounding values in a $smart\_cell$}

\begin{algorithm}[H]
\SetAlgoLined
\KwResult{New $smart\_cell$ $evol$ }
 Input: $s_i$, $f(s_i)$, $best_f$, $num_d$\;
 $evol=s_i$\;
 $sum=f(s_i)+best_f$\;
 $pond=1-(f(s_i)/sum)$\;
 \ForAll{$k$ in $length(evol)$}
 {
 \If{$rand<=pond$}
 {
  $evol_k=round(evol_k,num_d)$\;
 }
 }
\caption{Rounding rule}
\label{alg:regla_redondeo}
\end{algorithm}

The rule in Algorithm \ref{alg:regla_redondeo} takes the weight of $f(s_i)$ in contrast to $best_f$. If $f(s_i)$ is large, then $pond$ is small, and few changes are made; otherwise, there is a greater probability that changes will occur. Each change consists of rounding selected elements of $s_i$ to discretize the decimal part to as many digits as indicated by $num_d$.

This rule's effect is to autogenerate a change in $s_i$ that rounds its values, which is useful for optimization problems being primarily focused on finding values for a system's parameter specifications. In the CCAA, a single version $R_ {10}$ of this rule is used. Given the previous rules, Algorithm \ref{alg:CCAA} presents the general structure of the CCAA.

\subsection{Complete structure of the CCAA}

\begin{algorithm}[H]
\SetAlgoLined
\KwResult{Best $smart\_cell$ $b_S$ and fitness value $f(b_S)$}
 Input: $smart\_n$, $neighbor\_n$, $iteration\_n$, $elitism\_n$, $lower_b$, $upper_b$, $dim_n$, $f()$\;
 Set parameters $lower_p$, $upper_p$, $dist\_M$, $dist\_m$, $lower_d$ y $upper_d$\;
 Generate random population $S$ of $smart\_n$ $smart\_cells$\;
 Evaluate $S$ in $f()$\;
 \ForAll{$i=2$ to $iteration\_n$}
 {
 Keep the best $elitism\_n$ $smart\_cells$ in a new population\;
 \ForAll{$j=elitism\_n+1$ a $smart\_n$}
 {
 Take $smart\_cell_j$ and another random $neighbor$ from $S$ for rules requiring an extra $smart\_cell$\;
 \ForAll{$k=1$ to $neighbor\_n$}
 {
 Choose a rule $R$ in a random manner\;
 Obtain $evolution=R(smart\_cell_j,\mbox{additional parameters of the rule})$ \; 
 Check that the $dim$ values of $evolution$ are between $lower\_b$ and $upper\_b$ and correct if necessary\;
 Calculate $cost_k=f(evolution)$\;
 }
 Chose the best neighbor $new\_neigh$ from the $k$ generated neighbors, taking $new\_cost=f(best\_neigh)$\;
 \If{$rand < 0.5$ o $f(smart\_cell_j) > new\_cost$}
 {
  $smart\_cell_j=new\_neigh$\;
 }
 }
 }
 Return the best $smart\_cell$ $b_S$ in $S$ and its fitness value $f(b_S)$\;
\caption{Continuous-state cellular automata algorithm (CCAA)}
\label{alg:CCAA}
\end{algorithm}

The CCAA receives $4$ input values to operate, which are the number $smart_n$ of $smart\_cells$, the number $neighbor_n $ of neighbors of each $smart\_cell $, the number of iterations $ iteration_n $ and the number of elitist solutions $ elitism_n $. The other input values correspond to properties of the function that is to be optimized, such as the limits $ lower_b $ and $ upper_b $ for every element of each $ smart\_cell $ and the number of dimensions or elements $ dim_n $  of each $ smart\_cell $.

The CCAA has $ 6 $ parameters, $ lower_p $ and $ upper_p $ to define the $ prop $ parameter in the rules based on Algorithms \ref{alg:regla_acercamiento} and \ref{alg:regla_alejamiento}; the $ dist_M $ and $ dist_m $ parameters to define the $ dist $ parameter for the rules arising from Algorithms \ref{alg:regla_cambio}, \ref{alg:regla_incremento}, and \ref{alg:regla_mayoria}; and finally, the parameters $ lower_d $ and $ upper_d $ to generate a random number between these two values that defines the parameter $ num_d $ for the rule specified by Algorithm \ref{alg:regla_redondeo}.

\begin{figure}[htbp]
\begin{center}
\includegraphics[scale=0.46]{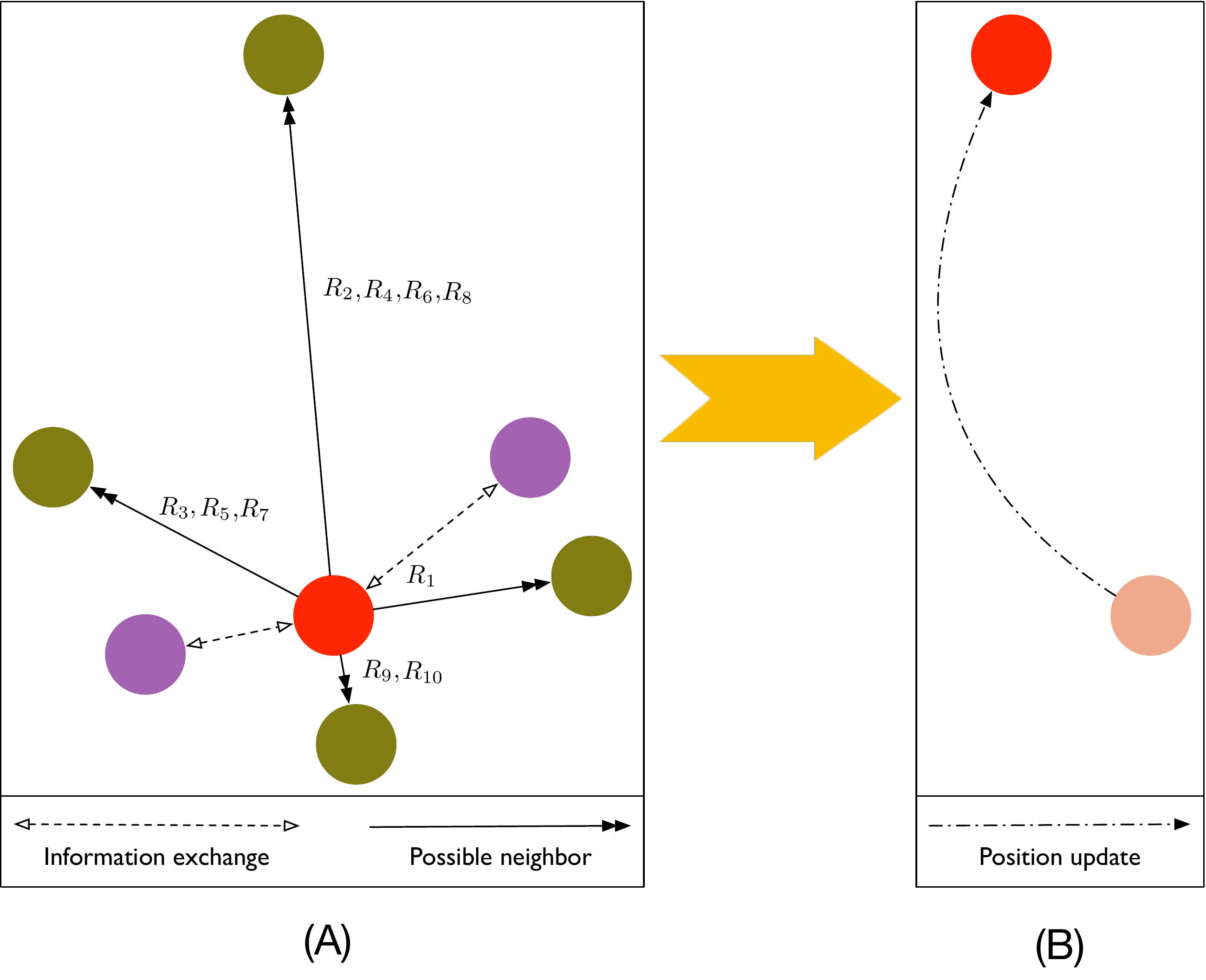}
\caption{Description of a neighborhood of each $smart\_cell $ in the CCAA}
\label{fig:Vecindad_CCAA}
\end{center}
\end{figure}

The CCAA's strategy consists of generating a random population of $smart\_cells$, qualifying it, and taking some elitist solutions to update the population. The rest of the solutions are updated one by one using a neighborhood of possible new $smart\_cells$, taking the available rules at random. A graphical description of how rules can update the position of a $smart\_cell $ is shown in Fig. \ref {fig:Vecindad_CCAA}. In part $ (A) $, the rules are applied randomly to generate new $smart\_cells$ that are possibly near or far from the original $smart\_cell$. The neighbors are produced either by exchanging information with other $smart\_cells$, taking as a reference the best fitness value obtained, or by taking the information of the $smart\_cell$.

\newpage

\begin{figure}[h!]
\begin{center}
\includegraphics[scale=0.75]{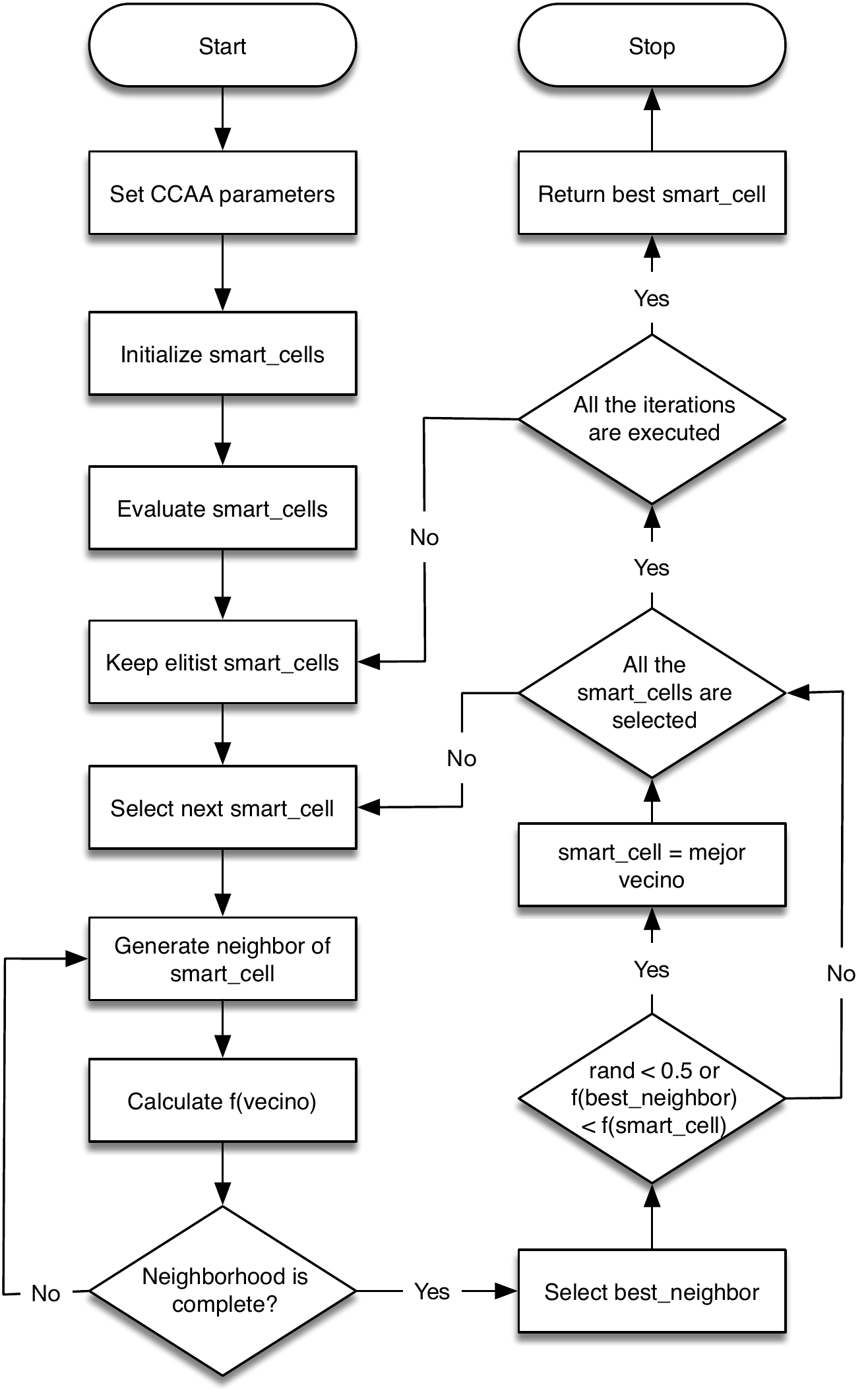}
\caption{CCAA flow chart}
\label{fig:DF_CCAA}
\end{center}
\end{figure}

Once the neighborhood has been generated, the best position with minimum cost is selected (Fig. \ref{fig:Vecindad_CCAA} $ (B) $)  to upgrade the $smart\_cell $ if it improves its cost, or otherwise with a probability of $ 50 \% $. The elitism of the CCAA serves to preserve the information of the best solutions generated by the optimization process. The probability of substituting a $smart\_cell $ for another with a worse value also helps the algorithm avoid stagnation at local minima. Figure \ref{fig:DF_CCAA} shows the CCAA flow chart.

\section{Experimental results}
\label{sec:experimentos}

To validate the effectiveness of the CCAA, $ 2 $ groups of experiments with different types of functions were developed.

The first group evaluated the CCAA with $23$ scalable proof problems in $30$ dimensions and $10$ problems with fixed dimensions. The results are compared with $7$ other recently published meta-heuristics, namely, the CPSO \citep{shi2011cellular}, the gray wolf optimizer (GWO) \citep{mirjalili2014grey}, the whale optimization algorithm (WOA) \citep{mirjalili2016whale}, the mayfly optimization algorithm (MA) \citep{zervoudakis2020mayfly}, the adaptive logarithmic spiral-Levy firefly algorithm (ADIFA) \citep{wu2020improved}, the double adaptive random spare reinforced whale optimization algorithm \citep{chen2020efficient}, and the modified sine cosine algorithm (MSCA) \citep{gupta2020modified}. The test problems used have been widely utilized to validate the performances of various algorithms \citep{gupta2020modified} \citep{zervoudakis2020mayfly} \citep{chen2020efficient} \citep{wu2020improved}. Most of the codes for these algorithms were taken from the sites noted in the references, so these are mostly implementations made by the same authors of the indicated papers, which provide a more objective comparison between algorithms. The only algorithm implemented was the MSCA, for which the original SCA code was taken and modified following the instructions published in \citep{gupta2020modified}.

The definitions of the test problems are presented in Tables \ref{tabla:funciones_unimodales}, \ref{tabla:funciones_multimodales}, and \ref{tabla:funciones_fijas}. Comparison with the other algorithms shows the average value and the standard deviation of their values obtained in the objective functions, taking $30$ independent runs of each algorithm. The second set of experiments contemplates the same $23$ scalable test problems but in $ 500 $ dimensions, making the same statistical comparisons of the average and standard deviations in another $30$ independent runs for each algorithm. A comparative statistical analysis was performed in each set of experiments applying the Wilcoxon rank-sum test, also showing some convergence and box diagrams for selected functions.

A preliminary study was conducted that considered different values of the CCAA parameters and they were applied to the problems $ F_6 $, $ F_ {11} $, and $ F_ {24} $ in $30$ dimensions to select the best parameters. For the population size and the number of generations, $ smart\_n = 12 $, $ neighbor\_n = 6 $, and $ iteration\_n = 500 $ were taken to have parameters similar to the algorithms used for comparison.

The other CCAA parameters were tested on two levels to choose the best combination. For $lower_p $, $ 0.5 $ and $ 1 $ were taken; for $ upper_p $, $ 2 $ and $ 3 $ were tested. For $ dist\_M $, $ 1 $ and $ 2 $ were analyzed; for $ dist\_m $, the values $ 0.15 $ and $ 0.3 $ were verified. For $ lower_d $, $ 0.5 $ and $ 1 $ were taken; for $ upper_d $, $ 3 $ and $ 4 $ were tested; and for $ elitism\_n $, the values $ 1 $ and $ 2 $ were studied; having a total of $128$ combinations. For each combination, $30$ independent runs of the CCAA were made in the three functions selected, and the combination of parameters that obtained the best average value was chosen. As a result of this analysis, the parameters selected to conduct the comparison of the CCAA with the other algorithms in the following experiments are $lower_p = 1$, $ upper_p = 2 $, $ dist\_M = 1 $, $ dist\_m = 0.3 $, $ lower_d = 1 $, $ upper_d = 4 $, and $ elitism\_n = 2 $.

The CCAA was implemented in Matlab 2015a, on a machine with Intel Xeon CPU at $ 3.1 $ GHz, $ 64 $ GB in the RAM, and $ 1 $ TB in the hard disk, running on Mac OS X Catalina operating system.

\subsection{Experiment 1: Functions on $30$ dimensions and functions with fixed dimensions}

This section compares the performance of the CCAA against 6 other recently published algorithms for global optimization, using $ 23 $ scalable test functions that were tested in this experiment with $ 30 $ dimensions and are described in Tables \ref{tabla:funciones_unimodales} and \ref{tabla:funciones_multimodales}. The comparison with the other $ 10 $ functions of fixed dimensions is also shown, which are defined in Table \ref{tabla:funciones_fijas}.

\begin{table}[h!]
\centering
\small
\caption{\label{tabla:funciones_unimodales}Description of scalable-dimensional unimodal benchmark functions} 
\begin{tabular}{llll}
\hline
Function & Dimensions & Range & $f_{min}$\\ 
\hline
$F1(x)=\sum_{i=1}^{n} x_i^2$ & $30$, $500$ & $[-100,100]$ & $0$  \\ 
$F2(x)=\sum_{i=1}^{n} ix_i^2$ & $30$, $500$ & $[-10,10]$ & $0$  \\ 
$F3(x)=\sum_{i=1}^{n} |x_i| + \prod_{i=1}^{n} |x_i| $ & $30$, $500$ & $[-10,10]$ & $0$  \\ 
$F4(x)=\sum_{i=1}^{n} \left( \sum_{j=1}^{i} x_j \right)^2 $ & $30$, $500$ & $[-100,100]$ & $0$  \\ 
$F5(x)=\max_i \left\{ |x_i|, 1 \leq i \leq n \right\} $ & $30$, $500$ & $[-100,100]$ & $0$  \\ 
$F6(x)=\sum_{i=1}^{n-1} \left( 100 \left( x_{i+1} - x_i^2 \right)^2 + (x_i-1)^2 \right) $ & $30$, $500$ & $[-30,30]$ & $0$  \\ 
$F7(x)=\sum_{i=1}^{n} \left( x_i + 0.5 \right)^2 $ & $30$, $500$ & $[-100,100]$ & $0$  \\ 
$F8(x)=\sum_{i=1}^{n} ix_i^4$ & $30$, $500$ & $[-1.28,1.28]$ & $0$  \\ 
$F9(x)=\sum_{i=1}^{n} ix_i^4 + random[0,1)$ & $30$, $500$ & $[-1.28,1.28]$ & $0$  \\ 
$F10(x)=\sum_{i=1}^{n} |x_i|^{i+1}$ & $30$, $500$ & $[-1,1]$ & $0$  \\ 
\hline
\end{tabular}
\end{table}

The first $ 10 $ test functions are unimodal and are useful in the evaluation of the exploitation properties of the $ smart\_cells $ in the CCAA. The following $ 13 $ test functions are multimodal and used to analyze the CCAA's ability to explore and escape local minima. Lastly, the test functions with fixed dimensionality have a lower number of local minima and are useful in the observation of the balance between the exploration and exploitation actions of the CCAA.

In this experiment, $30$ independent runs were made per algorithm; for the CPSO and the CCAA, $smart\_n = 12 $ and $ neighbor\_n = 6 $ were used; for the rest of the algorithms, $(smart\_n)(neighbor\_n) = 72 $ individuals were used. In all algorithms, $iteration\_n = 500 $ was employed. Table \ref{tabla:resultados30_funciones_unimodales} presents the results of the comparison with respect to the average and standard deviation for the unimodal functions, Table \ref{tabla:resultados30_funciones_multimodales} shows the same values for multimodal functions, and Table \ref{tabla:resultados30_funciones_fijas} presents the results for functions on fixed dimensions.

\begin{table}[h!]
\centering
\scriptsize
\caption{\label{tabla:funciones_multimodales}Description of scalable-dimensional multimodal benchmark functions} 
\begin{tabular}{llll}
\hline
Function & Dimensions & Range & $f_{min}$\\ 
\hline
$F11(x)=\sum_{i=1}^{n} -x_i sin\left( \sqrt{ | x_i | } \right)$ & $30$, $500$ & $[-500,500]$ & $-418.9829 \times n$  \\ 
$F12(x)=\sum_{i=1}^{n} \left( x_i^2 - 10 cos(2 \pi x_i) + 10 \right)$ & $30$, $500$ & $[-5.12, 5.12]$ & $0$  \\ 
$F13(x)= -20 \exp \left( -0.2 \sqrt{ \frac{1}{n} \sum_{i=1}^{n} x_i^2 } \right)$  & $30$, $500$ & $[-32, 32]$ & $0$  \\ 
$-\exp \left( \frac{1}{n} \sum_{i=1}^{n} cos(2 \pi x_i)  \right) + 20 + e$ & & \\

$F14(x)= \frac{1}{4000} \sum_{i=1}^{n} x_i^2  -\prod_{i=1}^{n} cos \left( \frac{x_i}{\sqrt{i}} \right) + 1 $ & $30$, $500$ & $[-600,600]$ & $0$  \\

$F15(x)= \frac{\pi}{n} \left( 10 sin (\pi y_1) + \sum_{i=1}^{n-1} (y_i-1)^2 \left( 1+10 sin^2 (\pi y_{i+1}) \right) \right. $ & $30$, $500$ & $[-50,50]$ & $0$  \\ 
$\left. + (y_n -1)^2\right) + \sum_{i=1}^{n} u(x_i,10,100,4); \; y_i=1+\frac{x_i+1}{4};$ \\ $u(x_i,a,k,m)= \left\{ \begin{array}{l}  k(x_i-a)^m, \; x_i > a \\ -a, \; -a \leq x_i \leq a \\ k(-x_i-a)^m, \; x_i < -a \\ \end{array} \right. $ & & & \\

$F16(x)= 0.1 \left[ sin^2(3 \pi x_1) + \sum_{i=1}^{n-1} (x_i-1)^2  \left( 1+sin^2 (3 \pi x_i +1) \right) \right.+    $ & $30$, $500$ & $[-50,50]$ & $0$  \\ 
$\left. (x_n-1)^2 \left( 1+sin^2 (2 \pi x_n) \right) \right] + \sum_{i=1}^{n} u(x_i,5,100,4)$ & & &\\
$F17(x)=\sum_{i=1}^{n} | x_i sin(x_i) + 0.1x_i | $ & $30$, $500$ & $[-10,10]$ & $0$  \\ 
$F18(x)=\sum_{i=1}^{n} 0.1n-\left( 0.1 \sum_{i=1}^{n} cos(5 \pi x_i) - \sum_{i=1}^{n} x_i^2 \right)$ & $30$, $500$ & $[-1, 1]$ & $0$  \\ 
$F19(x)=\sum_{i=1}^{n} \left( x_i^2 + 2x_{i+1}^2 \right)^{0.25} \times \left( 1+sin(50(x_i^2 + x_{i+1}^2)^{0.1}  \right)^2$ & $30$, $500$ & $[-1.28,1.28]$ & $0$  \\ 
$F20(x)=\sum_{i=1}^{n} (10^6)^{(i-1)/(n-1)}x_i^2$ & $30$, $500$ & $[-100,100]$ & $0$  \\ 
$F21(x)=(-1)^{n+1} \prod_{i=1}^{n} cos(x_i) \times \exp \left( -\sum_{i=1}^n (x_i - \pi)^2 \right)$ & $30$, $500$ & $[-100,100]$ & $-1$  \\ 
$F22(x)=1- cos \left( 2\pi \sqrt{ \left( \sum_{i=1}^n x_i^2 \right)} \right) + 0.1 \sqrt{\left( \sum_{i=1}^n x_i^2 \right)}$ & $30$, $500$ & $[-100,100]$ & $0$  \\ 
$F23(x)=0.5 + \left( sin^2 \left( \sqrt{\left( \sum_{i=1}^n x_i^2 \right)}  \right) -0.5 \right) / \left( 1+0.001 \left( \sum_{i=1}^n x_i^2 \right)^2 \right)$ & $30$, $500$ & $[-100,100]$ & $0$  \\ 
\hline
\end{tabular}
\end{table}

\begin{table}[h!]
\centering
\scriptsize
\caption{\label{tabla:funciones_fijas}Description of fixed-dimension multimodal benchmark functions} 
\begin{tabular}{llll}
\hline
Function & Dimensions & Range & $f_{min}$\\ 
\hline
$F24(x)= \left( \frac{1}{500} + \sum_{j=1}^{25} \frac{1}{j+\sum_{i=1}^2 (x_i-a_{ij})^6} \right)^{-1}$ & $2$ & $[-65, 65]$ & $1$  \\ 
$F25(x)=\sum_{i=1}^{11} \left( a_i - \frac{x_1(b_i^2+b_ix_2)}{b_i^2+b_ix_3+x_4} \right)^2$ & $4$ & $[-5, 5]$ & $0.00030$  \\ 
$F26(x)= 4x_1^2 - 2.1x_1^4 + (1/3)x_1^6 + x_1x_2 - 4x_2^2 + 4x_2^4$ & $2$ & $[-5, 5]$ & $-1.0316$  \\ 
$F27(x)= \left( x_2 - (5.1/4\pi^2)x_1^2 + (5/\pi)x_1 -6 \right)^2 + 10(1-(1/8\pi))cos(x_1) + 10 $ & $2$ & $[-5, 5]$ & $0.398$  \\ 
$F28(x)= \left( 1+ (x_1+x_2+1)^2 (19-14x_1+3x_1^2 - 14x_2 + 6x_1x_2 + 3x_2^2) \right) $ & $2$ & $[-2,2]$ & $3$  \\ 
$\times \left( 30+(2x_1-3x_2)^2 \times (18 - 32x_1 + 12x_1^2 + 48x_2 -36x_1x_2 +27x_2^2)  \right)$ & & & \\
$F29(x)=-\sum_{i=1}^{4} c_i \exp \left( -\sum_{j=1}^3 a_{ij}(x_j - p_{ij})^2 \right) $ & $3$ & $[1,3]$ & $-3.86$  \\ 
$F30(x)=-\sum_{i=1}^{4} c_i \exp \left( -\sum_{j=1}^6 a_{ij}(x_j - p_{ij})^2 \right) $ & $6$ & $[0,1]$ & $-3.32$  \\ 
$F31(x)=-\sum_{i=1}^{5} \left( (X-a_i)(X-a_i)^T + c_i \right)^{-1}$ & $4$ & $[0,10]$ & $-10.1532$  \\ 
$F32(x)=-\sum_{i=1}^{7} \left( (X-a_i)(X-a_i)^T + c_i \right)^{-1}$ & $4$ & $[0,10]$ & $-10.4028$  \\ 
$F33(x)=-\sum_{i=1}^{10} \left( (X-a_i)(X-a_i)^T + c_i \right)^{-1}$ & $4$ & $[0,10]$ & $-10.5363$  \\ 
\hline
\end{tabular}
\end{table}

For unimodal problems, the CCAA obtained $ 9 $ of the $ 10 $ best results with respect to the average value; i.e., the CCAA was able to achieve the optimal solutions for the problems $ F1 $ to $ F5 $, $ F7 $, $ F8 $, and $ F10 $ . Furthermore, in $ 8 $ of $ 10 $ cases, the CCAA produced the best values concerning the standard deviation, which gives evidence of the proposed algorithm's information exploitation capacity.

For the $ 13 $ multimodal problems, the CCAA also obtained $ 9 $ of the $ 13 $ best average values,($ F11 $, $ F13 $ to $ F16 $ and from $ F18 $ to $ F21 $). In $ 5 $ of them, the optimal values were calculated ($ F11 $, $ F14 $, and from $ F18 $ to $ F20 $). The performance of the CCAA is comparable to the MSCA, which produced $ 8$ better average values, where $ 7$ of them were optimal values. In $ 8 $ out of $ 13 $ cases, the CCAA achieved the best values referring to the standard deviation, showing the CCAA's exploration ability.

For the $ 10 $ fixed-dimension problems, the CCAA produced $ 7 $ of the best average values ($ F24 $, $ F27 $ and from $ F29 $ to $ F33 $). Optimal value was reached in $ 6 $ of them ($ F27 $ and from $ F29 $ to $ F33 $). The performance of the CCAA is comparable to the MA, which only got $ 5$ best average values, where $ 4$ of them are optimal values. In $ 5$ out of $ 10$ cases, the CCAA calculated the best values for the standard deviation, showing the robustness of the CCAA in carrying out exploration and exploitation actions simultaneously.

\newgeometry{margin=1.5cm}
\begin{landscape}

\begin{table}[th]
\tiny
\caption{\label{tabla:resultados30_funciones_unimodales}Performance of meta-heuristic algorithms compared with the CCAA on 30-dimensional unimodal problems} 
\begin{tabular}{lllllllllllllllll}
\hline
Function & \multicolumn{2}{l}{CPSO}& \multicolumn{2}{l}{GWO}& \multicolumn{2}{l}{WOA}& \multicolumn{2}{l}{MA}& \multicolumn{2}{l}{ADIFA}& \multicolumn{2}{l}{RDWOA}& \multicolumn{2}{l}{MSCA}& \multicolumn{2}{l}{CCAA}\\ 
\hline
& mean & std& mean & std& mean & std& mean & std& mean & std& mean & std& mean & std& mean & std \\ 
F1& 8.76e-07 & 1.32e-06 & 1.98e-37 & 2.33e-37 & 1.97e-89 & 1.39e-88 & 5.19e-07 & 8.70e-07 & 1.35e+03 & 4.09e+02 & 1.06e-141 & 4.37e-141 & \bf{0.00e+00} & \bf{0.00e+00} & \bf{0.00e+00} & \bf{0.00e+00}  \\ 
F2& 4.00e+00 & 2.83e+01 & 3.45e-38 & 5.41e-38 & 1.86e-91 & 9.12e-91 & 9.35e-08 & 1.33e-07 & 6.26e+01 & 3.22e+01 & 1.42e-141 & 5.92e-141 & \bf{0.00e+00} & \bf{0.00e+00} & \bf{0.00e+00} & \bf{0.00e+00}  \\ 
F3& 7.79e+00 & 1.17e+01 & 4.83e-22 & 3.40e-22 & 1.02e-55 & 4.97e-55 & 8.36e-05 & 3.79e-04 & 9.05e+00 & 3.37e+00 & 5.26e-88 & 1.75e-87 & \bf{0.00e+00} & \bf{0.00e+00} & \bf{0.00e+00} & \bf{0.00e+00}  \\ 
F4& 5.60e+03 & 4.23e+03 & 4.14e-09 & 1.92e-08 & 1.92e+04 & 7.87e+03 & 5.96e+03 & 2.19e+03 & 1.89e+03 & 7.15e+02 & 3.56e-64 & 1.54e-64 & \bf{0.00e+00} & \bf{0.00e+00} & \bf{0.00e+00} & \bf{0.00e+00}  \\ 
F5& 9.96e+00 & 5.31e+00 & 1.90e-09 & 2.87e-09 & 2.74e+01 & 3.08e+01 & 4.59e+01 & 8.64e+00 & 2.75e+01 & 6.59e+00 & 2.19e-39 & 3.98e-39 & \bf{0.00e+00} & \bf{0.00e+00} & \bf{0.00e+00} & \bf{0.00e+00}  \\ 
F6& 1.84e+03 & 1.27e+04 & 2.67e+01 & 8.22e-01 & 2.71e+01 & \bf{2.51e-01} & 6.88e+01 & 4.78e+01 & 9.07e+04 & 6.52e+04 & 2.34e+01 & 1.94e+00 & 9.61e+00 & 1.29e+01 & \bf{1.03e+00} & 5.12e+00  \\ 
F7& 2.01e-06 & 7.56e-06 & 3.11e-01 & 2.82e-01 & 1.85e-02 & 2.62e-02 & 3.94e-07 & 5.11e-07 & 1.34e+03 & 4.58e+02 & 2.42e-01 & 2.37e-01 & 5.89e-08 & 1.02e-07 & \bf{0.00e+00} & \bf{0.00e+00}  \\ 
F8& 6.20e-17 & 2.27e-16 & 4.48e-67 & 1.90e-66 & 2.54e-156 & 1.55e-155 & 7.12e-15 & 3.01e-14 & 2.81e-03 & 2.58e-03 & 5.28e-241 & \bf{0.00e+00} & \bf{0.00e+00} & \bf{0.00e+00} & \bf{0.00e+00} & \bf{0.00e+00}  \\ 
F9& 7.46e-02 & 8.96e-02 & 7.80e-04 & 6.06e-04 & 1.65e-03 & 1.85e-03 & 1.80e-02 & 5.81e-03 & 3.78e-01 & 1.80e-01 & 5.64e-05 & 5.07e-05 & \bf{3.91e-05} & \bf{4.02e-05} & 2.36e-04 & 1.94e-04  \\ 
F10& 3.80e-01 & 5.67e-01 & 3.31e-23 & 2.07e-23 & 4.79e-56 & 3.28e-55 & 4.71e-06 & 1.02e-05 & 7.27e-01 & 2.43e-01 & 5.94e-89 & 1.68e-88 & \bf{0.00e+00} & \bf{0.00e+00} & \bf{0.00e+00} & \bf{0.00e+00}  \\ 
\hline
\end{tabular}
\end{table}

\begin{table}[th]
\tiny
\caption{\label{tabla:resultados30_funciones_multimodales}Performance of meta-heuristic algorithms compared with the CCAA on 30-dimensional multimodal problems} 
\begin{tabular}{lllllllllllllllll}
\hline
Function & \multicolumn{2}{l}{CPSO}& \multicolumn{2}{l}{GWO}& \multicolumn{2}{l}{WOA}& \multicolumn{2}{l}{MA}& \multicolumn{2}{l}{ADIFA}& \multicolumn{2}{l}{RDWOA}& \multicolumn{2}{l}{MSCA}& \multicolumn{2}{l}{CCAA}\\ 
\hline
& mean & std& mean & std& mean & std& mean & std& mean & std& mean & std& mean & std& mean & std \\ 
F11& -9.97e+03 & 6.28e+02 & -6.41e+03 & 7.13e+02 & -1.14e+04 & 1.36e+03 & -1.00e+04 & 4.26e+02 & -6.68e+03 & 8.05e+02 & -8.88e+03 & 6.77e+02 & -1.26e+04 & 6.70e-03 & \bf{-1.26e+04} & \bf{7.35e-12}  \\ 
F12& 1.63e+02 & 4.58e+01 & 1.58e+00 & 3.07e+00 & \bf{0.00e+00} & \bf{0.00e+00} & 2.58e+01 & 1.03e+01 & 5.34e+01 & 1.93e+01 & \bf{0.00e+00} & \bf{0.00e+00} & \bf{0.00e+00} & \bf{0.00e+00} & 1.19e+00 & 5.91e+00  \\ 
F13& 1.82e+01 & 5.19e+00 & 3.31e-14 & 4.21e-15 & 4.58e-15 & 2.87e-15 & 1.57e+00 & 5.35e-01 & 8.40e+00 & 1.66e+00 & 4.16e-15 & 9.74e-16 & \bf{8.88e-16} & \bf{0.00e+00} & \bf{8.88e-16} & \bf{0.00e+00}  \\ 
F14& 1.27e-02 & 1.97e-02 & 1.79e-03 & 5.10e-03 & 5.66e-03 & 1.76e-02 & 1.79e-02 & 2.21e-02 & 1.59e+01 & 4.34e+00 & 8.87e-04 & 3.14e-03 & \bf{0.00e+00} & \bf{0.00e+00} & \bf{0.00e+00} & \bf{0.00e+00}  \\ 
F15& 2.90e-02 & 1.26e-01 & 2.12e-02 & 1.31e-02 & 3.68e-03 & 6.15e-03 & 4.15e-01 & 6.18e-01 & 1.42e+01 & 8.97e+00 & 6.86e-03 & 6.06e-03 & 8.96e-09 & 2.19e-08 & \bf{1.57e-32} & \bf{5.53e-48}  \\ 
F16& 3.32e-01 & 1.28e+00 & 2.51e-01 & 1.54e-01 & 8.23e-02 & 7.16e-02 & 2.25e-01 & 3.88e-01 & 5.17e+03 & 1.09e+04 & 2.51e-01 & 1.50e-01 & 5.93e-02 & 4.19e-01 & \bf{1.35e-32} & \bf{1.11e-47}  \\ 
F17& 3.11e-03 & 9.76e-03 & 2.85e-04 & 5.02e-04 & \bf{2.77e-41} & \bf{1.96e-40} & 1.04e-05 & 2.64e-05 & 3.12e+00 & 1.60e+00 & 1.97e-12 & 9.32e-12 & 9.78e-08 & 6.92e-07 & 1.25e-11 & 5.52e-11  \\ 
F18& 1.02e+00 & 3.68e-01 & \bf{0.00e+00} & \bf{0.00e+00} & \bf{0.00e+00} & \bf{0.00e+00} & 9.46e-02 & 1.15e-01 & 9.64e-01 & 3.33e-01 & \bf{0.00e+00} & \bf{0.00e+00} & \bf{0.00e+00} & \bf{0.00e+00} & \bf{0.00e+00} & \bf{0.00e+00}  \\ 
F19& 2.37e+01 & 1.31e+01 & 1.15e-10 & 4.57e-11 & 1.53e-31 & 5.45e-31 & 2.52e-02 & 1.64e-02 & 4.07e+01 & 7.30e+00 & 9.07e-46 & 2.71e-45 & \bf{0.00e+00} & \bf{0.00e+00} & \bf{0.00e+00} & \bf{0.00e+00}  \\ 
F20& 1.34e+06 & 1.88e+06 & 4.69e-34 & 6.10e-34 & 1.03e-87 & 5.80e-87 & 1.29e+00 & 8.21e+00 & 1.27e+07 & 6.98e+06 & 3.73e-138 & 2.48e-137 & \bf{0.00e+00} & \bf{0.00e+00} & \bf{0.00e+00} & \bf{0.00e+00}  \\ 
F21& 0.00e+00 & \bf{0.00e+00} & 0.00e+00 & \bf{0.00e+00} & 0.00e+00 & \bf{0.00e+00} & 0.00e+00 & \bf{0.00e+00} & 0.00e+00 & \bf{0.00e+00} & 0.00e+00 & \bf{0.00e+00} & 0.00e+00 & \bf{0.00e+00} & \bf{-7.00e-01} & 4.63e-01  \\ 
F22& 1.77e+00 & 1.31e+00 & 1.76e-01 & 4.31e-02 & 1.42e-01 & 7.02e-02 & 1.34e+00 & 5.42e-01 & 4.43e+00 & 6.56e-01 & 9.39e-02 & 2.40e-02 & \bf{0.00e+00} & \bf{0.00e+00} & 7.79e-02 & 4.18e-02  \\ 
F23& 4.97e-01 & 8.40e-03 & 1.00e-01 & 1.08e-01 & 1.19e-01 & 1.27e-01 & 4.99e-01 & 1.55e-03 & 5.00e-01 & 1.17e-03 & 4.19e-02 & 8.64e-03 & \bf{0.00e+00} & \bf{0.00e+00} & 2.97e-02 & 2.06e-02  \\ 
\hline
\end{tabular}
\end{table}

\begin{table}[th]
\tiny
\caption{\label{tabla:resultados30_funciones_fijas}Performance of meta-heuristic algorithms compared with the CCAA on fixed-dimensional problems} 
\begin{tabular}{lllllllllllllllll}
\hline
Function & \multicolumn{2}{l}{CPSO}& \multicolumn{2}{l}{GWO}& \multicolumn{2}{l}{WOA}& \multicolumn{2}{l}{MA}& \multicolumn{2}{l}{ADIFA}& \multicolumn{2}{l}{RDWOA}& \multicolumn{2}{l}{MSCA}& \multicolumn{2}{l}{CCAA}\\ 
\hline
& mean & std& mean & std& mean & std& mean & std& mean & std& mean & std& mean & std& mean & std \\ 
F24& 1.02e+00 & 1.41e-01 & 3.00e+00 & 3.29e+00 & 1.85e+00 & 2.00e+00 & \bf{9.98e-01} & \bf{0.00e+00} & 1.06e+00 & 2.38e-01 & 2.41e+00 & 3.30e+00 & 9.98e-01 & 3.57e-11 & \bf{9.98e-01} & 1.31e-16  \\ 
F25& 7.24e-03 & 1.14e-02 & 1.99e-03 & 5.48e-03 & 6.37e-04 & 2.69e-04 & 1.20e-03 & 3.96e-03 & 7.03e-04 & 4.35e-04 & 2.00e-03 & 1.19e-02 & \bf{3.59e-04} & \bf{1.83e-04} & 7.93e-04 & 2.84e-03  \\ 
F26& \bf{-1.03e+00} & 2.37e-16 & -1.03e+00 & 9.79e-09 & -1.03e+00 & 7.14e-11 & \bf{-1.03e+00} & \bf{2.24e-16} & -1.03e+00 & 1.33e-11 & \bf{-1.03e+00} & 5.60e-16 & -1.03e+00 & 2.00e-06 & -1.03e+00 & 1.41e-14  \\ 
F27& \bf{3.98e-01} & \bf{3.36e-16} & 3.98e-01 & 3.78e-07 & 3.98e-01 & 3.12e-07 & \bf{3.98e-01} & \bf{3.36e-16} & 3.98e-01 & 3.43e-12 & 3.98e-01 & 6.04e-16 & 3.98e-01 & 3.56e-05 & \bf{3.98e-01} & \bf{3.36e-16}  \\ 
F28& 3.00e+00 & \bf{7.59e-16} & 3.00e+00 & 8.24e-06 & 3.00e+00 & 1.23e-05 & \bf{3.00e+00} & 3.42e-15 & 3.00e+00 & 6.93e-10 & 3.00e+00 & 1.24e-12 & 3.00e+00 & 2.30e-06 & 4.08e+00 & 5.34e+00  \\ 
F29& \bf{-3.86e+00} & 3.14e-15 & -3.86e+00 & 2.15e-03 & -3.86e+00 & 2.20e-03 & \bf{-3.86e+00} & 3.14e-15 & -3.86e+00 & 1.43e-09 & \bf{-3.86e+00} & \bf{2.49e-15} & -3.86e+00 & 3.71e-03 & \bf{-3.86e+00} & 2.59e-15  \\ 
F30& -3.28e+00 & 5.60e-02 & -3.25e+00 & 7.83e-02 & -3.24e+00 & 9.09e-02 & -3.28e+00 & 5.83e-02 & -3.23e+00 & 5.13e-02 & -3.27e+00 & 5.99e-02 & -3.13e+00 & 1.12e-01 & \bf{-3.32e+00} & \bf{1.70e-09}  \\ 
F31& -9.13e+00 & 2.06e+00 & -9.75e+00 & 1.38e+00 & -9.48e+00 & 1.83e+00 & -6.25e+00 & 3.58e+00 & -8.85e+00 & 2.18e+00 & -9.24e+00 & 1.98e+00 & -1.02e+01 & 1.26e-04 & \bf{-1.02e+01} & \bf{5.45e-12}  \\ 
F32& -9.23e+00 & 2.22e+00 & -1.03e+01 & 7.52e-01 & -9.16e+00 & 2.39e+00 & -7.04e+00 & 3.57e+00 & -9.32e+00 & 2.14e+00 & -8.38e+00 & 2.61e+00 & -1.04e+01 & 1.92e-04 & \bf{-1.04e+01} & \bf{3.06e-12}  \\ 
F33& -1.01e+01 & 1.48e+00 & -1.01e+01 & 1.76e+00 & -8.90e+00 & 2.83e+00 & -7.58e+00 & 3.53e+00 & -9.65e+00 & 2.07e+00 & -8.70e+00 & 2.59e+00 & -1.05e+01 & 1.05e-04 & \bf{-1.05e+01} & \bf{1.33e-12}  \\ 
\hline
\end{tabular}
\end{table}

\end{landscape}
\restoregeometry

Table \ref{tabla:wilcoxon30} presents the results on the range that each algorithm obtained with respect to its average value in each test problem, for $ 30 $ dimensions and the functions with fixed dimension, as well as the results of the Wilcoxon rank-sum test to compare the CCAA with the other algorithms for each test function. Use of the $ + $ symbol indicates that the CCAA obtained a significantly better statistical result than the other algorithm, a $ - $ symbol indicates a significantly worse statistical result, and the $ \ approx $ symbol indicates no statistically significant difference. The penultimate row shows each algorithm's average rank and the difference between the number of positive tests ($ + $) minus the number of negative tests ($ - $), where a positive value indicates that the CCAA obtained a more significant number of positive statistical tests against the reference algorithm. Finally, the last row indicates each algorithm's rank with regard to the average range of the $ 33 $ test problems.

In Table \ref{tabla:wilcoxon30},  the CCAA always achieved a favorable comparison against the other algorithms for the Wilcoxon rank-sum test and its average rank is the best ($ 1.55 $), followed closely by the MSCA ($ 2.15$). These results show the excellent overall effectiveness of the CCAA for problems in $ 30 $ dimensions and with fixed dimensions.

\begin{table}[th]
\tiny
\caption{\label{tabla:wilcoxon30}Results of the Wilcoxon rank-sum test and ranking of all the compared algorithms on $30$-dimensional and fixed-dimensional problems} 
\begin{tabular}{cccccccccccccccc}
\hline
Function & \multicolumn{2}{l}{CPSO}& \multicolumn{2}{l}{GWO}& \multicolumn{2}{l}{WOA}& \multicolumn{2}{l}{MA}& \multicolumn{2}{l}{ADIFA}& \multicolumn{2}{l}{RDWOA}& \multicolumn{2}{l}{MSCA}& MSCA\\ 
\hline
& Rank & Test& Rank & Test& Rank & Test& Rank & Test& Rank & Test& Rank & Test& Rank & Test& Rank \\ 
F1&  6 & + &  4 & + &  3 & + &  5 & + &  7 & + &  2 & + &  1 & $ \approx $ &  1  \\ 
F2&  6 & + &  4 & + &  3 & + &  5 & + &  7 & + &  2 & + &  1 & $ \approx $ &  1  \\ 
F3&  6 & + &  4 & + &  3 & + &  5 & + &  7 & + &  2 & + &  1 & $ \approx $ &  1  \\ 
F4&  5 & + &  3 & + &  7 & + &  6 & + &  4 & + &  2 & + &  1 & $ \approx $ &  1  \\ 
F5&  4 & + &  3 & + &  5 & + &  7 & + &  6 & + &  2 & + &  1 & $ \approx $ &  1  \\ 
F6&  7 & + &  4 & + &  5 & + &  6 & + &  8 & + &  3 & + &  2 & + &  1  \\ 
F7&  4 & + &  7 & + &  5 & + &  3 & + &  8 & + &  6 & + &  2 & + &  1  \\ 
F8&  5 & + &  4 & + &  3 & + &  6 & + &  7 & + &  2 & + &  1 & $ \approx $ &  1  \\ 
F9&  7 & + &  4 & + &  5 & + &  6 & + &  8 & + &  2 & - &  1 & - &  3  \\ 
F10&  6 & + &  4 & + &  3 & + &  5 & + &  7 & + &  2 & + &  1 & $ \approx $ &  1  \\ 
F11&  5 & + &  8 & + &  3 & + &  4 & + &  7 & + &  6 & + &  2 & + &  1  \\ 
F12&  6 & + &  3 & + &  1 & $ \approx $ &  4 & + &  5 & + &  1 & $ \approx $ &  1 & $ \approx $ &  2  \\ 
F13&  7 & + &  4 & + &  3 & + &  5 & + &  6 & + &  2 & + &  1 & $ \approx $ &  1  \\ 
F14&  5 & + &  3 & + &  4 & + &  6 & + &  7 & + &  2 & + &  1 & $ \approx $ &  1  \\ 
F15&  6 & + &  5 & + &  3 & + &  7 & + &  8 & + &  4 & + &  2 & + &  1  \\ 
F16&  7 & + &  6 & + &  3 & + &  4 & + &  8 & + &  5 & + &  2 & + &  1  \\ 
F17&  7 & + &  6 & + &  1 & - &  5 & + &  8 & + &  2 & - &  4 & $ \approx $ &  3  \\ 
F18&  4 & + &  1 & $ \approx $ &  1 & $ \approx $ &  2 & + &  3 & + &  1 & $ \approx $ &  1 & $ \approx $ &  1  \\ 
F19&  6 & + &  4 & + &  3 & + &  5 & + &  7 & + &  2 & + &  1 & $ \approx $ &  1  \\ 
F20&  6 & + &  4 & + &  3 & + &  5 & + &  7 & + &  2 & + &  1 & $ \approx $ &  1  \\ 
F21&  2 & + &  2 & + &  2 & + &  2 & + &  2 & + &  2 & + &  2 & + &  1  \\ 
F22&  7 & + &  5 & + &  4 & $ \approx $ &  6 & + &  8 & + &  3 & + &  1 & - &  2  \\ 
F23&  6 & + &  4 & + &  5 & $ \approx $ &  7 & + &  8 & + &  3 & + &  1 & - &  2  \\ 
F24&  3 & + &  7 & + &  5 & + &  1 & $ \approx $ &  4 & + &  6 & + &  2 & + &  1  \\ 
F25&  8 & $ \approx $ &  6 & + &  2 & - &  5 & + &  3 & - &  7 & + &  1 & - &  4  \\ 
F26&  1 & - &  5 & + &  4 & + &  1 & - &  3 & + &  1 & $ \approx $ &  6 & + &  2  \\ 
F27&  1 & $ \approx $ &  5 & + &  4 & + &  1 & $ \approx $ &  3 & + &  2 & + &  6 & + &  1  \\ 
F28&  2 & - &  7 & - &  6 & - &  1 & - &  4 & - &  3 & $ \approx $ &  5 & - &  8  \\ 
F29&  1 & $ \approx $ &  3 & + &  4 & + &  1 & $ \approx $ &  2 & + &  1 & $ \approx $ &  5 & + &  1  \\ 
F30&  2 & + &  5 & + &  6 & + &  3 & + &  7 & + &  4 & $ \approx $ &  8 & + &  1  \\ 
F31&  6 & + &  3 & + &  4 & + &  8 & $ \approx $ &  7 & + &  5 & + &  2 & + &  1  \\ 
F32&  5 & + &  3 & + &  6 & + &  8 & $ \approx $ &  4 & + &  7 & $ \approx $ &  2 & + &  1  \\ 
F33&  3 & + &  4 & + &  6 & + &  8 & $ \approx $ &  5 & + &  7 & $ \approx $ &  2 & + &  1  \\ 
Avg. \& WRST& 4.91 & 26 & 4.36 & 30 & 3.79 & 23 & 4.64 & 23 & 5.91 & 29 & 3.12 & 21 & 2.15 &  9 & 1.55  \\ 
Overall rank&  7 & &  5 & &  4 & &  6 & &  8 & &  3 & &  2 & &  1  \\ 
\hline
\end{tabular}
\end{table}

\begin{figure}[h!]
\begin{center}
\includegraphics[scale=0.29]{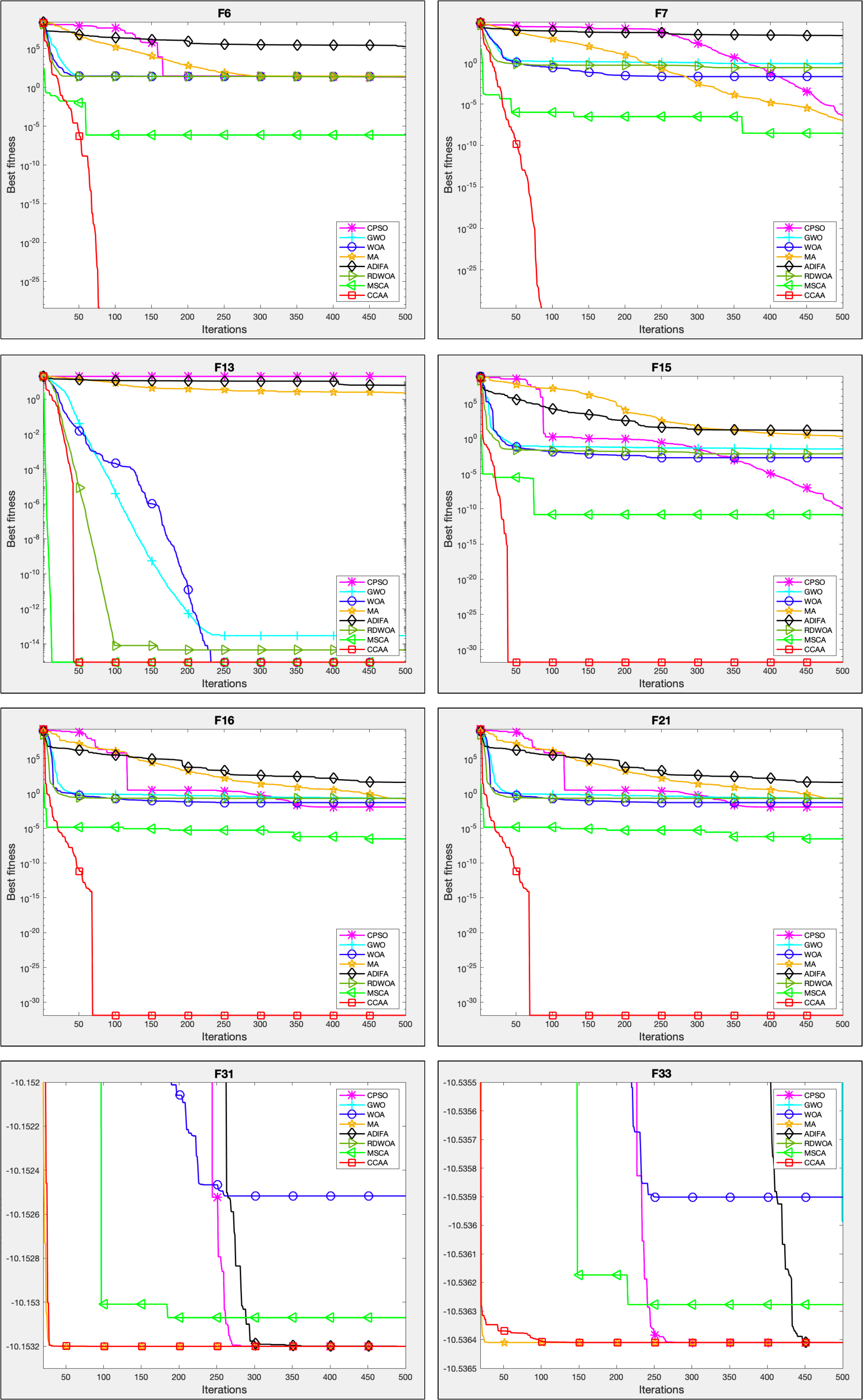}
\caption{Convergence curves of the different algorithms for test functions in $ 30 $ dimensions}
\label{fig:Convergencia_30}
\end{center}
\end{figure}

\begin{figure}[h!]
\begin{center}
\includegraphics[scale=0.3]{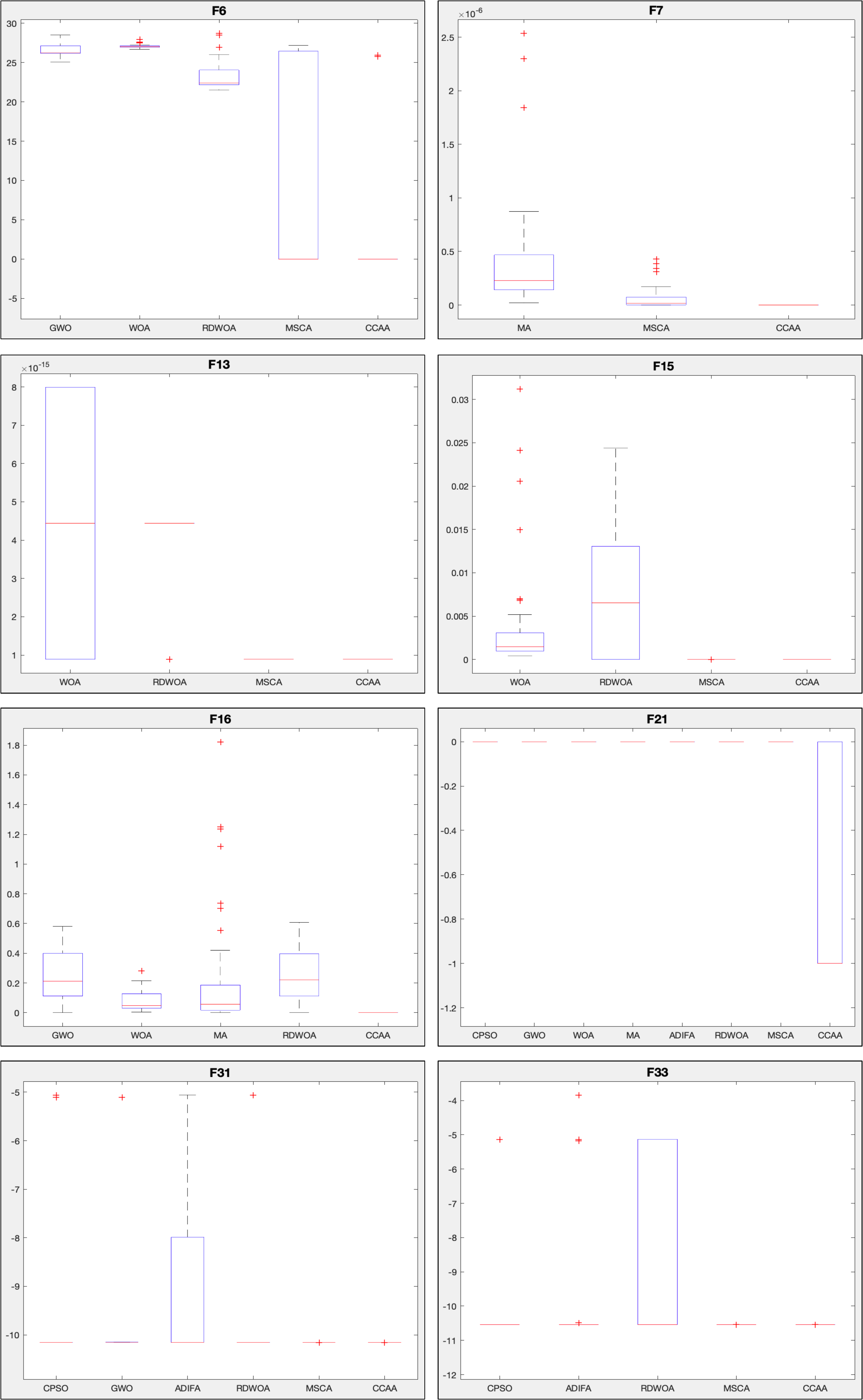}
\caption{Box plots with the best results of different algorithms for test functions in $30$ dimensions}
\label{fig:Box_Plot_30}
\end{center}
\end{figure}

Figure \ref{fig:Convergencia_30} presents a sample of the convergence curves of the algorithms used in this experiment for different test problems. Figure \ref{fig:Box_Plot_30} shows the box plots for the algorithms that obtained the best results in the test problems described in Fig. \ref{fig:Convergencia_30}.

\newpage

\subsection{Experiment 2: Functions on $ 500 $ dimensions and functions with fixed dimensions}

This section compares the performance of the CCAA against the  $ 6 $other algorithms for the $ 23 $ scalable test functions using $ 500 $ dimensions. A new comparison with the $ 10 $ fixed-dimension functions is also shown to result in findings similar to the first one.

There were $ 30 $ independent runs per algorithm in this case as well; for the CPSO and the CCAA, $ smart\_n = 12 $ and $ neighbor\_n = 6 $ were used; for the rest of the algorithms, $ (smart\_n) (neighbor\_n) = 72 $ individuals were applied. In all algorithms, $ iteration\_n = 500 $ was employed. Table \ref{tabla:resultados500_funciones_unimodales} presents the results of the comparison with respect to the mean and standard deviation for the unimodal functions, Table \ref{tabla:resultados500_funciones_multimodales} shows the same comparison for multimodal functions, and Table \ref{tabla:resultados500_funciones_fijas} describes the results for functions with fixed dimensions.

For unimodal problems, the CCAA obtains $ 8 $ of the $ 10 $ best average values, i.e., the CCAA was able to obtain optimal solutions for $ F1 $ to $ F3 $, $ F5 $, $ F7 $, $ F8 $, and $ F10 $. Moreover, in $ 7$ out of $ 10$ cases, the CCAA achieved the best values concerning the standard deviation, which again gives evidence of its exploitation capacity.

For the $ 13 $ multimodal problems, the CCAA also produced $ 10$ of the $ 13 $ best average values in the functions $ F11 $ to $ F16 $ and from $ F18 $ to $ F21 $. In $ 7 $ of them ($ F11 $, $ F12 $, $ F14 $, and from $ F18 $ to $ F21 $), the optimal values were calculated. The performance of the CCAA is comparable with that of the MSCA, which obtains $ 9 $ better average values, with $ 8 $ of them were optimal values. In $ 9 $ out of $ 13 $ cases, the CCAA achieved the best values referring to the standard deviation, which also shows the CCAA's exploration ability.

For the second run with $10$ fixed-dimension test functions, the CCAA produced $ 6$ of the best average values for the functions of $ F27 $ and from $ F29 $ to $ F33 $. In all of them, the CCAA obtained the optimal values. The performance of the CCAA is comparable to that of the MA, which produced $ 5 $ better average values, where again $ 4$ were optimal values. For this experiment, the CCAA reached the lowest value in $ 6$ of $ 10$ cases for the standard deviation, showing once again the robustness of the CCAA to carry out exploration and exploitation actions in a balanced way.

Table \ref{tabla:wilcoxon500} presents the results on the range that each algorithm obtained with respect to its average value in each test problem, for $ 500 $ dimensions and functions with fixed dimension, as well as the results of the Wilcoxon rank-sum test to compare the CCAA with the other algorithms for each test function. In Table \ref{tabla:wilcoxon500},  the CCAA always has a favorable comparison versus the other algorithms concerning the Wilcoxon rank-sum test and its average rank is the best ($ 1.45 $), followed again by the MSCA ($ 2.21$). These results show how effective the CCAA is in general for problems with $ 500$  and fixed dimensions.

\newgeometry{margin=1.5cm}
\begin{landscape}

\begin{table}[th]
\tiny
\caption{\label{tabla:resultados500_funciones_unimodales}Performance of meta-heuristic algorithms compared with the CCAA on 500-dimensional unimodal problems} 
\begin{tabular}{lllllllllllllllll}
\hline
Function & \multicolumn{2}{l}{CPSO}& \multicolumn{2}{l}{GWO}& \multicolumn{2}{l}{WOA}& \multicolumn{2}{l}{MA}& \multicolumn{2}{l}{ADIFA}& \multicolumn{2}{l}{RDWOA}& \multicolumn{2}{l}{MSCA}& \multicolumn{2}{l}{CCAA}\\ 
\hline
& mean & std& mean & std& mean & std& mean & std& mean & std& mean & std& mean & std& mean & std \\ 
F1& 5.01e+05 & 1.15e+05 & 3.99e-05 & 1.39e-05 & 1.19e-84 & 8.36e-84 & 9.74e+05 & 1.25e+05 & 5.96e+04 & 9.37e+03 & 3.29e-81 & 4.91e-81 & \bf{0.00e+00} & \bf{0.00e+00} & \bf{0.00e+00} & \bf{0.00e+00}  \\ 
F2& 1.13e+06 & 2.17e+05 & 7.18e-05 & 2.93e-05 & 3.37e-87 & 1.95e-86 & 1.56e+06 & 1.22e+05 & 1.73e+05 & 2.56e+04 & 6.49e-81 & 1.13e-80 & \bf{0.00e+00} & \bf{0.00e+00} & \bf{0.00e+00} & \bf{0.00e+00}  \\ 
F3& 1.32e+03 & 3.73e+02 & 1.33e-03 & 2.34e-04 & 4.60e-54 & 1.78e-53 & 3.83e+50 & 2.65e+51 & 9.34e+12 & 6.60e+13 & 8.32e-55 & 2.16e-54 & \bf{0.00e+00} & \bf{0.00e+00} & \bf{0.00e+00} & \bf{0.00e+00}  \\ 
F4& 3.23e+06 & 5.09e+05 & 1.87e+05 & 5.74e+04 & 2.60e+07 & 7.06e+06 & 3.99e+06 & 5.39e+05 & 6.03e+05 & 1.97e+05 & 7.50e-61 & 1.53e-61 & \bf{0.00e+00} & \bf{0.00e+00} & 3.08e-07 & 1.71e-06  \\ 
F5& 6.50e+01 & 6.09e+00 & 5.64e+01 & 6.23e+00 & 7.03e+01 & 2.86e+01 & 9.91e+01 & 1.86e-01 & 5.54e+01 & 6.73e+00 & 2.92e-32 & 6.61e-34 & \bf{0.00e+00} & \bf{0.00e+00} & \bf{0.00e+00} & \bf{0.00e+00}  \\ 
F6& 1.09e+09 & 3.42e+08 & 4.97e+02 & 2.83e-01 & 4.95e+02 & \bf{1.91e-01} & 7.10e+09 & 1.95e+08 & 1.60e+07 & 4.92e+06 & 4.95e+02 & 1.05e+00 & 1.88e+02 & 2.42e+02 & \bf{9.86e+00} & 6.97e+01  \\ 
F7& 4.69e+05 & 9.99e+04 & 8.34e+01 & 1.95e+00 & 1.19e+01 & 2.74e+00 & 9.40e+05 & 1.04e+05 & 6.10e+04 & 9.37e+03 & 5.43e+01 & 4.04e+00 & 1.93e-06 & 5.98e-06 & \bf{0.00e+00} & \bf{0.00e+00}  \\ 
F8& 7.61e+03 & 1.81e+03 & 7.45e-13 & 6.07e-13 & 2.13e-147 & 1.14e-146 & 5.39e+04 & 1.10e+04 & 1.59e+02 & 4.60e+01 & 7.55e-142 & 9.39e-142 & \bf{0.00e+00} & \bf{0.00e+00} & \bf{0.00e+00} & \bf{0.00e+00}  \\ 
F9& 7.32e+03 & 3.02e+03 & 2.15e-02 & 5.44e-03 & 2.56e-03 & 3.60e-03 & 5.15e+04 & 1.28e+04 & 1.51e+02 & 4.27e+01 & 5.92e-05 & 5.53e-05 & \bf{4.19e-05} & \bf{4.77e-05} & 2.24e-04 & 1.53e-04  \\ 
F10& 1.30e+02 & 1.58e+01 & 1.47e-04 & 1.75e-05 & 4.20e-54 & 2.09e-53 & 7.97e+01 & 7.49e+00 & 5.12e+01 & 4.08e+00 & 1.39e-55 & 3.82e-55 & \bf{0.00e+00} & \bf{0.00e+00} & \bf{0.00e+00} & \bf{0.00e+00}  \\ 
\hline
\end{tabular}
\end{table}

\begin{table}[th]
\tiny
\caption{\label{tabla:resultados500_funciones_multimodales}Performance of meta-heuristic algorithms compared with the CCAA on 500-dimensional multimodal problems} 
\begin{tabular}{lllllllllllllllll}
\hline
Function & \multicolumn{2}{l}{CPSO}& \multicolumn{2}{l}{GWO}& \multicolumn{2}{l}{WOA}& \multicolumn{2}{l}{MA}& \multicolumn{2}{l}{ADIFA}& \multicolumn{2}{l}{RDWOA}& \multicolumn{2}{l}{MSCA}& \multicolumn{2}{l}{CCAA}\\ 
\hline
& mean & std& mean & std& mean & std& mean & std& mean & std& mean & std& mean & std& mean & std \\ 
F11& -8.93e+04 & 3.14e+03 & -6.26e+04 & 8.10e+03 & -1.89e+05 & 2.54e+04 & -8.40e+04 & 4.76e+03 & -3.42e+04 & 3.40e+03 & -1.15e+05 & 1.34e+04 & -2.09e+05 & 1.63e-01 & \bf{-2.09e+05} & \bf{1.18e-10}  \\ 
F12& 5.61e+03 & 5.67e+02 & 4.67e+01 & 1.51e+01 & 1.82e-14 & 1.29e-13 & 4.91e+03 & 2.79e+02 & 3.62e+03 & 5.51e+02 & \bf{0.00e+00} & \bf{0.00e+00} & \bf{0.00e+00} & \bf{0.00e+00} & \bf{0.00e+00} & \bf{0.00e+00}  \\ 
F13& 2.00e+01 & 2.65e-04 & 2.94e-04 & 5.47e-05 & 4.37e-15 & 2.64e-15 & 2.04e+01 & 3.40e-01 & 1.26e+01 & 1.04e+00 & 4.37e-15 & 5.02e-16 & \bf{8.88e-16} & \bf{0.00e+00} & \bf{8.88e-16} & \bf{0.00e+00}  \\ 
F14& 4.59e+03 & 8.43e+02 & 2.94e-03 & 1.05e-02 & \bf{0.00e+00} & \bf{0.00e+00} & 9.02e+03 & 1.21e+03 & 4.63e+02 & 7.61e+01 & \bf{0.00e+00} & \bf{0.00e+00} & \bf{0.00e+00} & \bf{0.00e+00} & \bf{0.00e+00} & \bf{0.00e+00}  \\ 
F15& 1.68e+09 & 7.08e+08 & 6.57e-01 & 3.33e-02 & 1.96e-02 & 6.25e-03 & 1.76e+10 & 5.16e+08 & 7.05e+04 & 1.01e+05 & 2.24e-01 & 3.17e-02 & 1.39e-09 & 1.66e-09 & \bf{9.42e-34} & \bf{6.91e-49}  \\ 
F16& 4.18e+09 & 1.48e+09 & 4.65e+01 & 1.19e+00 & 5.76e+00 & 1.74e+00 & 3.19e+10 & 1.08e+09 & 1.03e+07 & 6.89e+06 & 3.15e+01 & 1.34e+00 & 9.89e-01 & 6.99e+00 & \bf{1.35e-32} & \bf{1.11e-47}  \\ 
F17& 6.19e+02 & 6.20e+01 & 4.00e-02 & 5.98e-03 & 2.19e-53 & 1.22e-52 & 5.06e+02 & 5.62e+01 & 3.57e+02 & 3.88e+01 & 2.98e-14 & 1.77e-13 & \bf{0.00e+00} & \bf{0.00e+00} & 6.08e-12 & 3.76e-11  \\ 
F18& 1.02e+02 & 1.08e+01 & 3.06e-08 & 1.01e-08 & \bf{0.00e+00} & \bf{0.00e+00} & 1.56e+02 & 4.44e+01 & 4.83e+01 & 5.41e+00 & \bf{0.00e+00} & \bf{0.00e+00} & \bf{0.00e+00} & \bf{0.00e+00} & \bf{0.00e+00} & \bf{0.00e+00}  \\ 
F19& 1.84e+03 & 9.99e+01 & 7.96e-01 & 6.42e-02 & 1.36e-30 & 5.53e-30 & 1.80e+03 & 3.86e+01 & 1.27e+03 & 7.13e+01 & 2.63e-32 & 5.59e-32 & \bf{0.00e+00} & \bf{0.00e+00} & \bf{0.00e+00} & \bf{0.00e+00}  \\ 
F20& 1.12e+10 & 3.10e+09 & 6.53e-02 & 1.99e-02 & 7.39e-82 & 4.79e-81 & 8.11e+09 & 1.22e+09 & 3.41e+09 & 6.40e+08 & 6.61e-78 & 1.71e-77 & \bf{0.00e+00} & \bf{0.00e+00} & \bf{0.00e+00} & \bf{0.00e+00}  \\ 
F21& 0.00e+00 & \bf{0.00e+00} & 0.00e+00 & \bf{0.00e+00} & 0.00e+00 & \bf{0.00e+00} & 0.00e+00 & \bf{0.00e+00} & 0.00e+00 & \bf{0.00e+00} & 0.00e+00 & \bf{0.00e+00} & 0.00e+00 & \bf{0.00e+00} & \bf{-2.60e-01} & 4.43e-01  \\ 
F22& 8.98e+01 & 4.86e+00 & 8.18e-01 & 8.50e-02 & 1.26e-01 & 6.94e-02 & 1.07e+02 & 3.27e+00 & 2.47e+01 & 1.97e+00 & 8.79e-02 & 3.28e-02 & \bf{0.00e+00} & \bf{0.00e+00} & 1.80e-02 & 3.88e-02  \\ 
F23& 5.00e-01 & 4.86e-11 & 4.97e-01 & 1.42e-03 & 1.00e-01 & 1.08e-01 & 5.00e-01 & 1.75e-11 & 5.00e-01 & 6.97e-08 & 4.19e-02 & 8.64e-03 & \bf{0.00e+00} & \bf{0.00e+00} & 7.86e-03 & 1.69e-02  \\ 
\hline
\end{tabular}
\end{table}

\begin{table}[th]
\tiny
\caption{\label{tabla:resultados500_funciones_fijas}Performance of meta-heuristic algorithms compared with the CCAA on fixed-dimensional problems} 
\begin{tabular}{lllllllllllllllll}
\hline
Function & \multicolumn{2}{l}{CPSO}& \multicolumn{2}{l}{GWO}& \multicolumn{2}{l}{WOA}& \multicolumn{2}{l}{MA}& \multicolumn{2}{l}{ADIFA}& \multicolumn{2}{l}{RDWOA}& \multicolumn{2}{l}{MSCA}& \multicolumn{2}{l}{CCAA}\\ 
\hline
& mean & std& mean & std& mean & std& mean & std& mean & std& mean & std& mean & std& mean & std \\ 
F24& 1.04e+00 & 1.97e-01 & 2.68e+00 & 3.25e+00 & 1.34e+00 & 7.12e-01 & \bf{9.98e-01} & \bf{4.49e-17} & 1.06e+00 & 3.11e-01 & 2.53e+00 & 3.04e+00 & 1.23e+00 & 1.65e+00 & 1.06e+00 & 3.11e-01  \\ 
F25& 8.44e-03 & 9.84e-03 & 2.35e-03 & 6.07e-03 & 7.52e-04 & 5.06e-04 & 1.21e-03 & 3.96e-03 & 9.00e-04 & 4.21e-04 & \bf{3.28e-04} & \bf{1.30e-04} & 3.64e-04 & 1.90e-04 & 3.62e-04 & 2.20e-04  \\ 
F26& \bf{-1.03e+00} & \bf{2.24e-16} & -1.03e+00 & 8.63e-09 & -1.03e+00 & 6.09e-11 & \bf{-1.03e+00} & \bf{2.24e-16} & -1.03e+00 & 8.45e-12 & \bf{-1.03e+00} & 5.74e-16 & -1.03e+00 & 1.21e-06 & -1.03e+00 & 4.51e-13  \\ 
F27& \bf{3.98e-01} & \bf{3.36e-16} & 3.98e-01 & 2.87e-07 & 3.98e-01 & 3.35e-07 & \bf{3.98e-01} & \bf{3.36e-16} & 3.98e-01 & 5.24e-12 & 3.98e-01 & 4.60e-16 & 3.98e-01 & 4.03e-05 & \bf{3.98e-01} & \bf{3.36e-16}  \\ 
F28& 3.00e+00 & \bf{1.76e-15} & 3.00e+00 & 7.63e-06 & 3.00e+00 & 7.48e-06 & \bf{3.00e+00} & 3.56e-15 & 3.00e+00 & 5.59e-10 & 3.54e+00 & 3.82e+00 & 3.00e+00 & 1.32e-06 & 3.00e+00 & 1.47e-14  \\ 
F29& \bf{-3.86e+00} & 3.14e-15 & -3.86e+00 & 2.04e-03 & -3.86e+00 & 1.88e-03 & \bf{-3.86e+00} & 3.14e-15 & -3.86e+00 & 2.39e-09 & -3.86e+00 & 1.11e-03 & -3.86e+00 & 3.84e-03 & \bf{-3.86e+00} & \bf{2.60e-15}  \\ 
F30& -3.27e+00 & 5.96e-02 & -3.23e+00 & 7.11e-02 & -3.24e+00 & 7.53e-02 & -3.27e+00 & 5.88e-02 & -3.23e+00 & 5.27e-02 & -3.26e+00 & 6.00e-02 & -3.18e+00 & 9.47e-02 & \bf{-3.32e+00} & \bf{1.68e-02}  \\ 
F31& -9.24e+00 & 1.98e+00 & -9.55e+00 & 1.66e+00 & -8.73e+00 & 2.62e+00 & -6.75e+00 & 3.63e+00 & -8.98e+00 & 2.12e+00 & -8.52e+00 & 2.40e+00 & -1.02e+01 & 1.90e-04 & \bf{-1.02e+01} & \bf{1.01e-11}  \\ 
F32& -9.49e+00 & 2.16e+00 & -1.02e+01 & 1.04e+00 & -9.13e+00 & 2.45e+00 & -7.07e+00 & 3.66e+00 & -8.99e+00 & 2.41e+00 & -8.06e+00 & 2.67e+00 & -1.04e+01 & 9.64e-05 & \bf{-1.04e+01} & \bf{4.60e-13}  \\ 
F33& -9.67e+00 & 2.00e+00 & -1.05e+01 & 7.14e-04 & -8.91e+00 & 2.66e+00 & -7.66e+00 & 3.73e+00 & -9.30e+00 & 2.37e+00 & -8.37e+00 & 2.68e+00 & -1.05e+01 & 1.22e-04 & \bf{-1.05e+01} & \bf{2.14e-12}  \\ 
\hline
\end{tabular}
\end{table}

\end{landscape}
\restoregeometry

\begin{table}[th]
\tiny
\caption{\label{tabla:wilcoxon500}Results of the Wilcoxon rank-sum test and ranking of all the compared algorithms on $500$-dimensional and fixed-dimensional problems.} 
\begin{tabular}{cccccccccccccccc}
\hline
Function & \multicolumn{2}{l}{CPSO}& \multicolumn{2}{l}{GWO}& \multicolumn{2}{l}{WOA}& \multicolumn{2}{l}{MA}& \multicolumn{2}{l}{ADIFA}& \multicolumn{2}{l}{RDWOA}& \multicolumn{2}{l}{MSCA}& MSCA\\ 
\hline
& Rank & Test& Rank & Test& Rank & Test& Rank & Test& Rank & Test& Rank & Test& Rank & Test& Rank \\ 
F1&  6 & + &  4 & + &  2 & + &  7 & + &  5 & + &  3 & + &  1 & $ \approx $ &  1  \\ 
F2&  6 & + &  4 & + &  2 & + &  7 & + &  5 & + &  3 & + &  1 & $ \approx $ &  1  \\ 
F3&  5 & + &  4 & + &  3 & + &  7 & + &  6 & + &  2 & + &  1 & $ \approx $ &  1  \\ 
F4&  6 & + &  4 & + &  8 & + &  7 & + &  5 & + &  2 & $ \approx $ &  1 & - &  3  \\ 
F5&  5 & + &  4 & + &  6 & + &  7 & + &  3 & + &  2 & + &  1 & $ \approx $ &  1  \\ 
F6&  7 & + &  5 & + &  3 & + &  8 & + &  6 & + &  4 & + &  2 & + &  1  \\ 
F7&  7 & + &  5 & + &  3 & + &  8 & + &  6 & + &  4 & + &  2 & + &  1  \\ 
F8&  6 & + &  4 & + &  2 & + &  7 & + &  5 & + &  3 & + &  1 & $ \approx $ &  1  \\ 
F9&  7 & + &  5 & + &  4 & + &  8 & + &  6 & + &  2 & - &  1 & - &  3  \\ 
F10&  7 & + &  4 & + &  3 & + &  6 & + &  5 & + &  2 & + &  1 & $ \approx $ &  1  \\ 
F11&  5 & + &  7 & + &  3 & + &  6 & + &  8 & + &  4 & + &  2 & + &  1  \\ 
F12&  6 & + &  3 & + &  2 & $ \approx $ &  5 & + &  4 & + &  1 & $ \approx $ &  1 & $ \approx $ &  1  \\ 
F13&  5 & + &  3 & + &  2 & + &  6 & + &  4 & + &  2 & + &  1 & $ \approx $ &  1  \\ 
F14&  4 & + &  2 & + &  1 & $ \approx $ &  5 & + &  3 & + &  1 & $ \approx $ &  1 & $ \approx $ &  1  \\ 
F15&  7 & + &  5 & + &  3 & + &  8 & + &  6 & + &  4 & + &  2 & + &  1  \\ 
F16&  7 & + &  5 & + &  3 & + &  8 & + &  6 & + &  4 & + &  2 & + &  1  \\ 
F17&  8 & + &  5 & + &  2 & - &  7 & + &  6 & + &  3 & - &  1 & $ \approx $ &  4  \\ 
F18&  4 & + &  2 & + &  1 & $ \approx $ &  5 & + &  3 & + &  1 & $ \approx $ &  1 & $ \approx $ &  1  \\ 
F19&  7 & + &  4 & + &  3 & + &  6 & + &  5 & + &  2 & + &  1 & $ \approx $ &  1  \\ 
F20&  7 & + &  4 & + &  2 & + &  6 & + &  5 & + &  3 & + &  1 & $ \approx $ &  1  \\ 
F21&  2 & + &  2 & + &  2 & + &  2 & + &  2 & + &  2 & + &  2 & + &  1  \\ 
F22&  7 & + &  5 & + &  4 & + &  8 & + &  6 & + &  3 & + &  1 & - &  2  \\ 
F23&  7 & + &  5 & + &  4 & + &  8 & + &  6 & + &  3 & + &  1 & - &  2  \\ 
F24&  2 & - &  7 & + &  5 & + &  1 & - &  3 & $ \approx $ &  6 & + &  4 & + &  3  \\ 
F25&  8 & $ \approx $ &  7 & + &  4 & + &  6 & + &  5 & + &  1 & - &  3 & + &  2  \\ 
F26&  1 & - &  5 & + &  4 & + &  1 & - &  3 & + &  1 & $ \approx $ &  6 & + &  2  \\ 
F27&  1 & $ \approx $ &  5 & + &  4 & + &  1 & $ \approx $ &  3 & + &  2 & $ \approx $ &  6 & + &  1  \\ 
F28&  2 & - &  7 & + &  6 & + &  1 & - &  4 & + &  8 & + &  5 & + &  3  \\ 
F29&  1 & $ \approx $ &  4 & + &  5 & + &  1 & $ \approx $ &  2 & + &  3 & + &  6 & + &  1  \\ 
F30&  3 & $ \approx $ &  7 & + &  5 & + &  2 & $ \approx $ &  6 & + &  4 & $ \approx $ &  8 & + &  1  \\ 
F31&  4 & + &  3 & + &  6 & + &  8 & $ \approx $ &  5 & + &  7 & $ \approx $ &  2 & + &  1  \\ 
F32&  4 & + &  3 & + &  5 & + &  8 & $ \approx $ &  6 & + &  7 & $ \approx $ &  2 & + &  1  \\ 
F33&  4 & + &  3 & + &  6 & + &  8 & + &  5 & + &  7 & $ \approx $ &  2 & + &  1  \\ 
Avg. \& WRST & 5.09 & 23 & 4.42 & 33 & 3.58 & 28 & 5.73 & 22 & 4.79 & 32 & 3.21 & 17 & 2.21 & 12 & 1.45  \\ 
Overall rank&  7 & &  5 & &  4 & &  8 & &  6 & &  3 & &  2 & &  1  \\ 
\hline
\end{tabular}
\end{table}

\begin{figure}[h!]
\begin{center}
\includegraphics[scale=0.29]{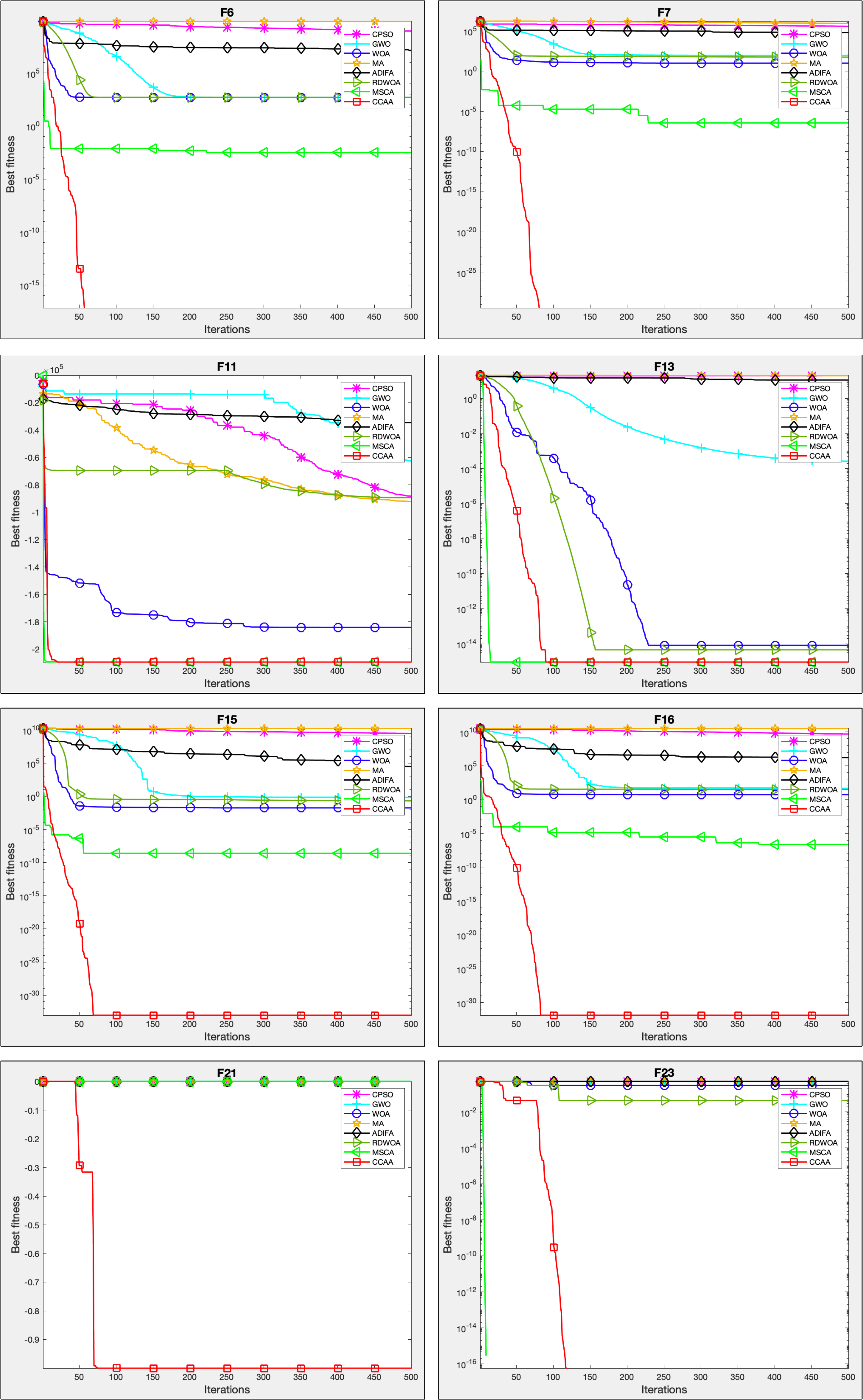}
\caption{Convergence curves of the different algorithms for problems in $ 500 $ dimensions}
\label{fig:Convergencia_500}
\end{center}
\end{figure}

\begin{figure}[h!]
\begin{center}
\includegraphics[scale=0.29]{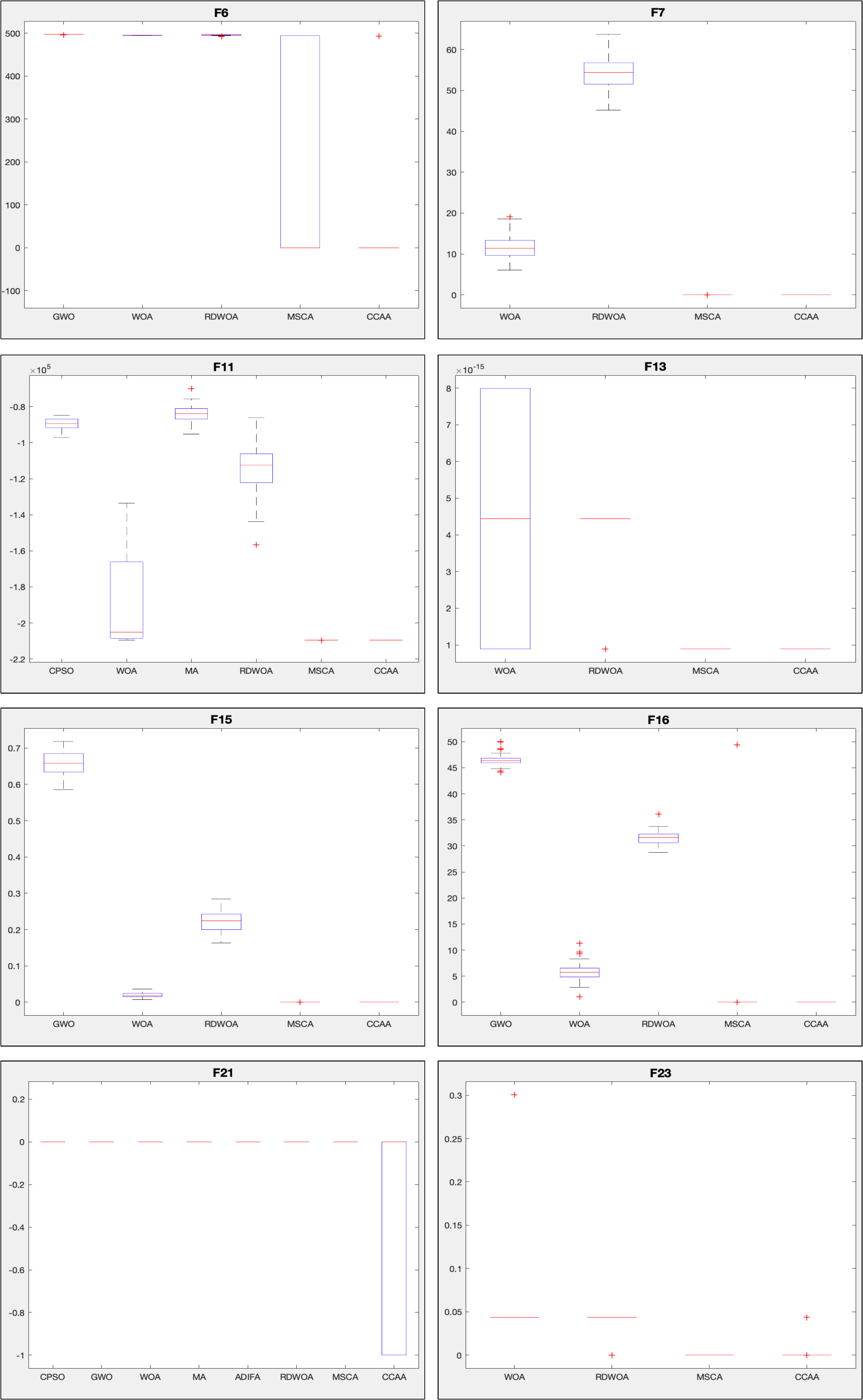}
\caption{Box plots with the best results of different algorithms for problems in $ 500 $ dimensions}
\label{fig:Box_Plot_500}
\end{center}
\end{figure}

Figure \ref{fig:Convergencia_500} presents a sample of the convergence curves of the algorithms used in this experiment for different test problems. Fig. \ref{fig:Box_Plot_500} shows the box plots for the algorithms that obtained the best results in the test problems described in Fig. \ref{fig:Convergencia_500}.

\subsection{Analysis of convergence behavior}

In the computational experiments developed, it was observed that $ smart\_cells $ tended to search extensively for optimal regions of the design space and exploit them concurrently. The $ smart\_cells $ change abruptly in the early stages of the optimization process and then gradually converge. Such behavior can ensure that a population-based algorithm will eventually converge \citep{van2006study}. The convergence curves are compared in Figs. \ref{fig:Convergencia_30} and \ref{fig:Convergencia_500} for some of the problems. In these results, it can be observed that the CCAA is competitive with other recently proposed meta-heuristic algorithms.

The convergence curves of the CCAA show that its convergence tends to accelerate during the initial iterations of the optimization process. This convergence is due to the neighborhood mechanism and the different evolution rules that the $ smart\_cells $ follow which help find promising regions of the search space and exploit them concurrently. Some evolution rules focus on exploration and others on exploitation, which produces an adequate balance to find the global optimum. In general, the experimental success rate of the CCAA algorithm is high as seen in Tables \ref{tabla:wilcoxon30} and \ref{tabla:wilcoxon500}.

In summary, the high exploration capacity is due to the mechanism for updating the position of the $smart\_cells $ using the rules derived from the algorithms \ref{alg:regla_acercamiento}, \ref{alg:regla_alejamiento} and \ref{alg:regla_cambio}.  Simultaneously, the exploitation and convergence is performed using the different versions of the rules that are described in the algorithms \ref{alg:regla_alejamiento}, \ref{alg:regla_incremento}, \ref{alg:regla_mayoria} and \ref{alg:regla_redondeo}. These rules allow $ smart\_cells $ to rapidly relocate themselves to improve their position by exchanging information and using their own values. Since these rules are applied concurrently, the CCAA shows a high speed to avoid converging to local optima. The following sections verify the performance of the CCAA in engineering problems.

\subsection{Engineering design problems}
\label{sec:aplicaciones}

In engineering, there are many problems where mathematical models are applied that are later optimized. To show the utility of the CCAA in these cases, this section tests for $ 4 $ design problems: a gear train, a pressure vessel, a welded beam, and a cantilever beam.

Given that these problems have different restrictions, the cost function is penalizing with an additional extra-large cost (because they are minimization problems) if one of the restrictions is not fulfilled. This objective function increment is simple to implement and has a low computational cost \citep{mirjalili2016whale}.


\subsubsection{Gear train design (GTD)}

\begin{figure}[h!]
\begin{center}
\includegraphics[scale=0.1]{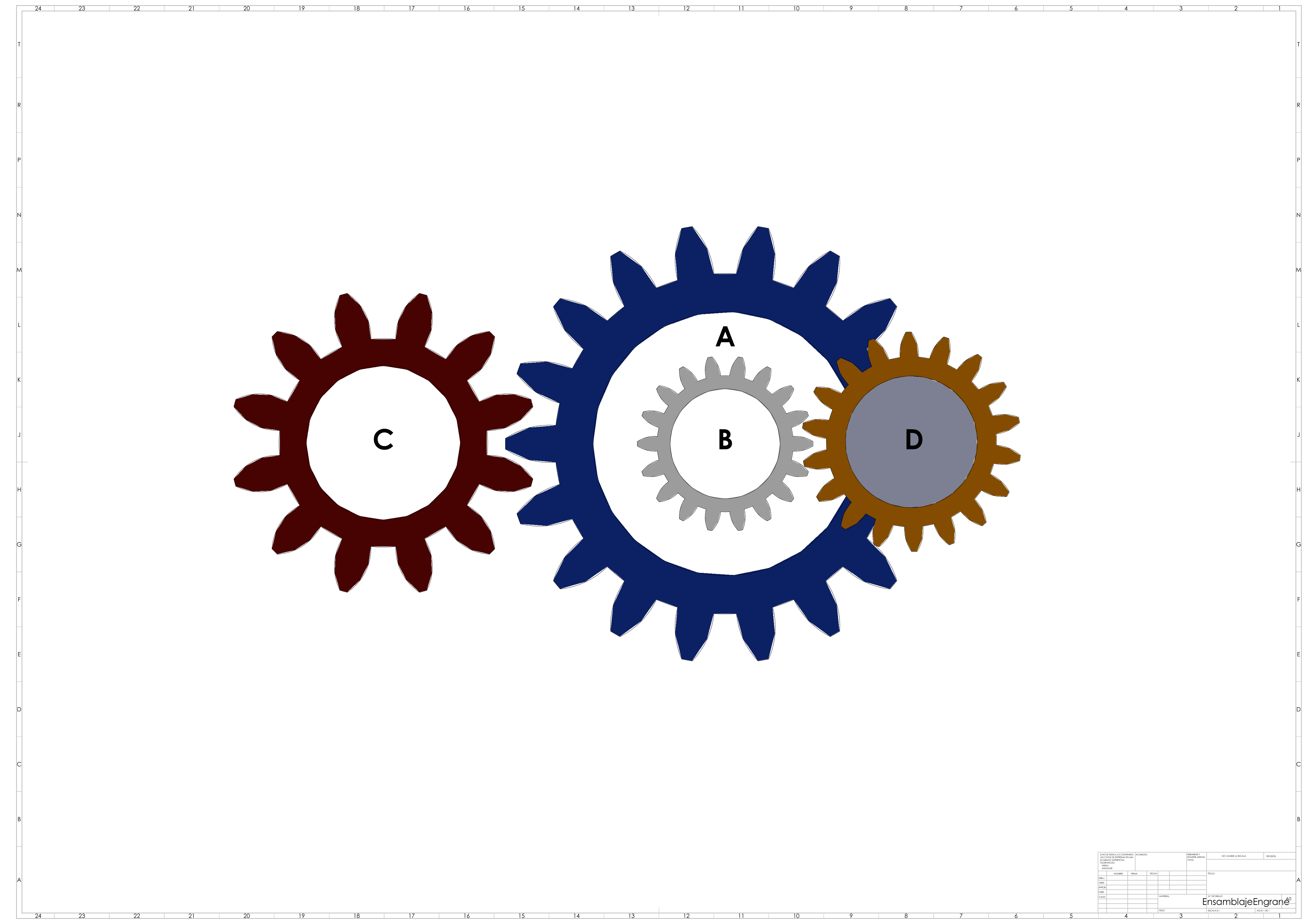}
\caption{Gear train}
\label{fig:Grain-train}
\end{center}
\end{figure}

The GTD problem was defined in \citep{sandgren1990nonlinear} and has no restrictions other than being discrete and that each parameter is within a possible range of values. In this problem, we have $ 4 $ decision parameters, $ y_1 $, $ y_2 $, $ y_3 $, and $ y_4 $, which represent the number of teeth that a gear can have (Fig. \ref{fig:Grain-train}).

The goal of this problem is to determine the minimum cost of gear ratio that can be set as follows:

\begin{equation}
    min\; F(y_1,y_2,y_3,y_4) = \left( \frac{1}{6.931} - \frac{y_2y_3}{y_1y_4} \right)^2
\end{equation}

\noindent where $12 \leq y_i \leq 60$ for $i=(1,2,3,4)$ and each $y_i$ must be integer.

For this case, the result obtained by the CCAA is compared with the results published in \citep{gupta2020modified}, which takes $17$ different meta-heuristic algorithms. Fifty independent runs of the CCAA were made and the best result was taken. In each run, $ 5 $ $smart\_cells $ and $ 4 $ neighbors were used for each $ smart\_cell $, in addition to $ 200 $ evaluations of $F$ to make the setting of the experiment similar to that of the one reported in \citep{gupta2020modified}.

The reference values and the best design obtained by the CCAA are shown in Table \ref{tabla:gear_design}. It can be seen that the CCAA obtained a design with a similar cost to the MSCA and the FA and a better result compared to the other algorithms. These outcomes show that the CCAA is competitive with the best algorithms for this discrete design problem.

\begin{table}[th]
\centering
\scriptsize
\caption{\label{tabla:gear_design}Comparison of meta-heuristics for the GTD problem} 
\begin{tabular}{lccccc}
\hline
Algorithm & $y_1$ & $y_2$ & $y_3$ & $y_4$ & $F_{min}$ \\ 
\hline
\bf{CCAA} & $49$ & $19$ & $16$ & $43$ & $2.7009e-12$ \\
\bf{MSCA} & $49$ & $16$ & $19$ & $43$ & $2.7009e-12$ \\
\bf{FA} & $43$ & $19$ & $16$ & $49$ & $2.7009e-12$ \\
\bf{m-SCA} & $34$ & $20$ & $13$ & $53$ & $2.3078e-11$ \\
\bf{TLBO} & $53$ & $30$ & $13$ & $51$ & $2.3078e-11$ \\
\bf{SCA-GWO} & $51$ & $26$ & $15$ & $53$ & $2.3078e-11$ \\
\bf{CMA-ES} & $53$ & $13$ & $20$ & $34$ & $2.3078e-11$ \\
\bf{wPSO} & $57$ & $12$ & $37$ & $54$ & $8.8876e-10$ \\
\bf{ISCA} & $46$ & $26$ & $12$ & $47$ & $9.9216e-10$ \\
\bf{PSO} & $47$ & $13$ & $12$ & $23$ & $9.9216e-10$ \\
\bf{GWO} & $23$ & $13$ & $12$ & $47$ & $9.9216e-10$ \\
\bf{SCA-PSO} & $46$ & $24$ & $13$ & $47$ & $9.9216e-10$ \\
\bf{SCA} & $60$ & $17$ & $28$ & $55$ & $1.3616e-09$ \\
\bf{SinDE} & $30$ & $17$ & $14$ & $55$ & $1.3616e-09$ \\
\bf{modSCA} & $60$ & $18$ & $25$ & $52$ & $2.3576e-09$ \\
\bf{SSA} & $57$ & $27$ & $17$ & $60$ & $3.6358e-09$ \\
\bf{OBSCA} & $43$ & $12$ & $31$ & $60$ & $8.7008e-09$ \\
\hline
\end{tabular}
\end{table}

\subsubsection{Pressure vessel design (PVD)}

For this problem, a mathematical model calculates the total cost of a cylindrical pressure vessel, showing the cost relationships concerning the material, the structure, and the weld. The design variables to specify are the thickness of the hull ($ y_1 $) and the head ($ y_2 $), the radius of the vessel ($ y_3 $), and the length of the vessel without counting the head ($ y_4 $), to minimize the total cost of the design \citep{gandomi2013cuckoo}. The cost function and its restrictions are shown in Eq. \ref{eq:PVD}.

\begin{figure}[h!]
\begin{center}
\includegraphics[scale=0.3]{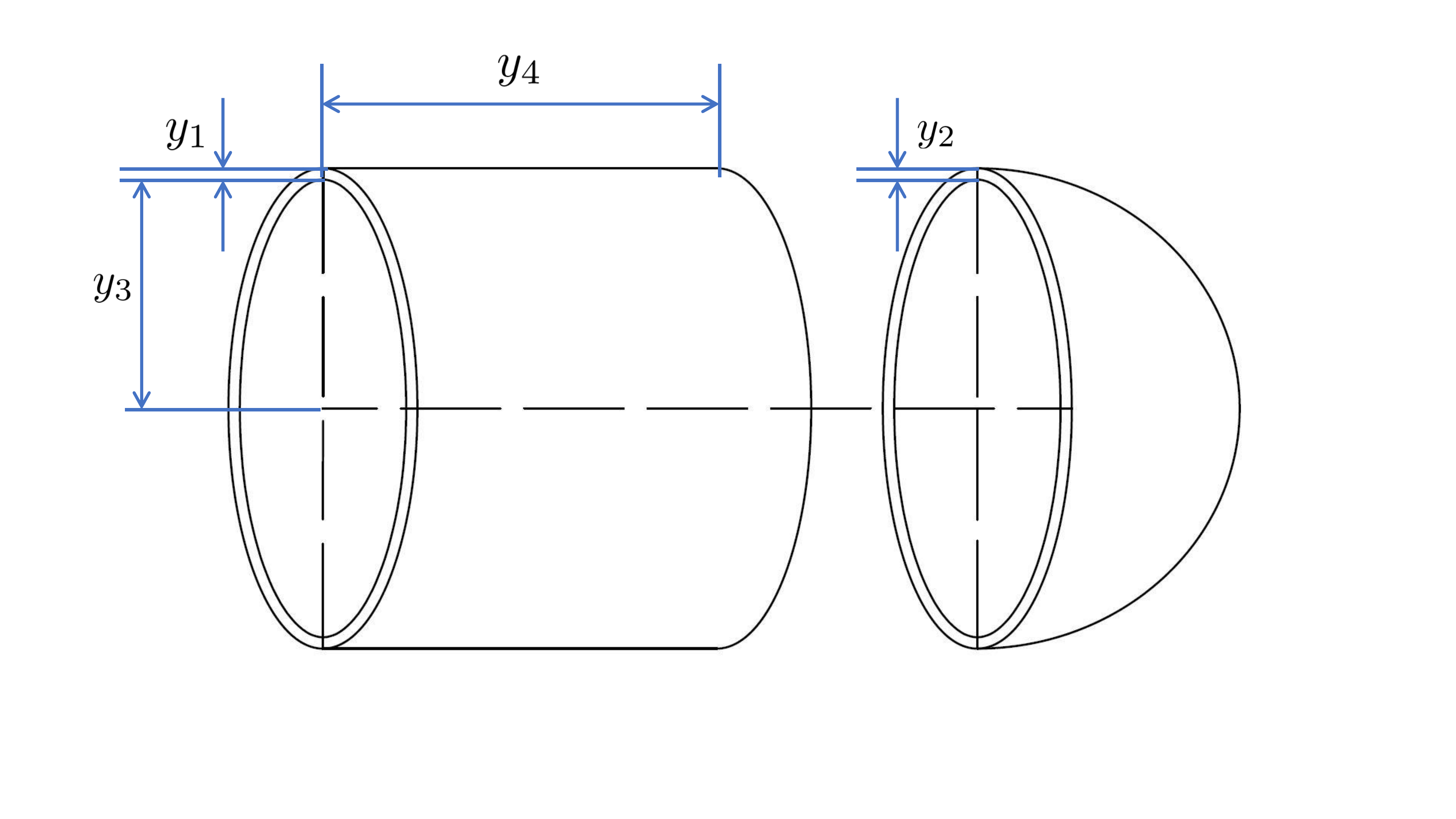}
\caption{Pressure vessel}
\label{fig:Pressure_Vessel}
\end{center}
\end{figure}

\begin{equation}
\label{eq:PVD}
\begin{array}{l}
    min\; F(y_1,y_2,y_3,y_4) = 0.6224y_1y_3y_4 + 1.7781y_1^2y_3 + 3.1661y_1^2y_4 + 19.84y_1^2y_3 \\
    \mbox{Subject to:} \\
    -y_1+0.0193y_3 \leq 0 \\
    -y_3+0.00954y_3 \leq 0 \\
    -\pi y_3^2y_4 - \frac{4}{3} \pi y_3^3 + 1296000 \leq 0 \\
    y_4 - 240 \leq 0
\end{array}
\end{equation}

\noindent where $ 0 \leq y_1, y_2 \leq 99 $ and $ 10 \leq y_3, y_4 \leq 200 $. According to \citep{gandomi2013cuckoo}, the thicknesses $y_1$ and $y_2$ must be multiples of 0.0625 inches because the steel gauges are commercial ones. For this model, the results obtained by the CCAA is compared with the results published in \citep{chen2020efficient}, which considers $ 7 $ different meta-heuristic algorithms. Notice that the best result reported in \citep{chen2020efficient} does not hold the steel gauge restriction for $y_1$ and $y_2$. Therefore, for this case, two different experiments were conducted using the CCAA: one with no gauge restrictions over $y_1$ and $y_2$ and the second taking into account that $y_1$ and $y_2$ are multiples of $0.0625$. Again, $ 50 $ independent runs of the CCAA were made for each experiment and the best result was taken in every case. In each run, $ 6 $ $smart\_cells $ and $ 10 $ neighbors were used for each $smart\_cell $, in addition to $ 15000 $ evaluations of $ F $ to make an experiment similar to the one reported in \citep{gandomi2013cuckoo} and \citep{chen2020efficient}.

The reference values and the best design obtained by the CCAA are shown in Table \ref{tabla:PVD}. It can be seen that the CCAA obtained the best design with a cost of $5885.5328$ in the case of  no steel-gauge restriction. The CCAA also calculated the best design with a cost of $6059.7144$, applying the steel-gauge restriction. Therefore, the CCAA can be very useful for the PVD problem.

\begin{table}[th]
\centering
\scriptsize
\caption{\label{tabla:PVD}Comparison of meta-heuristics for the PVD problem} 
\begin{tabular}{lccccc}
\hline
Algorithm & $y_1$ & $y_2$ & $y_3$ & $y_4$ & $F_{min}$ \\ 
\hline
\bf{CCAA - no gauge restriction} & $0.7781858$ & $0.3846655$ & $40.31985$ & $199.9995$ & $5885.5328$ \\
\bf{RDWOA} & $0.793769$ & $0.39236$ & $41.127973$ & $189.045124$ & $5912.53868$ \\
\bf{CCAA - gauge restriction} & $0.8125$ & $0.4375$ & $42.09845$ & $176.6366$ & $6059.7144$ \\
\bf{ES} & $0.8125$ & $0.4375$ & $42.098087$ & $176.640518$ & $6059.7456$ \\
\bf{PSO} & $0.8125$ & $0.4375$ & $42.091266$ & $176.7465$ & $6061.0777$ \\
\bf{GA} & $0.9375$ & $0.5$ & $48.329$ & $112.679$ & $6410.3811$ \\
\bf{IHS} & $1.125$ & $0.625$ & $58.29015$ & $43.69268$ & $7197.73$ \\
\bf{Lagrangian multiplier} & $1.125$ & $0.625$ & $58.291$ & $43.69$ & $7198.0428$ \\
\bf{Branch-and-bound} & $1.125$ & $0.625$ & $47.7$ & $117.71$ & $8129.1036$ \\
\hline
\end{tabular}
\end{table}


\subsubsection{Welded beam design (WBD)}

The objective of the WBD problem is to obtain the minimum manufacturing cost \citep{coello2000use} subject to the constants of shear stress $\tau $, bending stress $ \sigma $, buckling load $ \gamma $, and deflection $ \delta $ on the beam. This model has $4$ design variables $ \mathbf{y} = (y_1, y_2, y_3, y_4) $ and $7$ restrictions. The variables are the thickness of welds $ y_1 $, the length of clamped bar $ y_2 $, the height of the beam $ y_3 $, and the thickness of the beam $y_4$. The optimization model is presented in Eq. \ref{eq:WBD}.

\begin{figure}[h!]
\begin{center}
\includegraphics[scale=0.3]{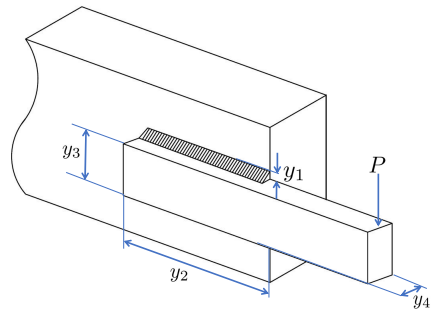}
\caption{Welded beam}
\label{fig:Welded_beam}
\end{center}
\end{figure}

\begin{equation}
\small
\label{eq:WBD}
\begin{array}{l}
    min\; F(y_1,y_2,y_3,y_4) = 1.10471y_1^2 + 0.04811 y_3y_4 (14.0+y_2) \\
    \mbox{Subject to:} \\
    g1(\mathbf{y}) = \tau(\mathbf{y}) - \tau_{max} \leq 0 \\
    g2(\mathbf{y}) = \sigma(\mathbf{y}) - \sigma_{max} \leq 0 \\
    g3(\mathbf{y}) = \delta(\mathbf{y}) - \delta_{max} \leq 0 \\
    g4(\mathbf{y}) = y_1 - y_4 \leq 0 \\
    g5(\mathbf{y}) = P - \gamma(\mathbf{y}) \leq 0 \\
    g6(\mathbf{y}) = 0.125 - y_1 \leq 0 \\
    g7(\mathbf{y}) = 0.10471y_1^2 + 0.04811 y_3y_4 (14.0+y_2)-5.0 \leq 0 \\
    \mbox{where:} \\
    P=6000\; lb,\; L=14\; in,\; \sigma_{max}= 30000\; psi, \\
    \delta_{max}=0.25\; in,\; E=30e6\; psi,\; G=12e6\; psi \\
    M=P(L+\frac{y_2}{2}), \; R=\sqrt{\frac{y_2^2}{4} + \left( \frac{y_1+y_3}{2} \right)^2} \\
    J=2 \left( \sqrt{2}y_1y_2\left( \frac{y_2^2}{12} + \left( \frac{y_1+y_3}{2} \right)^2 \right) \right) \\
    \tau'=\frac{P}{\sqrt{2}y_1y_2}, \;\tau''=\frac{MR}{J} \\
    \tau(\mathbf{y}) = \sqrt{(\tau')^2 + 2 \tau' \tau'' \frac{y_2}{2R} + (\tau'')^2},\; \sigma(\mathbf{y}) = \frac{6PL}{y_3^2y_4} \\
    \delta(\mathbf{y}) = \frac{4PL^3}{Ey_3^3y_4}, \; \gamma(\mathbf{y}) = \frac{4.013E \sqrt{\frac{y_3^2 y_4^6}{36}}}{L^2} \left(1-\frac{y_3}{2L} \sqrt{\frac{E}{4G}} \right) \\
\end{array}
\end{equation}

\noindent where $ 0.1 \leq y_1 \leq 2 $, $ 0.1 \leq y_2 \leq 10 $, $ 0.1 \leq y_3 \leq 10 $ y $ 0.1 \leq y_4 \leq 2 $. For this model, the result obtained by the CCAA is compared with the results published in \citep{chen2020efficient} and \citep{coello2000use}, which take into consideration $ 8 $ different meta-heuristic algorithms. Again, $ 50 $ independent runs of the CCAA were made and the best result was taken. In each run, $ 5 $ $smart\_cells$ and $ 4 $ neighbors were used for each $smart\_cell$, in addition to $ 2000 $ evaluations of $ F $ in order to make the experiment comparable to the references consulted. The best design obtained by the CCAA are shown in Table \ref{tabla:WBD}. It can be seen that the CCAA calculated the second-best design at the cost of $ 1.7248$. Therefore, the CCAA is competitive in the optimization of the WBD problem.

\begin{table}[th]
\centering
\scriptsize
\caption{\label{tabla:WBD}Comparison of meta-heuristics for the WBD problem.} 
\begin{tabular}{lccccc}
\hline
Algorithm & $y_1$ & $y_2$ & $y_3$ & $y_4$ & $F_{min}$ \\ 
\hline
\bf{RDWOA} & $0.205618$ & $3.252958$ & $9.04447$ & $0.20569$ & $1.6961$ \\
\bf{CCAA} & $0.20573$ & $3.4705$ & $9.03662$ & $0.20573$ & $1.7248$ \\
\bf{IHS} & $0.20573$ & $3.47049$ & $9.03662$ & $0.20573$ & $1.7248$ \\
\bf{RO} & $0.203687$ & $3.528467$ & $9.004233$ & $0.207241$ & $1.735344$ \\
\bf{GA-Coello} & $0.2088$ & $3.4205$ & $8.9975$ & $0.2100$ & $1.7483$ \\
\bf{HS} & $0.2442$ & $6.2231$ & $8.2915$ & $0.2433$ & $2.3807$ \\
\bf{David} & $0.2434$ & $6.2552$ & $8.2915$ & $0.2444$ & $2.3841$ \\
\bf{Simple} & $0.2792$ & $5.6256$ & $7.7512$ & $0.2796$ & $2.5307$ \\
\bf{Random} & $0.4575$ & $4.7313$ & $5.0853$ & $0.6600$ & $4.1185$ \\
\hline
\end{tabular}
\end{table}

\subsubsection{Cantilever beam design (CBD)}

\begin{figure}[h!]
\begin{center}
\includegraphics[scale=0.1]{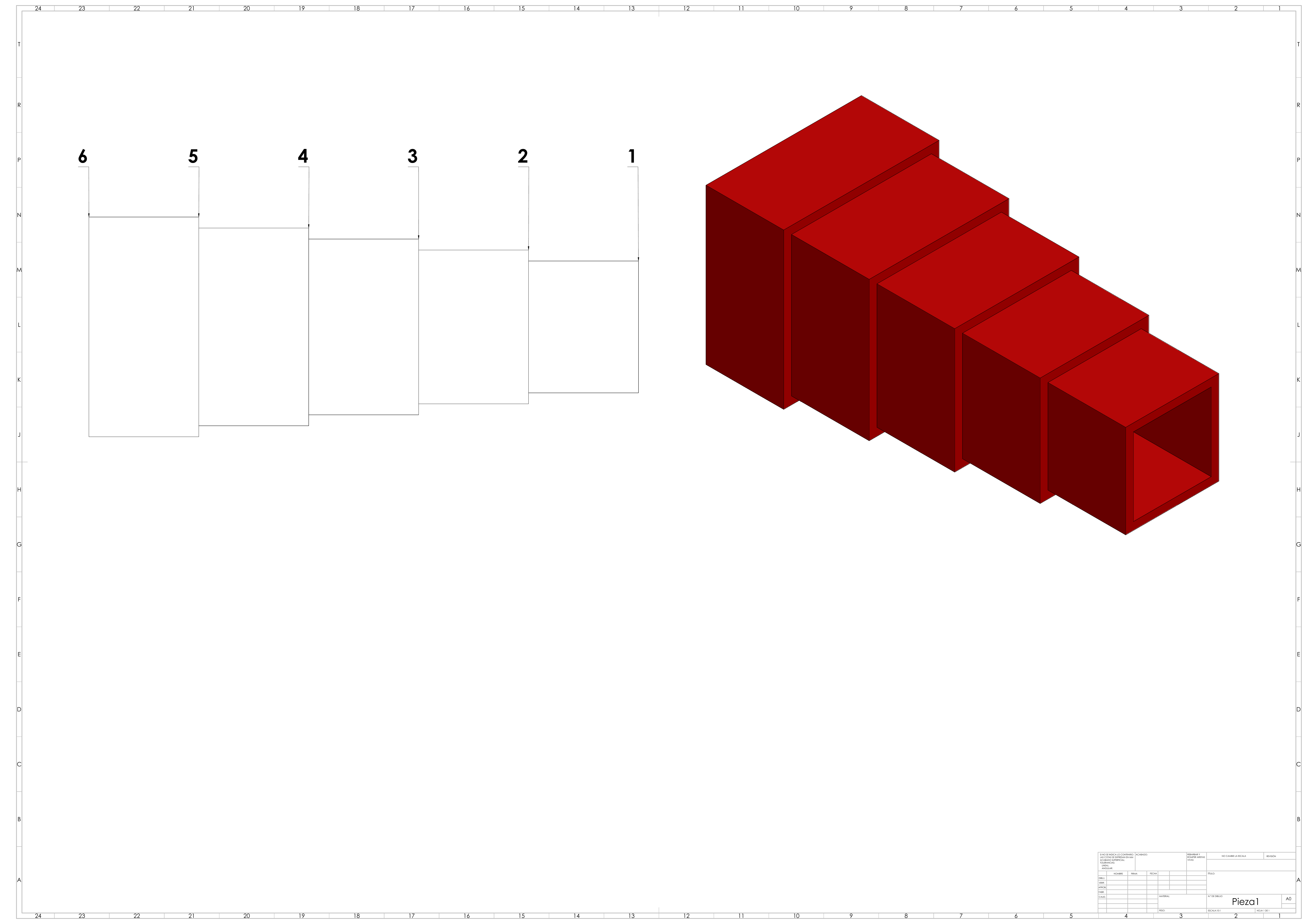}
\caption{Cantilever beam.}
\label{fig:Cantilever}
\end{center}
\end{figure}

The cantilever beam design problem (CBD) objective is to obtain the minimum weight of a rigid structural element supported only on one side by another vertical element \citep{mirjalili2015ant}. The model consists of five hollow elements that have a square section. There is a vertical load in the last part of the structure, and there is a vertical displacement restriction. The system to be designed can be seen in Fig. \ref{fig:Cantilever}. 

The problem is formulated in Eq. \ref{eq:CAN}.

\begin{equation}
\small
\label{eq:CAN}
\begin{array}{l}
    min\; F(y_1,y_2,y_3,y_4,y_5) = 0.6224(y_1 + y_2 + y_3 + y_4 + y_5) \\
    \mbox{Subject to:} \\
    g(y_1,y_2,y_3,y_4,y_5) = \frac{61}{y_1^3} +\frac{37}{y_2^3} + \frac{19}{y_3^3}+ \frac{7}{y_4^3}+ \frac{1}{y_5^3} \leq 1 \\
\end{array}
\end{equation}

\noindent where $ 0.01 \leq y_1, y_2, y_3, y_4, y_5 \leq 100 $. For this model, the result obtained by the CCAA is compared with the results published in \citep{wu2020improved} and \citep{mirjalili2015ant} taking $ 7 $ different meta-heuristic algorithms. Again, $ 50$ independent runs of the CCAA were made and the best result was taken. In each run, $ 5 $ $smart\_cells $ and $ 4 $ neighbors were used for each $ smart\_cell $, with a maximum of $ 12000 $ evaluations of $ F $ to have an experiment similar to the one reported in \citep{mirjalili2015ant}.

\begin{table}[th]
\centering
\scriptsize
\caption{\label{tabla:CAN}Comparison of meta-heuristics for the CBD problem.} 
\begin{tabular}{lcccccc}
\hline
Algorithm & $y_1$ & $y_2$ & $y_3$ & $y_4$ & $y_5$ & $F_{min}$ \\ 
\hline
\bf{ALO} & $6.01812$ & $5.31142$ & $4.48836$ & $3.49751$ & $2.158329$ & $1.33995$ \\
\bf{CCAA} & $6.01698$ & $5.30917$ & $4.49428$ & $3.50147$ & $2.15266$ & $1.33996$ \\
\bf{SOS} & $6.01878$ & $5.30344$ & $4.49587$ & $3.49896$ & $2.15564$ & $1.33996$ \\
\bf{CS} & $6.0089$ & $5.3049$ & $4.5023$ & $3.5077$ & $2.1504$ & $1.33999$ \\
\bf{MMA} & $6.0100$ & $5.3000$ & $4.4900$ & $3.4900$ & $2.1500$ & $1.3400$ \\
\bf{GCA\_I} & $6.0100$ & $5.30400$ & $4.4900$ & $3.4980$ & $2.1500$ & $1.3400$ \\
\bf{GCA\_II} & $6.0100$ & $5.3000$ & $4.4900$ & $3.4900$ & $2.1500$ & $1.3400$ \\
\hline
\end{tabular}
\end{table}

The reference values and the best design achieved by the CCAA are shown in Table \ref{tabla:CAN}. It can be seen that the CCAA had the second-best design with a cost of $ 1.33996 $. Therefore, the CCAA is also well-suited to optimize this engineering problem.

\section{IIR Filter Design Optimization}

IIR filters are one of the main types of filters applied in digital signal processing to separate overlapping signals and restore signals that have been distorted \citep{harris2004multirate}. This section reports the design of adaptive IIR filters using the CCAA algorithm and comparing the results with other specialized algorithms recently published.

The design of infinite impulse response (IIR) filters has become a recurring research topic since they are more efficient than other types of filters. The IIR filter design problem has been addressed using meta-heuristic algorithms, determining the parameters to obtain an optimal model of the unknown plant, minimizing the error's cost function. The IIR filter transfer function is defined by:

\begin{equation}\label{Eq:FiltroIIR}
    \frac{Y(z)}{U(z)} = \frac{b_0+b_1z^{-1}+b_2z^{-2}+\hdots +b_nz^{-n}}{1+a_1z^{-1}+a_2z^{-2}+\hdots+ a_mz^{-m}}
\end{equation}

\noindent where $ Y (z) $ and $ U (z) $ are the output and input respectively, $ [a_1, a_2, \hdots a_n] $ and  $ [b_0, b_1,$ $ \hdots, b_n] $ are the real coefficients of the numerator and denominator, $ n $ and $ m $ are the maximum degree of the polynomials of the numerator and denominator respectively.

\begin{figure}[h!]
    \centering
    \includegraphics[scale=0.5]{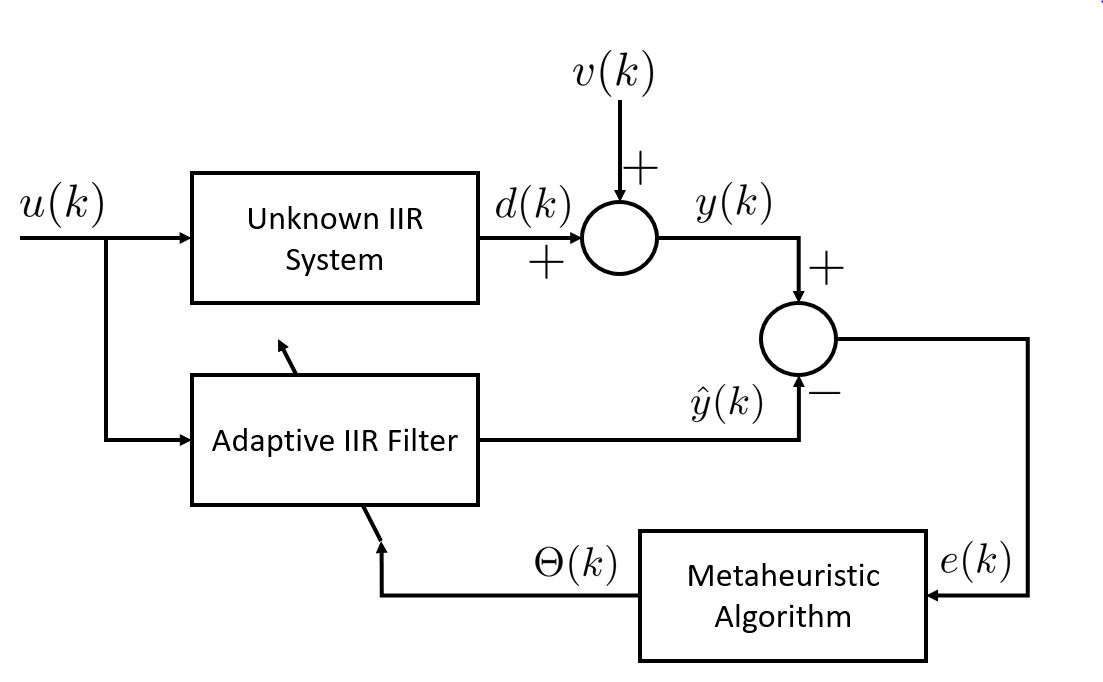}
    \caption{Parameter identification scheme in adaptive IIR filters.}
    \label{fig:EsquemaFiltroIIRAdaptive}
\end{figure}

According to Fig. \ref{fig:EsquemaFiltroIIRAdaptive}, the problem of the adaptive IIR filter is to identify the unknown model whose transfer function is described by Eq. \ref{Eq:FiltroIIR}, $ d (k) $ is the output of the unknown system, and a noise component $ v (k) $ is added to generate the output $ y (k) $. The adaptive IIR filter block receives the same input as the unknown IIR system block $ u (k) $. It also receives as input the coefficients of the numerator and denominator generated by the meta-heuristic algorithm $ \Theta (k ) = [a_1, a_2. \hdots a_m, b_0, b_1, \hdots, b_n] $, the output generated is an estimated response $ \hat{y}(k) $. The deviation of the estimated signal $ \hat{y}(k) $ to the output signal of the unknown system $y(k)$ is known as the mean square error MSE and is represented by $ e (k) $. Mathematically, the adaptive IIR filter problem can be modeled as an optimization problem of minimizing the signal $e (k) $, that is:

\begin{equation}
   Min \; e(k) = \frac{1}{L}\sum_{k=1}^L (y(k)-\hat{y}(k))^2
\end{equation}

\noindent where $L$ is the length of the vector or input signal $ u (k) $.

The $10$ cases analyzed in this work are IIR filter models reviewed in the references \citep{krusienski2005design}, \citep{panda2011iir}, \citep{jiang2015new}, \citep{Lagos-Eulogio2017} and \citep{zhao2019selfish}. Table \ref{Tabla:17TF} shows the models for examples I, II, III, IV, V, VI, VII, VIII, IX, and X. The identification of parameters is carried out through a transfer function of full order.

\begin{table}[h!]
	\centering
	\caption{Ten IIR system identification problems}\label{Tabla:17TF}
	\begin{spacing}{1.7}
		\scalebox{0.75}{
			\begin{tabular}{lll}
				\hline
				Problems & The transfer function & The transfer function of the\\
				&	of IIR plant $H_p(z)$& adaptive IIR filter model $H_a(z)$	\\
				\hline
				Example I&$ \frac{0.05-0.4z^{-1}}{1-1.1314z^{-1}+0.25z^{-2}}$&$\frac{b_0-b_1z^{-1}}{1-a_1z^{-1}-a_2z^{-2}}$ \\
				Example II& $\frac{-0.2-0.4z^{-1}+0.5z^{-2}}{1-0.6z^{-1}+0.25z^{-2}-0.2z^{-3}}$&$\frac{b_0+b_1z^{-1}+b_2z^{-2}}{1-a_1z^{-1}-a_2z^{-2}-a_3z^{-3}}$\\			
				Example III&$\frac{1-0.9z^{-1}+0.81z^{-2}-0.729z^{-3}}{1+0.04z^{-1}+0.2775z^{-2}-0.2101z^{-3}+0.14z^{-4}}$&$\frac{b_0+b_1z^{-1}+b_2z^{-2}+b_3z^{-3}}{1-a_1z^{-1}-a_2z^{-2}-a_3z^{-3}-a_4z^{-4}}$\\
				Example IV &$\frac{0.1084+0.5419z^{-1}+1.0837z^{-2}+1.0837z^{-3}+0.5419z^{-4}+0.1084z^{-5}}{1+0.9853z^{-1}+0.9738z^{-2}-0.3854z^{-3}+0.1112z^{-4}+0.0113z^{-5}}$&$\frac{b_0+b_1z^{-1}+b_2z^{-2}+b_3z^{-3}+b_4z^{-4}+b_5z^{-5}}{1-a_1z^{-1}-a_2z^{-2}-a_3z^{-3}-a_4z^{-4}-a_5z^{-5}}$\\
				Example V &$\frac{1-0.4z^{-2}-0.65z^{-4}+0.26z^{-6}}{1-0.77z^{-2}-0.8498z^{-4}+0.6486z^{-6}}$&$\frac{b_0+b_2z^{-2}+b_4z^{-4}+b_6z^{-6}}{1-a_2z^{-2}-a_4z^{-4}-a_6z^{-6}}$\\
				Example VI&$\frac{1}{1-1.4z^{-1}+0.49z^{-2}}$&$\frac{b_0}{1-a_1z^{-1}-a_2z^{-2}}$\\
				Example VII &$\frac{1}{1-1.2z^{-1}+0.6z^{-2}}$&$\frac{b_0}{1-a_1z^{-1}-a_2z^{-2}}$\\
				Example VIII&$\frac{1.25z^{-1}-0.25z^{-2}}{1-0.3z^{-1}+0.4z^{-2}}$&$\frac{b_1z^{-1}+b_2z^{-2}}{1-a_1z^{-1}-a_2z^{-2}}$\\
				Example IX&$\frac{1}{(1-0.5z^{-1})^3}$&$\frac{b_0}{1-a_1z^{-1}-a_2z^{-2}-a_3z^{-3}}$\\
				Example X&$\frac{-0.3 +0.4z^{-1}-0.5z^{-2}}{1-1.2z^{-1}+0.5z^{-2}-0.1z^{-3}}$&$\frac{b_0+ b_1z^{-1}+b_2z^{-2}}{1-a_1z^{-1}-a_2z^{-2}-a_3z^{-3}}$\\
				\hline
			\end{tabular}
		}
	\end{spacing}
\end{table}

It is necessary to mention that the adaptive IIR filter design problem requires robust algorithms since the systems are non-linear, non-differentiable, and multimodal. Recent hybrid algorithms have been proposed for this problem. For example, in \citep{jiang2015new}, a hybrid algorithm is implemented with a PSO (particle swarm optimization) and a gravitational algorithm (GSA), which is called HPSO-GSA. This algorithm generates a co-evolutionary technique, taking advantage of the PSO's ability to share information between particles and the memory of the best position obtained, and the GSA's force and mass acceleration to determine the best search direction. Similarly, in \citep{Lagos-Eulogio2017} the parameter estimation problem is addressed using a hybrid algorithm, in this case, a combination of concepts of a cellular-automata neighborhood, PSO, and the Differential Evolution algorithm (CPSO-DE). In this way, CPSO-DE is able to modify the trajectory of the particles to leave the local optimum, while the PSO and the DE focus on finding a higher quality solution. The performance of the CCAA algorithm is compared with two algorithms that also use cellular automata concepts (CPSO and CPSO-DE), with a gravitational hybrid algorithm (HPSO-GSA), and with two recent algorithms (GWO and MSCA). To make a more fair comparison, a population of $100 $ individuals was applied for the HPSO-GSA, GWO, and MSCA. For the CPSO, CPSO-DE and CCAA, $ 20 $ $smart\_cells $ and $ 5 $ neighbors for $smart\_cell$  were utilized. Besides, $50$ runs were carried out for every algorithm, each applying a total number of $500$ iterations.

Table \ref{IIR:FullOrder1_5} shows the MSE statistical results for the best, worst, average, mean, and standard deviation for each example for the first $5$ IIR filters. For the cases I and II, the CCAA and CPSO-DE algorithms identified the parameters with an MSE of $0$, obtaining the best of the $50$ runs. For the case III to V, the CPSO-DE algorithm obtained a higher performance than the others for calculating the best parameter estimation. In these cases, the CCAA obtained positions $3$, $3$ and $2$ respectively in the best value, and positions $3$, $2$ and $2$ in the best average value. The best parameter estimations for these $ 5 $ filters are presented in Tables \ref{P_E_E1_C1} - \ref{P_E_E5_C1}.

\newpage

\begin{table}[h!]
\small
\begin{center}
\caption{Experimental results (MSE) of algorithms for Test Examples I,II, III, IV, V}\label{IIR:FullOrder1_5}
	\begin{spacing}{1.2}
		\scalebox{0.8}{
			\begin{tabular}{l|lcccccccc}
				\hline
				Problems & Measure & \multicolumn{6}{l}{Mean square error (MSE) }\\
				\cline{3-8}
				& & HPSO-GSA &CPSO-DE &CPSO &GWO &MSCA &CCAA\\
	\hline
		Example I &Best &2.305e-22 &\textbf{0} &1.249e-23 &1.237e-07 &1.914e-05 &\textbf{0}\\
				&Worst &1.006e-20 &\textbf{0} &2.504e-01 &1.192e-02 &1.124e-01 &\textbf{0}\\
				&Average &3.384e-21 &\textbf{0} &9.334e-03 &2.389e-03 &3.184e-03 &\textbf{0}\\
				&Median &2.524e-21 &\textbf{0} &2.734e-18 &3.807e-06 &2.867e-04 &\textbf{0}\\
				&SD &2.348e-21 &\textbf{0} &4.632e-02 &4.215e-03 &1.592e-02 &\textbf{0}\\
				\hline
				Example II &Best &1.118e-21 &\textbf{0} &1.914e-32 &5.788e-08 &2.865e-04 &\textbf{0}\\
				&Worst &5.902e-03 &\textbf{1.222e-32} &6.121e-03 &4.858e-03 &3.943e-02 &5.589e-03\\
				&Average &1.453e-03 &\textbf{6.079e-34} &1.424e-03 &1.210e-03 &6.084e-03 &1.123e-04\\
				&Median &4.134e-21 &\textbf{0} &4.572e-04 &2.451e-07 &4.250e-03 &\textbf{0}\\
				&SD &2.092e-03 &\textbf{2.221e-33} &1.865e-03 &1.643e-03 &7.301e-03 &7.904e-04\\
				\hline
				Example III &Best &9.300e-21 &\textbf{0} &8.728e-12 &1.242e-05 &1.648e-03 &1.528e-14\\
				&Worst &2.594e-02 &\textbf{1.082e-31} &2.794e+00 &1.026e-01 &3.384e-01 &4.980e-02\\
				&Average &3.288e-03 &\textbf{8.660e-33} &1.270e-01 &1.173e-02 &5.761e-02 &3.321e-03\\
				&Median &1.566e-13 &\textbf{0} &1.206e-03 &1.357e-03 &2.416e-02 &5.887e-05\\
				&SD &7.083e-03 &\textbf{1.852e-32} &4.351e-01 &2.357e-02 &6.200e-02 &9.963e-03\\
				\hline
				Example IV &Best &3.124e-06 &\textbf{1.768e-08} &8.645e-06 &1.139e-05 &1.277e-03 &1.316e-05\\
				&Worst &3.054e-01 &\textbf{5.904e-06} &2.456e+00 &8.225e-03 &5.007e-02 &4.019e-03\\
				&Average &8.206e-03 &\textbf{2.540e-06} &2.037e-01 &1.449e-03 &1.220e-02 &6.618e-04\\
				&Median &3.056e-04 &\textbf{2.715e-06} &1.151e-03 &9.029e-04 &1.091e-02 &2.033e-04\\
				&SD &4.308e-02 &\textbf{1.280e-06} &5.263e-01 &1.760e-03 &9.180e-03 &9.653e-04\\
				\hline
				Example V &Best &5.336e-04 &\textbf{2.964e-31} &5.072e-04 &7.645e-04 &1.811e-03 &2.416e-05\\
				&Worst &3.620e-02 &\textbf{3.139e-09} &2.673e-01 &3.050e-02 &1.015e-01 &1.191e-02\\
				&Average &8.303e-03 &\textbf{6.453e-11} &1.790e-02 &9.688e-03 &1.746e-02 &2.697e-03\\
				&Median &9.395e-03 &\textbf{2.014e-18} &8.639e-03 &9.184e-03 &1.518e-02 &8.339e-04\\
				&SD &6.501e-03 &\textbf{4.437e-10} &4.837e-02 &5.463e-03 &1.471e-02 &3.745e-03\\			
	\hline
			\end{tabular}
		}
	\end{spacing}
\end{center}
\end{table}

\begin{table}[h!]
    \small
	\begin{center}
		\caption{The parameter estimation of algorithms for Example I}\label{P_E_E1_C1}
		\begin{spacing}{1.1}
			\scalebox{0.7}{
				\begin{tabular}{cccccccc}
					\hline
					Parameters &Actual values & \multicolumn{6}{l}{Parameter estimation value}\\
					\cline{3-8}
					& & HPSO-GSA &CPSO-DE &CPSO &GWO &MSCA &CCAA\\
					\hline
					$b_0$ &\textbf{0.050}  &{0.050} &\textbf{0.050} &{0.050} &0.049 &0.043 &\textbf{0.050}\\
					$b_1$ &\textbf{-0.400} &{-0.400} &\textbf{-0.400} &{-0.400} &-0.399 &-0.393 &\textbf{-0.400}\\ 
					$a_1$ &\textbf{1.131} &{1.131} &\textbf{1.131} &{1.131} &1.132 &1.133 &\textbf{1.131}\\
					$a_2$ &\textbf{-0.250} &{-0.250} &\textbf{-0.250} &{-0.250} &-0.251 &-0.2514 &\textbf{-0.250}\\
					\hline
				\end{tabular}
			}
		\end{spacing}
	\end{center}
\end{table}

\newpage

\begin{table}[h!]
    \small
	\begin{center}
		\caption{The parameter estimation of algorithms for Example II }\label{P_E_E2_C1}
		\begin{spacing}{1.1}
			\scalebox{0.7}{
				\begin{tabular}{cccccccccc}
					\hline
					Parameters &Actual values & \multicolumn{6}{l}{Parameter estimation value}\\
					\cline{3-8}
					& & HPSO-GSA &CPSO-DE &CPSO &GWO &MSCA &CCAA\\
					\hline
					$b_0$ &\textbf{-0.20} &{-0.20} &\textbf{-0.20} &{-0.20} &-0.19 &-0.19 &\textbf{-0.20}\\
					$b_1$ &\textbf{-0.40} &{-0.40} &\textbf{-0.40} &{-0.40} &-0.39 &-0.39 &\textbf{-0.40}\\
					$b_2$ &\textbf{0.50} &{0.50} &\textbf{0.50} &{0.50} &0.51 &0.51 &\textbf{0.50}\\
					$a_1$ &\textbf{0.60} &{0.60} &\textbf{0.60} &{0.60} &0.61 &0.64 &\textbf{0.60}\\
					$a_2$ &\textbf{-0.25} &{-0.25} &\textbf{-0.25} &{-0.25} &-0.24 &-0.23 &\textbf{-0.25}\\
					$a_3$ &\textbf{0.20} &{0.20} &\textbf{0.20} &{0.20} &0.21 &0.21 &\textbf{0.20}\\
					\hline
				\end{tabular}
			}
		\end{spacing}
	\end{center}
\end{table}

\begin{table}[h!]
    \small
	\begin{center}
		\caption{The parameter estimation of algorithms for Example III }\label{P_E_E3_C1}
		\begin{spacing}{1.1}
			\scalebox{0.7}{
				\begin{tabular}{cccccccc}
					\hline
					Parameters &Actual values & \multicolumn{6}{l}{Parameter estimation value}\\
					\cline{3-8}
					& & HPSO-GSA &CPSO-DE &CPSO &GWO &MSCA &CCAA\\
					\hline
					$b_0$ &\textbf{1.0000} &{1.0000} &{1.0000} &0.9998 &0.9996 &1.0250 &0.9991\\
					$b_1$ &\textbf{-0.9000} &{-0.9000} &{-0.9000} &-0.9042 &-0.8928 &-0.9313 &-0.8975\\
					$b_2$ &\textbf{0.8100}  &{0.8100} &{0.8100} &0.8093 &0.8071 &0.7354 &0.8154\\
					$b_3$ &\textbf{-0.7290} &{-0.7290} &{-0.7290} &-0.7309 &-0.7283 &-0.7019 &-0.7261\\
					$a_1$ &\textbf{-0.0400} &{-0.0400} &{-0.0400} &-0.0362 &-0.0463 &-0.0031 &-0.0424\\
					$a_2$ &\textbf{-0.2775} &{-0.2775} &{-0.2775} &-0.2733 &-0.2795 &-0.1994 &-0.2855\\
					$a_3$ &\textbf{0.2101} &{0.2101} &{0.2101} &0.2131 &0.2110 &0.2579 &0.2018\\
					$a_4$ &\textbf{-0.1400} &{-0.1400} &{-0.1400} &-0.1395 &-0.1363 &-0.1390 &-0.1442\\
					\hline
				\end{tabular}
			}
		\end{spacing}
	\end{center}
\end{table}

\begin{table}[h!]
    \small
	\begin{center}
		\caption{The parameter estimation of algorithms for Example IV}\label{P_E_E4_C1}
		\begin{spacing}{1.1}
			\scalebox{0.7}{
				\begin{tabular}{cccccccc}
					\hline
					Parameters &Actual values & \multicolumn{6}{l}{Parameter estimation value}\\
					\cline{3-8}
					& & HPSO-GSA &CPSO-DE &CPSO &GWO &MSCA &CCAA\\
					\hline
					$b_0$ &\textbf{0.1084} &0.1083 &{0.1084} &0.1083 &0.1106 &0.1040 &0.1078\\
					$b_1$ &\textbf{0.5419} &0.4900 &0.4609 &0.4536 &0.4676 &0.4798 &0.4332\\
					$b_2$ &\textbf{1.0840} &0.8416 &0.7088 &0.6748 &0.7262 &0.7193 &0.6129\\
					$b_3$ &\textbf{1.0840}  &0.6541 &0.4176 &0.3589 &0.3822 &0.4713 &0.3065\\
					$b_4$ &\textbf{0.5419} &0.1908 &-0.0023 &-0.0484 &-0.1161 &0.03742 &-0.0184\\
					$b_5$ &\textbf{0.1084} &-0.0038 &-0.0675 &-0.0821 &-0.1463 &-0.0045 &-0.0189\\
					$a_1$ &\textbf{-0.9853} &-0.5027 &-0.2384 &-0.1682 &-0.3329 &-0.3064 &-0.0012\\
					$a_2$ &\textbf{-0.9738} &-0.6778 &-0.5157 &-0.4804 &-0.3086 &-0.5344 &-0.5806\\
					$a_3$ &\textbf{-0.3864} &-0.0534 &0.1316 &0.1806 &0.1153 &-0.0100 &0.2203\\
					$a_4$ &\textbf{-0.1112} &-0.0536 &-0.0227 &-0.0183 &0.0952 &-0.0090 &-0.0916\\
					$a_5$ &\textbf{-0.0113} &0.0079 &0.0192 &0.0236 &-0.0006 &-0.0013 &0.0267\\
					\hline
				\end{tabular}
			}
		\end{spacing}
	\end{center}
\end{table}

For cases VI, VII, VIII, IX, and X, the CCAA achieves better performance than the other algorithms (Table \ref{IIR:FullOrder6_10}). In all cases, the CCAA achieves an MSE of $0$ for all statistics, as well as the CPSO algorithm achieves it for cases VII and VIII and CPSO-DE for case IX. 
 
\newpage

\begin{table}[h!]
    \small
	\begin{center}
		\caption{The parameter estimation of algorithms for Example V}\label{P_E_E5_C1}
		\begin{spacing}{1.1}
			\scalebox{0.7}{
				\begin{tabular}{cccccccc}
					\hline
					Parameters &Actual values & \multicolumn{6}{l}{Parameter estimation value}\\
					\cline{3-8}
					& & HPSO-GSA &CPSO-DE &CPSO &GWO &MSCA &CCAA\\
					\hline
					$b_0$ &\textbf{1.0000} &0.9952 &{1.0000} &0.9911 &1.0030 &0.9290 &1.0020\\
					$b_2$ &\textbf{-0.4000} &0.0951 &-0.4016 &0.0310 &0.8110 &0.3474 &0.1533\\
					$b_4$ &\textbf{-0.6500} &-0.4787 &-0.6504 &-0.4870 &-0.2160 &-0.2686 &-0.4490\\
					$b_6$ &\textbf{0.2600} &0.0587 &0.2606 &0.0794 &-0.2203 &-0.0571 &0.0338\\
					$a_2$ &\textbf{0.7700} &0.2884 &0.7716 &0.3439 &-0.4426 &-0.0001 &0.2150\\
					$a_4$ &\textbf{0.8498} &0.8518 &0.8497 &0.8420 &0.8445 &0.8281 &0.8424\\
					$a_6$ &\textbf{-0.6486} &-0.2361 &-0.6499 &-0.2729 &0.4056 &0.0442 &-0.1625\\
					\hline
				\end{tabular}
			}
		\end{spacing}
	\end{center}
\end{table}

	\begin{table}[h!]
	    \footnotesize
		\begin{center}
\caption{Experimental results (MSE) of algorithms for Test Examples VI, VII, VIII, IX, X}
\label{IIR:FullOrder6_10}
			\begin{spacing}{1.2}
				\scalebox{0.8}{
					\begin{tabular}{l|lcccccccc}
						\hline
						Problems & Measure & \multicolumn{6}{l}{Mean square error (MSE) }\\
						\cline{3-8}
						& & HPSO-GSA &CPSO-DE &CPSO &GWO &MSCA &CCAA\\										
						\hline
Example VI &Best &4.779e-22 &\textbf{0} &6.777e-27 &9.614e-07 &5.878e-05 &\textbf{0}\\
							&Worst &2.158e-20 &\textbf{0} &3.560e+00 &2.908e+00 &5.688e-01 &\textbf{0}\\
							&Average &6.638e-21 &\textbf{0} &7.121e-02 &8.731e-02 &1.494e-02 &\textbf{0}\\
							&Median &5.104e-21 &\textbf{0} &1.456e-18 &6.153e-06 &2.239e-03 &\textbf{0}\\
							&SD &5.239e-21 &\textbf{0} &5.034e-01 &4.234e-01 &8.003e-02 &\textbf{0}\\
							\hline
							Example VII &Best &6.540e-23 &\textbf{0} &\textbf{0} &1.156e-07 &6.273e-05 &\textbf{0}\\
							&Worst &3.137e-21 &2.128e-31 &\textbf{0} &5.108e-06 &3.023e-03 &\textbf{0}\\
							&Average &1.157e-21 &8.506e-33 &\textbf{0} &1.523e-06 &7.333e-04 &\textbf{0}\\
							&Median &9.449e-22 &\textbf{0} &\textbf{0} &1.194e-06 &4.279e-04 &\textbf{0}\\
							&SD &7.713e-22 &4.209e-32 &\textbf{0} &1.142e-06 &6.875e-04 &\textbf{0}\\
							\hline
							Example VIII &Best &1.626e-22 &\textbf{0} &\textbf{0} &3.298e-08 &1.997e-05 &\textbf{0}\\
							&Worst &2.581e-21 &4.578e-33 &\textbf{0} &7.673e-03 &8.759e-03 &\textbf{0}\\
							&Average &8.244e-22 &9.156e-35 &\textbf{0} &9.694e-04 &1.005e-03 &\textbf{0}\\
							&Median &6.897e-22 &\textbf{0} &\textbf{0} &2.495e-07 &1.737e-04 &\textbf{0}\\
							&SD &5.130e-22 &6.475e-34 &\textbf{0} &2.432e-03 &2.289e-03 &\textbf{0}\\
							\hline
							Example IX &Best &9.748e-14 &\textbf{0} &1.798e-05 &2.513e-05 &7.443e-04 &\textbf{0}\\
							&Worst &8.902e-02 &\textbf{0} &2.059e-01 &3.162e+00 &1.448e-01 &\textbf{0}\\
							&Average &6.291e-03 &\textbf{0} &4.343e-02 &1.076e-01 &4.313e-02 &\textbf{0}\\
							&Median &6.838e-04 &\textbf{0} &3.058e-02 &2.315e-02 &2.312e-02 &\textbf{0}\\
							&SD &1.461e-02 &\textbf{0} &4.427e-02 &4.426e-01 &4.259e-02 &\textbf{0}\\
							\hline
							Example X &Best &1.772e-21 &\textbf{0} &2.172e-16 &9.734e-07 &2.365e-04 &\textbf{0}\\
							&Worst &9.592e-02 &2.640e-32 &8.775e-02 &8.765e-02 &9.275e-02 &\textbf{0}\\
							&Average &5.284e-03 &2.833e-33 &7.671e-03 &1.015e-02 &1.519e-02 &\textbf{0}\\
							&Median &1.582e-17 &\textbf{0} &1.450e-09 &2.376e-03 &5.322e-03 &\textbf{0}\\
							&SD &2.116e-02 &7.816e-33 &2.154e-02 &1.775e-02 &2.393e-02 &\textbf{0}\\
						\hline
					\end{tabular}
				}
			\end{spacing}
		\end{center}
	\end{table}

\begin{figure}[h!]
	\caption{Convergence of the MSE curve for the IIR filters }\label{fig:convergencia_filtros}
	\centering				
	\subfigure[Example I]{\includegraphics[scale=0.31]{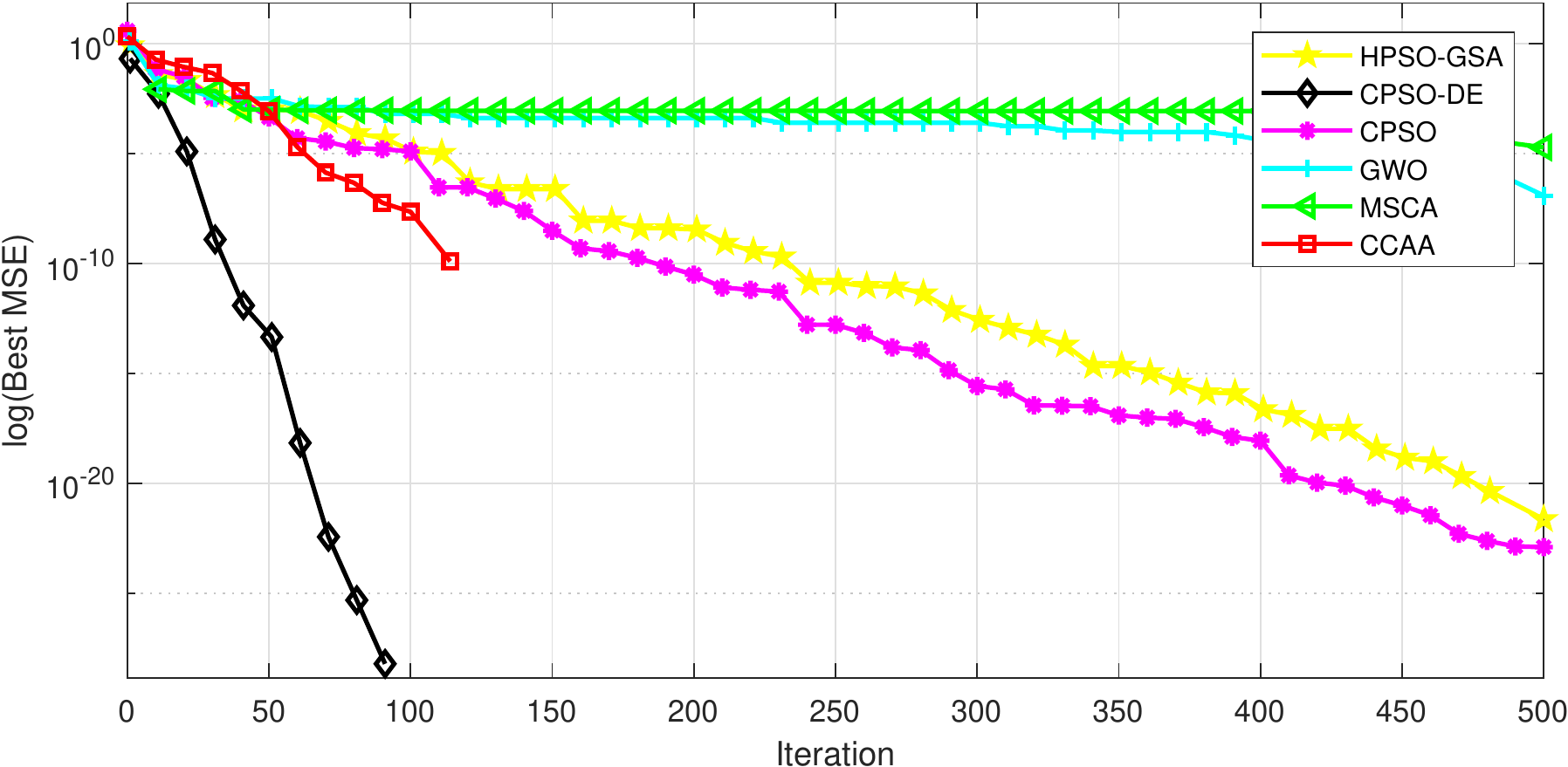} }\vspace{00mm}
	\subfigure[Example II]{\includegraphics[scale=0.31]{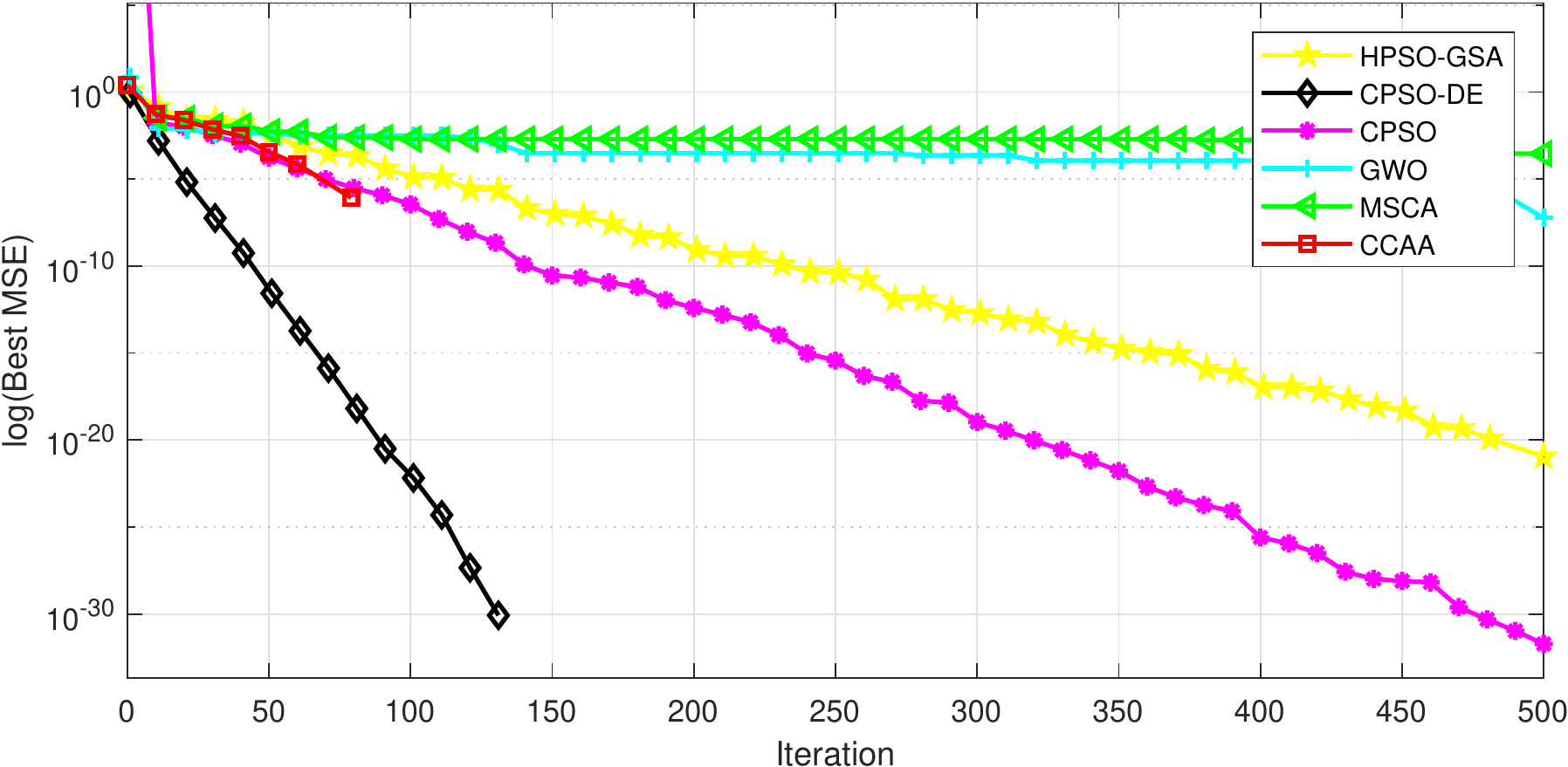}}\hspace{10mm}
	\subfigure[Example III]{\includegraphics[scale=0.31]{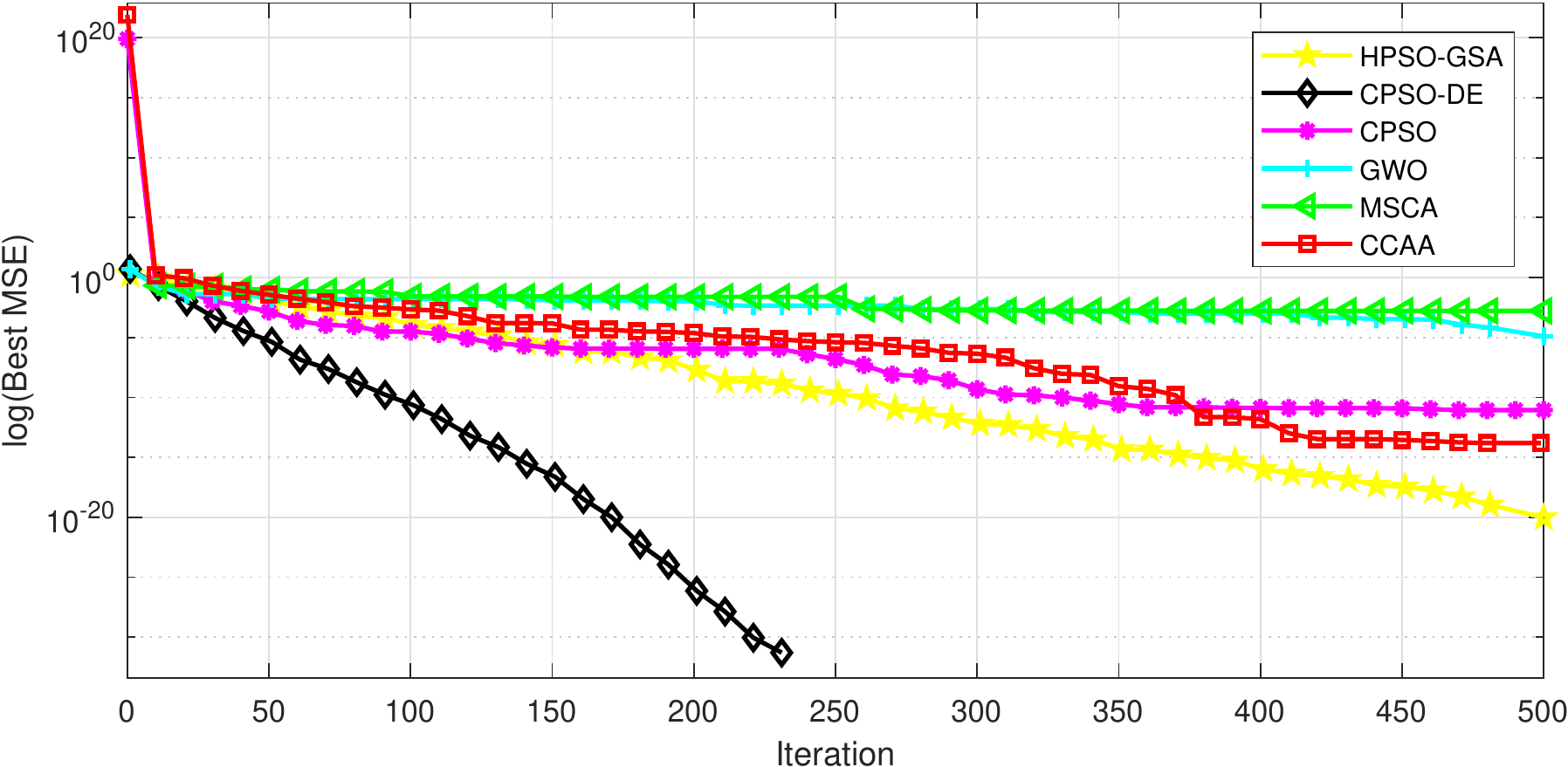} }\vspace{00mm}
	\subfigure[Example IV]{\includegraphics[scale=0.31]{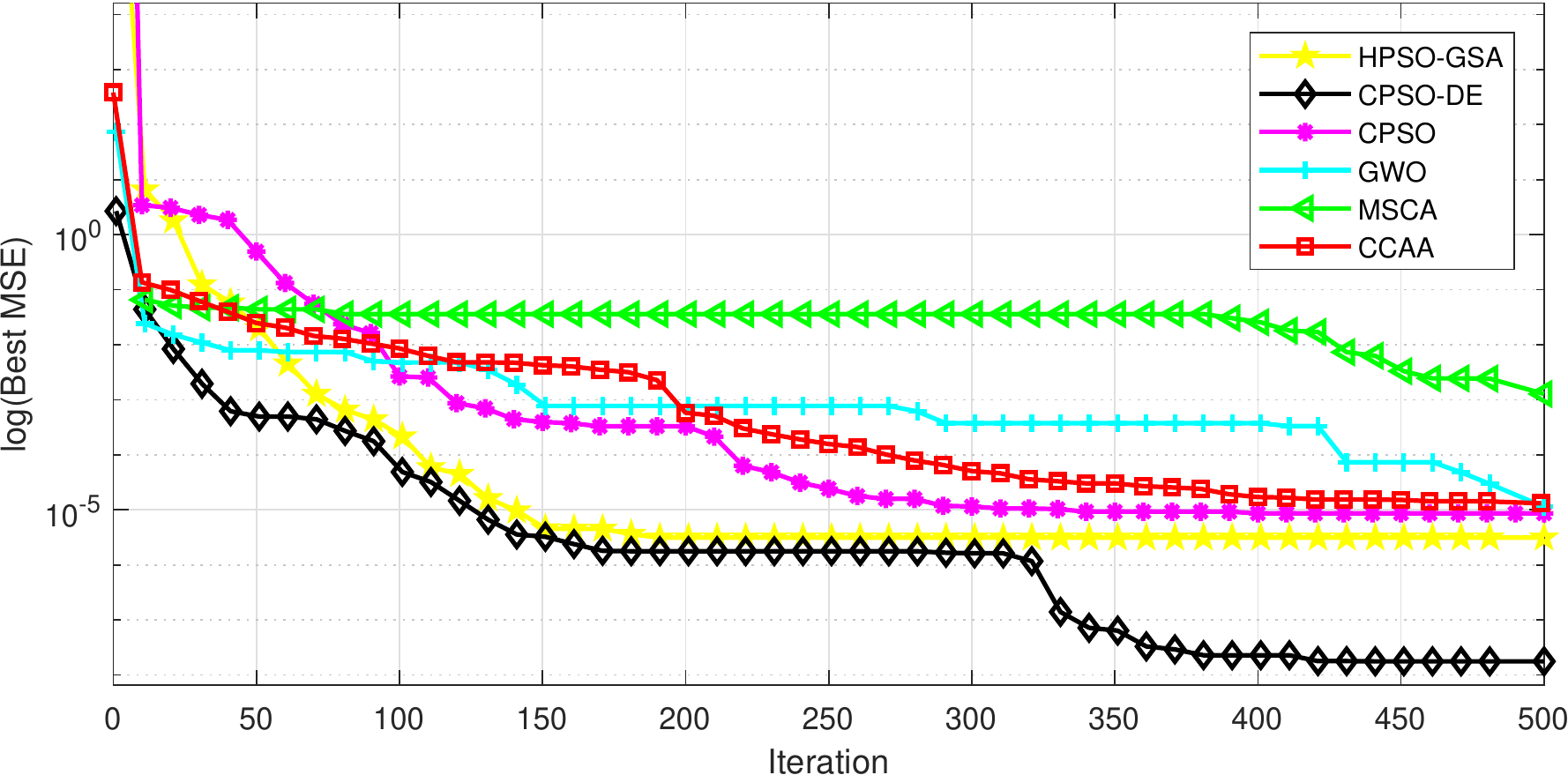}}\hspace{10mm}
	\subfigure[Example V]{\includegraphics[scale=0.31]{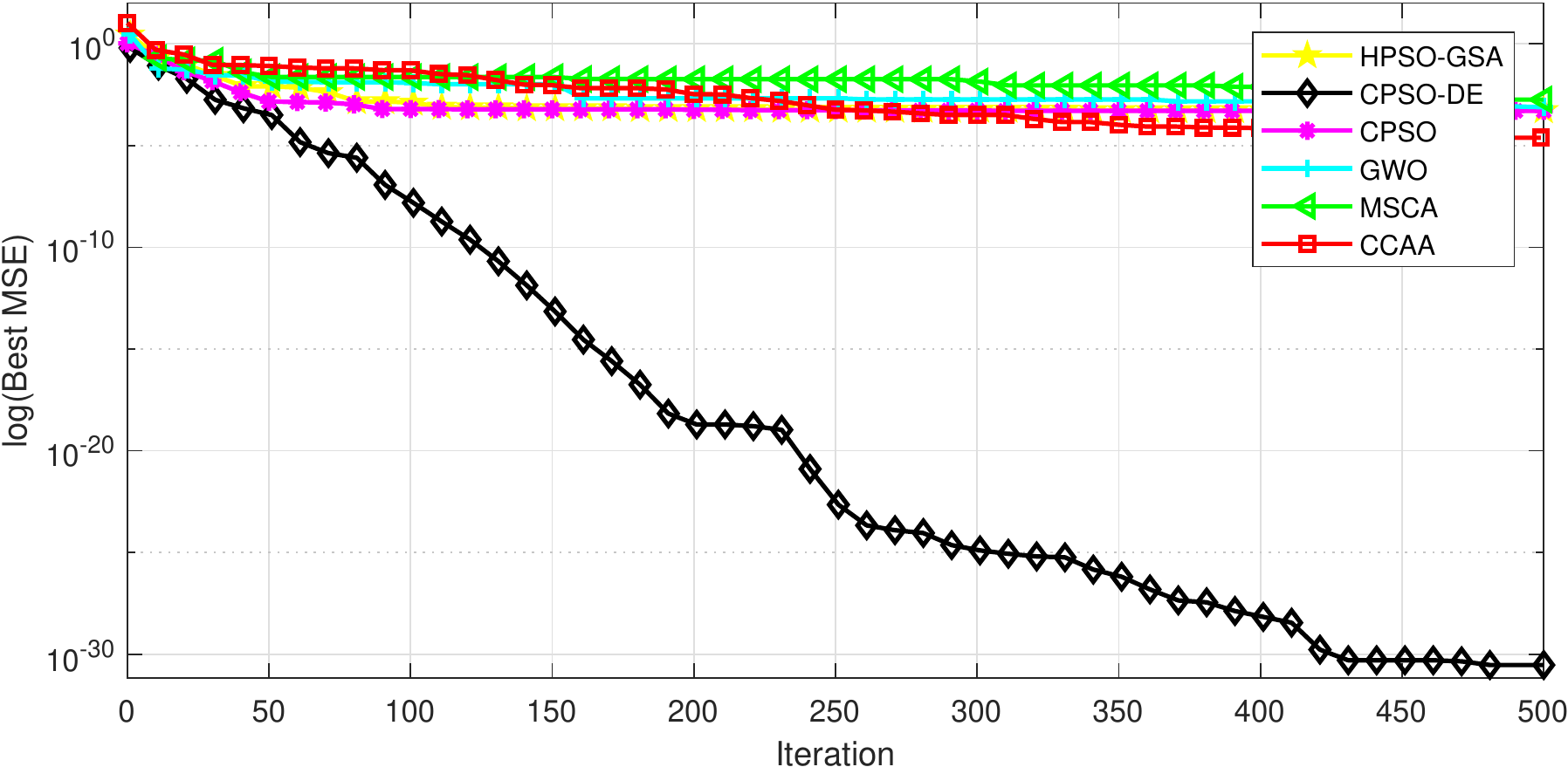} }\vspace{00mm}		
	\subfigure[Example VI]{\includegraphics[scale=0.31]{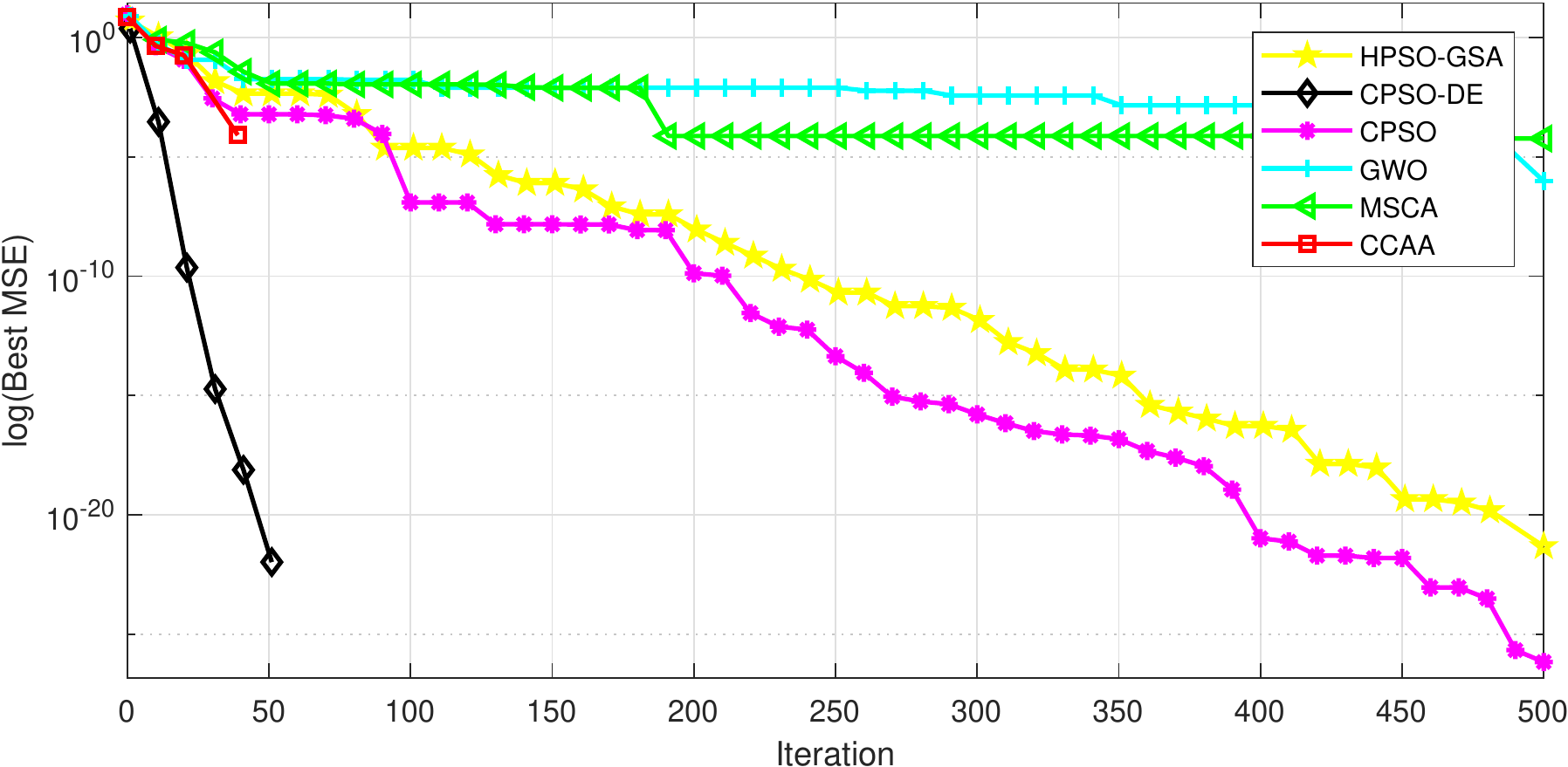}}\hspace{10mm}
	\subfigure[Example VII]{\includegraphics[scale=0.31]{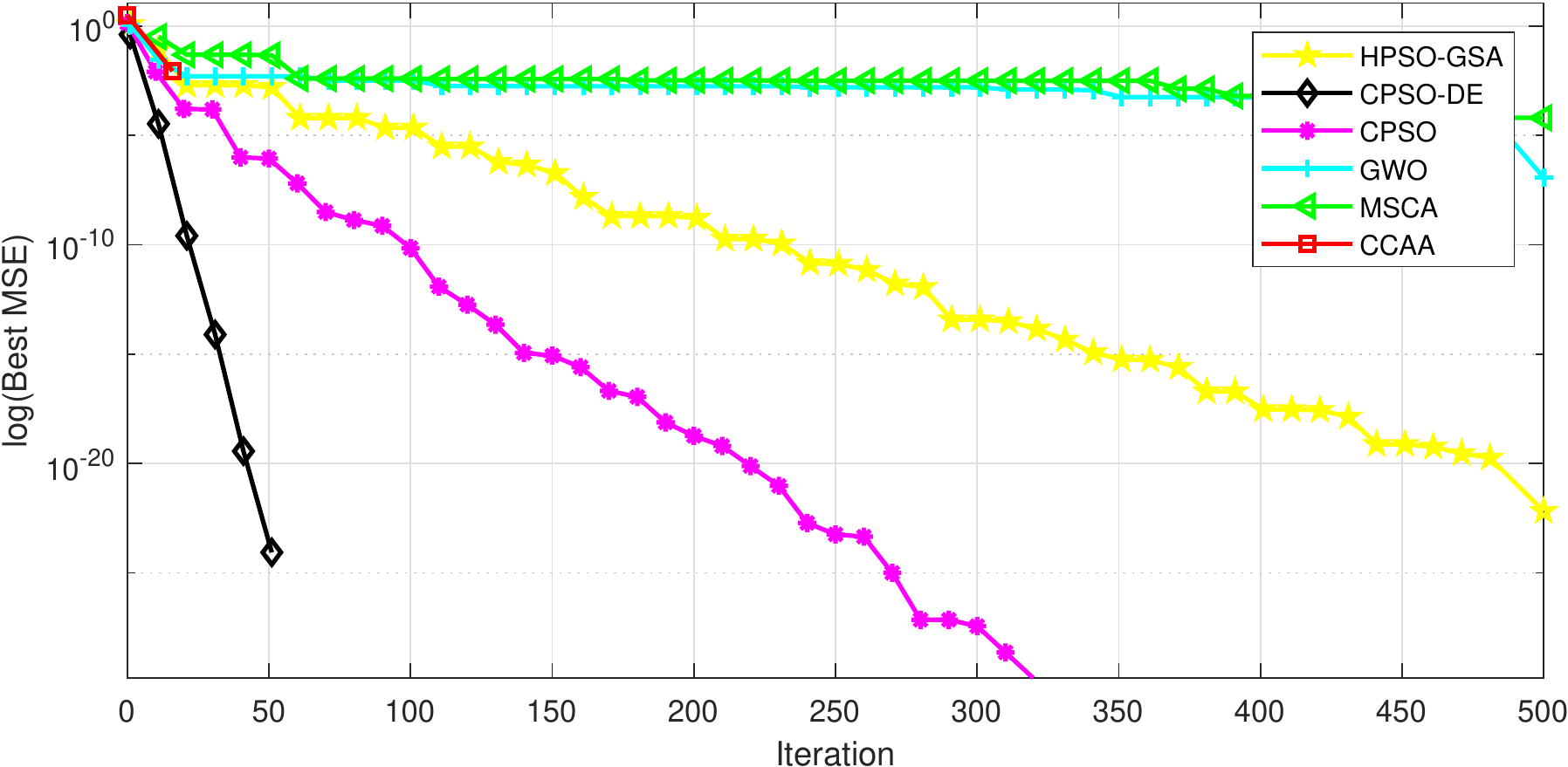} }\vspace{00mm}		
	\subfigure[Example VIII]{\includegraphics[scale=0.31]{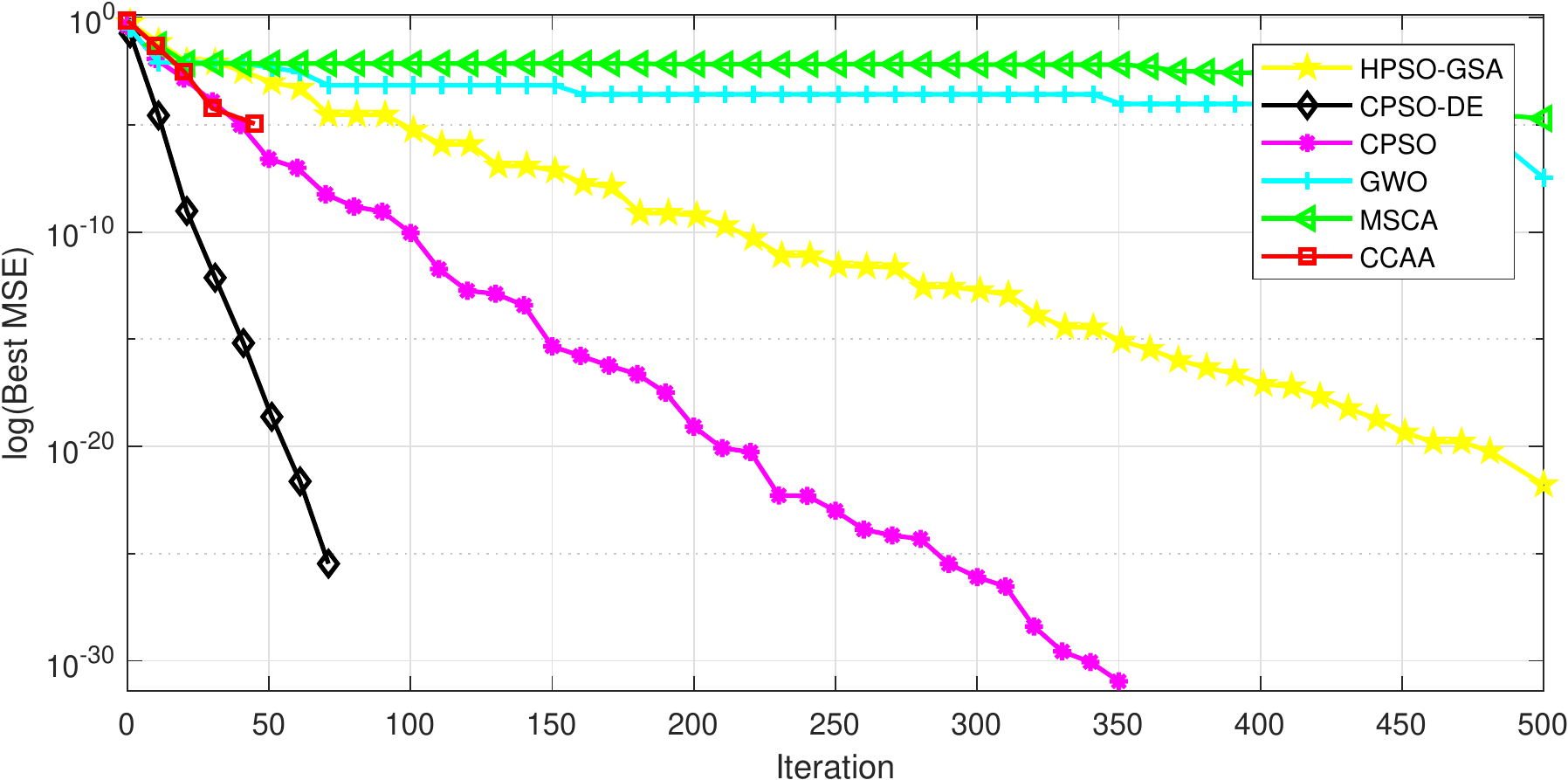}}\hspace{10mm}
	\subfigure[Example IX]{\includegraphics[scale=0.31]{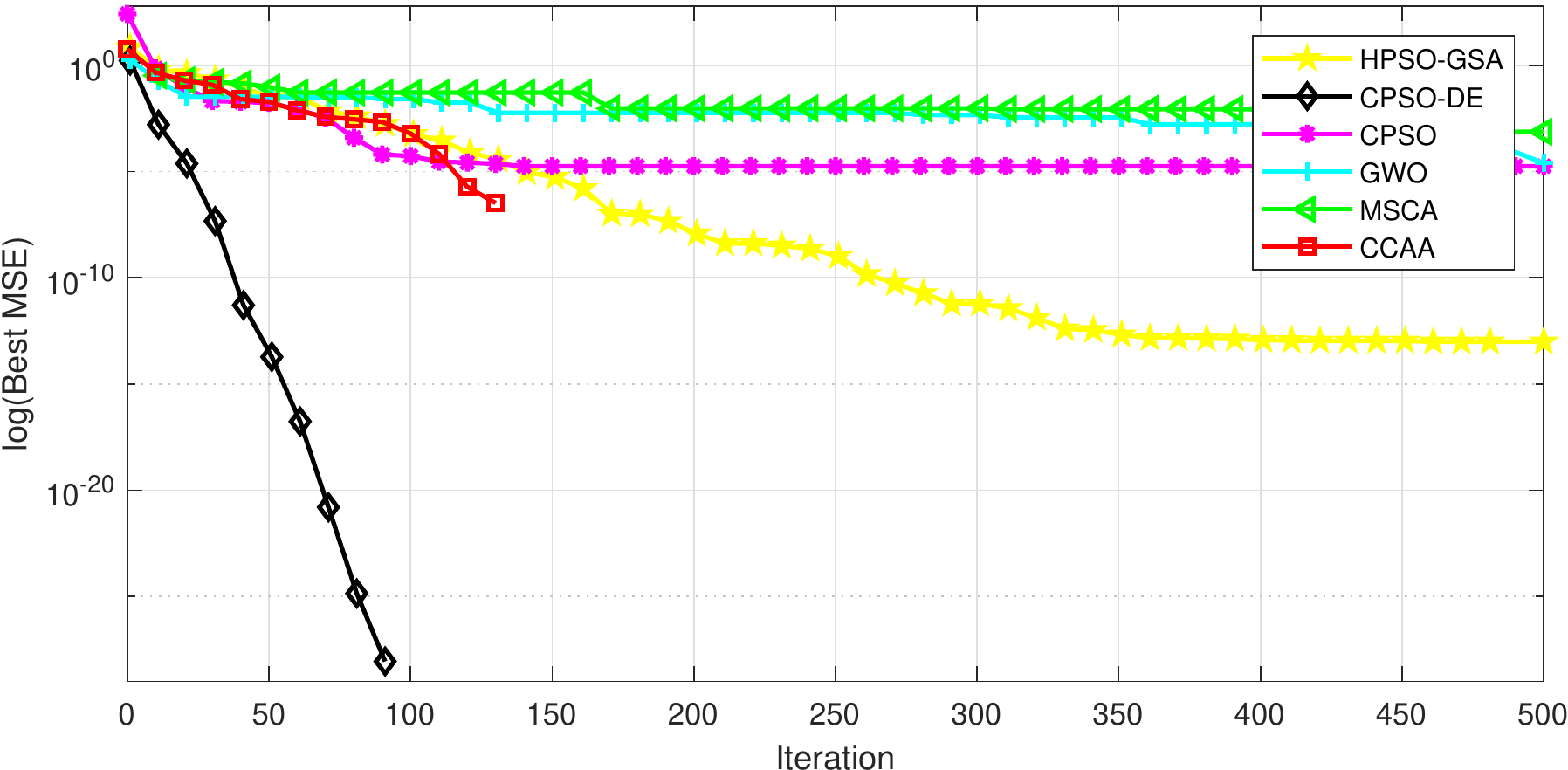} }\vspace{00mm}		
	\subfigure[Example X]{\includegraphics[scale=0.31]{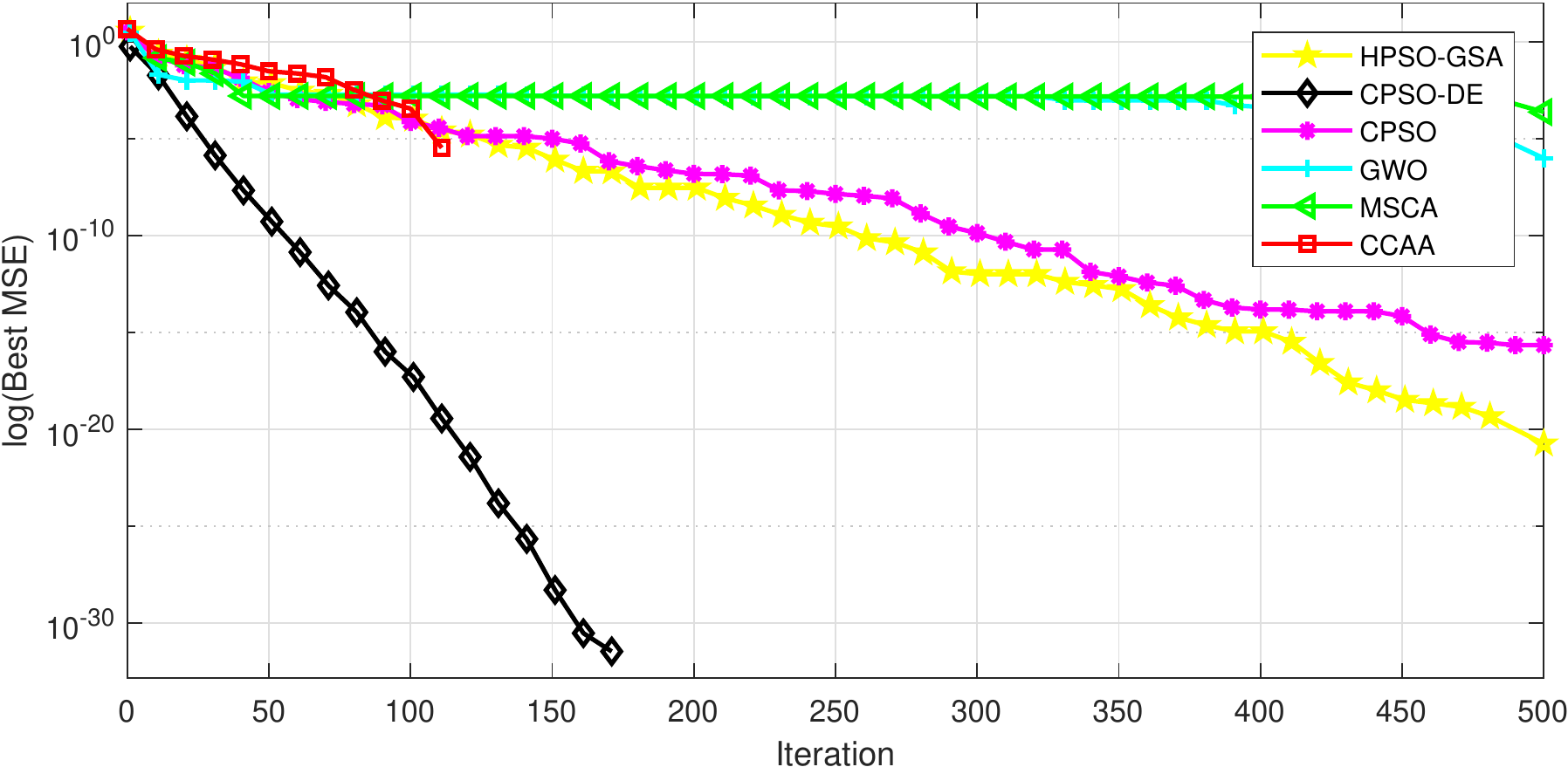}}\hspace{0mm}				
\end{figure}

The best parameter estimations for the last $ 5 $ filters are presented in Tables \ref{P_E_E6_C1} - \ref{P_E_E10_C1}, and the convergence curves in the parameter estimation of the $10$ cases are presented in Fig. \ref{fig:convergencia_filtros}. 

\begin{table}[h!]
	\small
	\begin{center}
		\caption{The parameter estimation of algorithms for Example VI}\label{P_E_E6_C1}
		\begin{spacing}{1.1}
			\scalebox{0.7}{
				\begin{tabular}{cccccccc}
					\hline
					Parameters &Actual values & \multicolumn{6}{l}{Parameter estimation value}\\
					\cline{3-8}
					& & HPSO-GSA &CPSO-DE &CPSO &GWO &MSCA &CCAA\\
					\hline
					$b_0$ &\textbf{1.00} &{1.00} &\textbf{1.00} &{1.00} &1.01 &1.01 &\textbf{1.00}\\
					$a_1$ &\textbf{1.40} &{1.40} &\textbf{1.40} &{1.40} &{1.40} &1.39 &\textbf{1.40}\\
					$a_2$ &\textbf{-0.49} &{-0.49} &\textbf{-0.49} &{-0.49} &-0.48 &-0.48 &\textbf{-0.49}\\
					\hline
				\end{tabular}
			}
		\end{spacing}
	\end{center}
\end{table}

\newpage

\begin{table}[h!]
	\small
	\begin{center}
		\caption{The parameter estimation of algorithms for Example VII }\label{P_E_E7_C1}
		\begin{spacing}{1.1}
			\scalebox{0.7}{
				\begin{tabular}{cccccccc}
					\hline
					Parameters &Actual values & \multicolumn{6}{l}{Parameter estimation value}\\
					\cline{3-8}
					& & HPSO-GSA &CPSO-DE &CPSO &GWO &MSCA &CCAA\\
					\hline
					$b_0$ &\textbf{1.00} &{1.00} &\textbf{1.00} &\textbf{1.00} &0.99 &0.99 &\textbf{1.00}\\
					$a_1$ &\textbf{1.20} &{1.20} &\textbf{1.20} &\textbf{1.20} &{1.20} &1.201 &\textbf{1.20}\\
					$a_2$ &\textbf{-0.60} &{-0.60} &\textbf{-0.60} &\textbf{-0.60} &{-0.60} &-0.61 &\textbf{-0.60}\\
					\hline
				\end{tabular}
			}
		\end{spacing}
	\end{center}
\end{table}

\begin{table}[h!]
	\small
	\begin{center}
		\caption{The parameter estimation of algorithms for Example VIII}\label{P_E_E8_C1}
		\begin{spacing}{1.1}
			\scalebox{0.7}{
				\begin{tabular}{cccccccc}
					\hline
					Parameters &Actual values & \multicolumn{6}{l}{Parameter estimation value}\\
					\cline{3-8}
					& & HPSO-GSA &CPSO-DE &CPSO &GWO &MSCA &CCAA\\
					\hline
					$b_1$ &\textbf{1.25} &{1.25} &\textbf{1.25} &\textbf{1.25} &{1.25} &1.24 &\textbf{1.25}\\
					$b_2$ &\textbf{-0.25} &{-0.25} &\textbf{-0.25} &\textbf{-0.25} &{-0.25} &-0.24 &\textbf{-0.25}\\
					$a_1$ &\textbf{0.30} &{0.30} &\textbf{0.30} &\textbf{0.30} &{0.30} &0.29 &\textbf{0.30}\\
					$a_2$ &\textbf{-0.40} &{-0.40} &\textbf{-0.40} &\textbf{-0.40} &{-0.40} &{-0.40} &\textbf{-0.40}\\
					\hline
				\end{tabular}
			}
		\end{spacing}
	\end{center}
\end{table}

\begin{table}[h!]
	\small
	\begin{center}
		\caption{The parameter estimation of algorithms for Example IX }\label{P_E_E9_C1}
		\begin{spacing}{1.1}
			\scalebox{0.7}{
				\begin{tabular}{cccccccc}
					\hline
					Parameters &Actual values & \multicolumn{6}{l}{Parameter estimation value}\\
					\cline{3-8}
					& & HPSO-GSA &CPSO-DE &CPSO &GWO &MSCA &CCAA\\
					\hline
					$b_0$ &\textbf{1.000} &0.999 &\textbf{1.000} &0.999 &0.995 &0.969 &\textbf{1.000}\\
					$a_1$ &\textbf{1.500} &{1.500} &\textbf{1.500} &1.501 &1.507 &1.524 &\textbf{1.500}\\
					$a_2$ &\textbf{-0.750} &-0.750 &\textbf{-0.750} &-0.751 &-0.761 &-0.786 &\textbf{-0.750}\\
					$a_3$ &\textbf{0.125} &0.125 &\textbf{0.125} &0.125 &0.130 &0.140 &\textbf{0.125}\\
					\hline
				\end{tabular}
			}
		\end{spacing}
	\end{center}
\end{table}

\begin{table}[h!]
	\small
	\begin{center}
		\caption{The parameter estimation of algorithms for Example X}\label{P_E_E10_C1}
		\begin{spacing}{1.1}
			\scalebox{0.7}{
				\begin{tabular}{cccccccc}
					\hline
					Parameters &Actual values & \multicolumn{6}{l}{Parameter estimation value}\\
					\cline{3-8}
					& & HPSO-GSA &CPSO-DE &CPSO &GWO &MSCA &CCAA\\
					\hline
					$b_0$ &\textbf{-0.300} &{-0.300} &\textbf{-0.300} &{-0.300} &-0.300 &-0.328 &\textbf{-0.300}\\
					$b_1$ &\textbf{0.400} &{0.400} &\textbf{0.400} &{0.400} &0.400 &0.422 &\textbf{0.400}\\
					$b_2$ &\textbf{-0.500} &{-0.500} &\textbf{-0.500} &{-0.500} &-0.500 &-0.522  &\textbf{-0.500}\\
					$a_1$ &\textbf{1.200} &{1.200} &\textbf{1.200} &{1.200} &1.194 &1.169 &\textbf{1.200}\\
					$a_2$ &\textbf{-0.500} &{-0.500} &\textbf{-0.500} &{-0.500} &-0.487 &-0.471 &\textbf{-0.500}\\
					$a_3$ &\textbf{0.100}  &{0.100} &\textbf{0.100} &{0.100} &0.093 &0.096 &\textbf{0.100}\\
					\hline
				\end{tabular}
			}
		\end{spacing}
	\end{center}
\end{table}

\newpage

In summary, for cases I, II, and VI to X, the CCAA algorithm achieved a complete parameter identification, and for cases I and VI to X, the CCAA obtained an MSE of $0$ in all the statistics. In general, for the 10 cases, the CCAA, CPSO-DE and CPSO were the only algorithms with a complete parameter identification, the CPSO-DE in $8$ cases and the CCAA in $7$ cases. The CCAA had $6$ best average values, and the CPSO-DE had $7$ best average values, resulting the best algorithms overall. These results demonstrate that the CCAA can be very useful for the design of adaptive IIR filters, and that the meta-heuristics applying cellular automata concepts can be extremely helpful to optimize challenging problems. Note that the source codes of the CCAA algorithm can be downloaded from \url{https://github.com/juanseck/CCAA.git}.

\section{Conclusions and future work}
\label{sec:conclusiones}

This study presented a new global optimization algorithm called CCAA, which uses a neighborhood and evolution rules inspired by cellular automata concepts, such as local interactions between solutions and different evolution rules. These elements allow the exploration and exploitation operations to be carried out concurrently and randomly for each $ smart\_cell $ in the solution population.

This strategy allows the CCAA to establish a proper balance of local and global changes and information exchange between $ smart\_cells $ during the search for the optimal solution. The experimental validation was carried out with $ 33 $ test functions that were widely used in recent literature to analyze the exploration, exploitation, escape of local optima, and the convergence behavior of the proposed algorithm, analyzing the performance of the CCAA for problems on 30, 500, and fixed dimensions. The experimental results obtained were compared with several recent meta-heuristic algorithms recognized for their excellent performance. The comparison showed that the CCAA is competitive enough to calculate an optimal solution for most problems.

To verify the performance of the CCAA, $5$ engineering design problems (GTD, PVD, WBD, CBD, and IIR filter design) were also addressed. The results also demonstrated the effectiveness of the CCAA to find quality solutions in these types of problems, showing that the CCAA is very competitive with respect to recently published meta-heuristic optimizers.

In future applications, it is intended to apply improvements of the CCAA  for other engineering problems, such as the design of reduced-order IIR filters. This article proposed the application of the CCAA to solve single-objective optimization problems. However, many problems involve multiple objectives for real situations. In this sense, a proposal for future work is to use the CCCAA to optimize continuous and discrete multi-objective problems, such as task scheduling problems, route scheduling, aeronautical modeling problems, classification problems, power dispatch problems, among others. The application of different and diverse evolution rules is also proposed to improve the performance of the CCAA.

\section*{Acknowledgement}
This study was supported by the National Council for Science and Technology (CONACYT) with project number CB- 2017-2018-A1-S-43008. Nadia S. Zuñiga-Peña was supported by CONACYT grant number 785376.

\bibliographystyle{elsarticle-harv}
\bibliography{ReferencesCCAA}


\end{document}